\documentclass{article}
\usepackage{amsmath, graphicx, bbm, amssymb, amsthm, cite}
\usepackage{authblk}
\usepackage[a4paper, centering, scale=0.8]{geometry}
\usepackage[pdfborder={0 0 0},colorlinks=true,linkcolor=blue,CJKbookmarks=true]{hyperref}
\usepackage{aliascnt}

\numberwithin{equation}{section}
\theoremstyle{plain}

\newaliascnt{theorem}{definition}
\newtheorem{theorem}[theorem]{\hspace{2em}Theorem}
\aliascntresetthe{theorem}
\newaliascnt{lemma}{definition}
\newtheorem{lemma}[lemma]{\hspace{2em}Lemma}
\aliascntresetthe{lemma}
\newaliascnt{proposition}{definition}
\newtheorem{proposition}[proposition]{\hspace{2em}Proposition}
\aliascntresetthe{proposition}
\newaliascnt{corollary}{definition}
\newtheorem{corollary}[corollary]{\hspace{2em}Corollary}
\aliascntresetthe{corollary}
\newaliascnt{example}{definition}

\aliascntresetthe{example}
\newaliascnt{assumption}{definition}

\aliascntresetthe{assumption}
\newtheorem{remark}{\hspace{2em}Remark}[section]

\title{High-Order Gaussian Approximation of Besov Functions on Manifolds $^\dag$\footnotetext{\dag~The work of Lei Shi is supported by the National Natural Science Foundation of China (Grants Nos. 12171093 and 12571099). The work of Xiaoming Yuan is supported by the Croucher Senior Research Fellowship.}}

\author[1]{Guanhang Lei}
\author[2]{Lei Shi}
\author[1]{Xiaoming Yuan}

\affil[1]{Department of Mathematics, \linebreak
The University of Hong Kong, Hong Kong, China \linebreak
Email: ghlei@hku.hk, xmyuan@hku.hk}

\affil[2]{School of Mathematical Sciences, \linebreak
Shanghai Key Laboratory for Contemporary Applied Mathematics, \linebreak
Center for Applied Mathematics, \linebreak
Fudan University, Shanghai, China \linebreak
Email: leishi@fudan.edu.cn}

\begin{document}
	\maketitle

\begin{abstract}
    We study high-order Gaussian approximation for Besov functions $f\in B^\alpha_{2,\infty}(X)$ with $\alpha>0$ on a compact smooth embedded manifold $X$ without boundary. For each real $r<\alpha$, a suitable signed combination of Gaussian integral operators at different bandwidths gives an $H^r(X)$ error bounded by $C\gamma^{\alpha-r}\|f\|_{B^\alpha_{2,\infty}(X)}$. The squared Gaussian reproducing kernel Hilbert space norm of the approximant is bounded by $C\gamma^{-\varrho}\|f\|_{L_2(X)}^2$, where $\varrho$ is the intrinsic dimension of $X$. A uniform Sobolev expansion identifies the low-order bias terms canceled by the signed combination, and uniform Sobolev stability controls the error in the target norm. We apply these approximation results to two statistical learning problems. For regularized Gaussian logistic classification, we combine the approximation bounds with a localized oracle inequality. For a potentially unbounded Bayes function in $B^\alpha_{2,\infty}(X)$ and a marginal density in $L_p(X)$, the excess logistic risk has upper bounds of order $\bigl((\log n)^{\varrho+1}/n\bigr)^{2\beta/(2\beta+3\varrho)}$ for the unclipped predictor and $\bigl((\log n)^{\varrho+2}/n\bigr)^{\beta/(\beta+\varrho)}$ for clipped predictions with suitably selected parameters, where $\beta=2\alpha(1-1/p)$. Under bounded marginal densities, our excess misclassification risk bound matches, up to logarithmic factors, the classical Hölder-class exponent $\alpha/(2\alpha+\varrho)$. For elliptic operator learning, we assemble a Gaussian Galerkin solution operator from random manifold samples. When $\alpha>\varrho/2$, its error in the operator norm from $B^\alpha_{2,\infty}(X)$ to $H^1(X)$ is bounded with high probability by two terms. The trial-space term has a decay exponent in the number of Gaussian centers that can approach $(\alpha+1)/\varrho$. The bias and sampling term is of order $n^{-2N/(4N+\varrho)}$ for a fixed trial space, where $n$ is the number of sample pairs and $N$ is the number of Gaussian scales used for matrix assembly. The convergence rates for both problems depend on the intrinsic dimension $\varrho$ and are independent of the ambient dimension $d$.
\end{abstract}


{\textbf{MSC2020 subject classifications:} 41A25, 41A35, 62H30, 65N30.}

\section{Introduction}\label{sec:introduction}

Learning theory studies how to infer the structure of an unknown function from sample data. In high-dimensional spaces, learning algorithms often face the curse of dimensionality, as the sample size $n$ required to achieve a given accuracy can grow rapidly with the ambient dimension $d$. This motivates the manifold assumption underlying many manifold learning methods, namely that the data lie on or near a low-dimensional manifold embedded in the ambient space $\mathbb{R}^d$. A central question is then whether learning rates can be governed by the intrinsic dimension $\varrho$ rather than the ambient dimension $d$. 

Kernel methods, including support vector machines (SVMs), are a classical family of machine learning algorithms with a solid mathematical foundation in reproducing kernel Hilbert space (RKHS) theory \cite{Aronszajn1950Theory,Steinwart2008Support}. The choice of kernel determines the hypothesis space and its approximation properties. In this paper, we focus on the Gaussian kernel, which is widely used and has been extensively studied in numerical analysis and statistical learning. Two quantities are essential when it comes to the Gaussian approximation analysis: the approximation error must reflect the smoothness of the target in the norm relevant to the application, and the norm in the Gaussian RKHS must remain controlled as the bandwidth decreases. 

On Euclidean domains, classical work on radial basis functions includes high-order error estimates for Gaussian interpolation of scattered data \cite{Wu1993Local} and meshfree methods for solving partial differential equations by multiquadric collocation \cite{Kansa1990Multiquadrics}. In statistical learning, the analysis of SVMs motivated estimates that control approximation error together with the RKHS norm. Such estimates are then used to establish learning rates for hinge loss classification, classification with general convex losses, least-squares regression and correntropy loss regression. See \cite{Steinwart2007Fast,Xiang2009Classification,Eberts2013Optimal,Lv2021Optimal}. 

For manifold learning, the Gaussian kernel on a manifold measures similarity through Euclidean distances, while the volume at the kernel scale is governed by the intrinsic dimension. Despite the dependence on the ambient metric, existing studies show that Gaussian kernel SVMs can overcome the curse of ambient dimensionality under suitable manifold and regularity assumptions, with learning rates governed by the intrinsic dimension \cite{Ye2008Learning, Ye2009SVM,Yang2016Bayesian,Hamm2021Adaptive,Lei2024Pairwise,Tang2026Adaptive}. Related results have also been established for operator learning on manifolds \cite{Gilani2019Approximating, Cheng2026Learninga}.

The early work in \cite{Ye2008Learning,Ye2009SVM} develops the connection between Gaussian approximation and learning on manifolds and establishes approximation estimates and learning bounds governed by the manifold dimension. However, the approximation results in \cite{Ye2008Learning,Ye2009SVM} identify a saturation issue. For the direct Gaussian integral approximation, increasing the regularity $\alpha$ of the target beyond $\alpha = 2$ does not in general improve the approximation order. On a manifold, the error includes contributions from the target function, the volume density, and the difference between Euclidean and geodesic distances. Estimating these contributions separately without correcting their combined effect leaves a second-order error even for smoother target functions. 

Subsequent work in \cite{Yang2016Bayesian} studies Gaussian process regression on manifolds and derives intrinsic-dimensional RKHS entropy estimates, but the accompanying approximation result remains restricted to smoothness at most two, leaving the second-order saturation issue unresolved.

The work \cite{Hamm2021Adaptive} introduces a more general box-counting dimension condition. By expressing the low-dimensional structure through covering numbers of the input domain, their work obtains learning rates for Gaussian SVMs on supports that need not be smooth manifolds. Related ideas are developed for pairwise learning in our earlier work \cite{Lei2024Pairwise}, where oracle inequalities and a box-counting dimension assumption yield fast rates for Gaussian ranking estimators. In their squared-loss approximation analyses, both works use signed combinations of Gaussian kernels with different bandwidths to cancel lower-order approximation terms. Under their smoothness assumptions on extensions to the ambient Euclidean space, these constructions achieve arbitrarily high approximation orders and thus overcome second-order saturation of the approximation error. However, the use of ambient-space extensions yields squared RKHS norm bounds of order $\gamma^{-d}$ for regression, where $\gamma$ denotes the Gaussian bandwidth. As we explain in Remark~\ref{rem:learning-rate-comparison}, this dependence on the ambient dimension would also enter the learning rates obtained by combining these bounds with our oracle inequality for logistic loss. 

The work \cite{Gilani2019Approximating} uses local kernels adapted to the drift and diffusion coefficients of an elliptic operator on a manifold. Discretizing the kernel integral on a random point cloud yields a linear system whose right-hand side consists of the source values at the sampled points. For sufficiently smooth solutions, they bound the maximum solution error over these points in probability, with the intrinsic dimension entering the sampling terms. The work \cite{Cheng2026Learninga} approximates the heat semigroup by iterating a graph transition matrix constructed from Gaussian affinities. The smoothing property of the heat semigroup allows their analysis to cover bounded input functions. For bounded diffusion times and a suitable bandwidth, they obtain high-probability supremum-norm error bounds of order $n^{-2/(\varrho+6)}$ up to logarithmic factors, both at the sampled points and after extending the approximation to the manifold by kernel convolution.

Our principal inspiration is the recent work \cite{Tang2026Adaptive} on adaptive Bayesian regression with low intrinsic dimensionality. The results in \cite[Lemma 4.1 and Proposition 4.2]{Tang2026Adaptive} establish high-order Gaussian expansions and remove the smoothness restriction by recursively correcting the input function. They provide approximation of intrinsically defined H\"older functions of arbitrary positive order, with error measured in the supremum norm. By constructing the approximant directly on the manifold without extending the target function to the ambient Euclidean space, they also obtain a squared RKHS norm bound of order $\gamma^{-\varrho}$ whose growth as the bandwidth decreases is governed by the intrinsic dimension $\varrho$. Their key technique is a high-order expansion in geodesic normal coordinates that includes the target function, the squared Euclidean distance, and the volume density. Uniform bounds on the remainder and on the H\"older norms of the coefficient functions then justify recursive corrections to the input that cancel the lower-order bias terms. 

Building on the high-order Gaussian expansion in \cite{Tang2026Adaptive}, we extend the approximation framework from intrinsic H\"older classes to the broader Besov spaces $B^\alpha_{2,\infty}(X)$ for every $\alpha>0$, with the approximation error measured in the Sobolev norm $H^r(X)$ for any real $r<\alpha$. Our construction shares their principle of canceling lower-order terms in the Gaussian expansion, but implements this cancellation through a finite signed combination of Gaussian integral operators at different bandwidths. The resulting operators act directly on $f$, without evaluating its derivatives or explicitly computing geometric correction coefficients, and are bounded on $H^r(X)$ uniformly in the bandwidth. Combining this boundedness with a Sobolev operator expansion, we obtain an error bound of order $\gamma^{\alpha-r}$ by real interpolation and a dyadic spectral decomposition. The approximants also satisfy a squared RKHS norm bound of the form $C\gamma^{-\varrho}\|f\|_{L_2(X)}^2$, whose bandwidth exponent is governed by the intrinsic dimension and whose dependence on the target is through its $L_2(X)$ norm alone. At the same smoothness order, this extension admits a larger class of targets, including potentially unbounded functions at low smoothness. Related approximation and regression guarantees for deep neural networks over Besov classes are also established in \cite{Suzuki2019Adaptivity, Suzuki2021Deepa, Tsuji2021Estimation}.

We illustrate the utility of our approximation results in two applications.

The first application is regularized logistic classification. The log-odds Bayes function
\[
    f^*=\log\frac{\eta}{1-\eta}
\]
can diverge when the conditional class probability $\eta$ approaches zero or one. Standard oracle inequalities based on fixed-level clipping, such as \cite[Theorem 7.22]{Steinwart2008Support}, are not directly applicable to logistic loss. Some existing analyses, such as \cite{Farrell2021Deep}, impose a boundedness assumption on the Bayes function $f^*$, which is equivalent to requiring the conditional class probability to stay uniformly away from zero and one. A fundamental difficulty in the analysis of logistic loss is therefore to obtain effective approximation and concentration bounds without imposing such a boundedness assumption on the Bayes function $f^*$. Logistic classification with Gaussian kernels is studied directly in \cite{Xiang2011Logistic}. Its stochastic analysis does not exploit variance bounds and hence only yields a slow rate for the misclassification error. Previous works \cite{Alquier2019Estimation, Farrell2021Deep} consider the variance estimate, but with an exponential dependence on the supremum bound for logistic loss. Hence their bound can deteriorate rapidly when regularization decreases. The unboundedness issue also motivates the work \cite{Zhang2024Classification} on classification with deep neural networks and logistic loss. 

The concentration argument in this paper follows the localization strategy developed in \cite{Steinwart2006new}. We adapt this framework using a logistic Bernstein-type inequality whose constant grows only linearly with the supremum norm of the predictor. This variance control permits an oracle analysis without a uniform bound on the Bayes function $f^*$ and connects the statistical estimate to our manifold approximation bounds. We further use this linear variance dependence to obtain faster rates by clipping predictions at a level that grows logarithmically with the sample size. 

The second application is the approximation of the elliptic solution operator $(\kappa I-\Delta_X)^{-1}$ from random samples on $X$. For the Poisson equation with $C^1(X)$ sources, the results in \cite[Theorem~1]{Shi2017Convergence} bound the $H^1(X)$ error of the point integral method in terms of the kernel scale and the quadrature accuracy. Similar results in \cite[Theorem~3.17 and Remark~3.18]{Yan2023Spectral} give a high-probability $L_2(X)$ solution error bound for a spectral Galerkin method, with explicit dependence on the number of retained modes and the sample size. Quantitative error bounds for point-based approximations of elliptic equations and related operators on manifolds are also given in \cite{Liang2024Solving, Trillos2023Rates, Davydov2023Error, Sanz-Alonso2024Analysis}.

In our setting, we fix the elliptic differential operator and vary the source functions. From points sampled with a known density, we assemble the Galerkin matrix using a symmetric high-order Gaussian approximation of the Laplacian for the stiffness term. The sampled values of each new source determine its load vector. Combining the $H^1(X)$ approximation and RKHS bounds with the sampling analysis yields a high-probability operator norm estimate from $B^\alpha_{2,\infty}(X)$ to $H^1(X)$ for $\alpha>\varrho/2$, controlling the solution error uniformly for all source functions in the Besov unit ball. The estimate separates the trial-space error, quantified by the Gaussian basis bandwidth and the fill distance of the centers, from the high-order Gaussian approximation bias and sampling error. 

Our main contributions are summarized as follows.
\begin{itemize}
    \item We establish a Sobolev expansion of the Gaussian integral operator $T_\gamma$ to arbitrary order in Theorem~\ref{thm:gaussian-sobolev-expansion}. The coefficients are differential operators independent of bandwidth $\gamma$, with leading terms proportional to powers of $\Delta_X$. This expansion has an $L_2(X)$ remainder bounded by $C\gamma^{2N+2}\|f\|_{H^{2N+2}(X)}$. For every real $r$, we also establish $H^r(X)$ stability of $T_\gamma$ in  Theorem~\ref{thm:gaussian-sobolev-stability}.
    \item We construct a signed multiscale Gaussian approximant $S_{\gamma,s}f$ that cancels the lower-order terms in the Sobolev expansion. For $f\in B^\alpha_{2,\infty}(X)$ with $\alpha>0$, any real $r<\alpha$, and an integer $s>\alpha+\max\{0,-r\}$, Theorems~\ref{thm:rkhs-norm} and \ref{thm:besov-approximation} give, for $0<\gamma\leq1$,
    \[
        \begin{aligned}
            \|S_{\gamma,s}f\|_{\mathcal H_\gamma(X)}
            &\leq C\gamma^{-\varrho/2}\|f\|_{L_2(X)},\\
            \|S_{\gamma,s}f-f\|_{H^r(X)}
            &\leq C\gamma^{\alpha-r}\|f\|_{B^\alpha_{2,\infty}(X)}.
        \end{aligned}
    \]
    Thus the construction removes second-order saturation while retaining a squared RKHS norm bound of order $\gamma^{-\varrho}$.
    \item For regularized logistic classification, we establish an oracle inequality for logistic loss without assuming a uniform bound on the Bayes function $f^*$. The proof uses a Bernstein-type bound whose constant grows only linearly with the supremum norm of the predictor. Theorem~\ref{thm:oracle-inequality} applies to separable normed function spaces, with constants uniform in the relevant entropy exponent. Corollary~\ref{cor:truncated-entropy-oracle} further allows covering estimates that are available only above a positive radius $b$, retaining the additional error $b^2/\lambda$. We combine the oracle and Gaussian approximation bounds with the manifold covering estimate in Proposition~\ref{prop:manifold-entropy}. Suppose that $f^*$ has a measurable extension in $B^\alpha_{2,\infty}(X)$ for $\alpha>0$ and the marginal density satisfies $g\in L_p(X)$ for $2\leq p\leq\infty$. With $\beta:=2\alpha(1-1/p)$ and suitable bandwidth and regularization choices, Theorems~\ref{thm:besov-learning-rate} and \ref{thm:clipped-learning-rate} give high-probability logistic excess-risk bounds of orders
    \[
        \left(\frac{(\log n)^{\varrho+1}}{n}\right)^{\frac{2\beta}{2\beta+3\varrho}}
        \quad\text{and}\quad
        \left(\frac{(\log n)^{\varrho+2}}{n}\right)^{\frac{\beta}{\beta+\varrho}}
    \]
    for the unclipped predictor and the predictor clipped at $M_n=3\log n$, respectively. For bounded densities, the clipped bounds yield polynomial exponents $2\alpha/(2\alpha+\varrho)$ for excess logistic risk and $\alpha/(2\alpha+\varrho)$ for excess misclassification risk. The latter matches the classical minimax exponent for binary classification in dimension $\varrho$ without an additional noise condition.
    \item We approximate the elliptic solution operator $G=(\kappa I-\Delta_X)^{-1}$ from $n$ random point pairs $(X_i,Y_i)$ sampled from a known density bounded away from zero and the values of each source at $X_i$. The symmetric Gaussian approximation of $\Delta_X$ in Lemma~\ref{lem:pde-centered-consistency} uses $N$ Gaussian bandwidths, yielding bias of order $\gamma^{2N}$ for smooth functions in the stiffness matrix assembly. The $H^1(X)$ approximation and RKHS norm bounds quantify the accuracy of the Gaussian trial space. For $\alpha>\varrho/2$ and integer $s>\max\{1,\varrho/2\}$, fix a Gaussian trial space $W$ with $J$ quasi-uniform centers. For $0<\delta<1$ and sufficiently large $n/\log(2/\delta)$, the bandwidth choices in Corollaries~\ref{cor:pde-operator-rate} and \ref{cor:pde-besov-trial-bound} yield, with probability at least $1-\delta$,
    \[
        \|\widehat G_{n,\gamma,W}-G\|_{B^\alpha_{2,\infty}(X)\to H^1(X)}
        \leq C J^{-\frac{(\alpha+1)(s-1)}{\varrho(s+\alpha+1+\varrho/2)}}
        + C_W\left(\frac{\log(2/\delta)}{n}\right)^{\frac{2N}{4N+\varrho}}.
    \]
    The exponent of the trial-space term approaches $(\alpha+1)/\varrho$ as $s$ increases. The second term controls the approximation bias and the sampling error and its exponent approaches $1/2$ as $N$ increases.
\end{itemize}

The rest of this paper is organized as follows. Section~\ref{sec:preliminaries} introduces the manifold setting, Gaussian kernels, and function spaces. Section~\ref{sec:manifold-approximation} presents the Gaussian approximation results. Sections~\ref{sec:logistic-application} and \ref{sec:elliptic-operator-learning} develop the logistic classification and elliptic solution operator applications. Section~\ref{sec:conclusions} concludes and discusses further questions. All proofs are collected in Appendix~\ref{sec:proofs}.

\section{Preliminaries}\label{sec:preliminaries}
This section reviews some basic notions of Gaussian kernels and function spaces on manifolds.

\subsection{Manifold setting and Gaussian kernels}
For a data domain $X \subset \mathbb{R}^d$, it is well-known that typical learning rates are subject to the curse of dimensionality, and the sample size $n$ required to achieve a given accuracy can grow rapidly with the dimension $d$. For example, the minimax nonparametric squared-loss regression rate for $f^* \in C^{\alpha}([0,1]^d)$ is $n^{-2\alpha/(2\alpha+d)}$, cf. \cite{Stone1982Optimal}. To address the curse of dimensionality, a common assumption is that the data is supported on a low-dimensional manifold $X\subset\mathbb R^d$ with intrinsic dimension $\varrho\ll d$. This motivates the study of learning and approximation on manifolds. 

We now specify the manifold assumption throughout this paper. Suppose that $X\subset\mathbb R^d$ is a connected compact $C^\infty$ embedded submanifold without boundary, of intrinsic dimension $1\leq\varrho<d$, and is equipped with the metric induced by the Euclidean embedding. We write $d_X$, $V$, $\mathcal{E}_x$, and $\operatorname{inj}(X)$ for its geodesic distance, volume measure, exponential map at $x \in X$, and injectivity radius. Compactness implies $\operatorname{inj}(X)>0$ and provides a finite collection of relatively compact coordinate neighborhoods carrying smooth orthonormal frames.

For a bandwidth $\gamma>0$, let $\mathcal{H}_\gamma(X)$ be the RKHS of the Gaussian kernel
\[
    k_\gamma(x,y):=\exp\left(-\frac{\|x-y\|^2_{\mathbb{R}^d}}{\gamma^2}\right), \quad x,y\in X.
\]
$\mathcal{H}_\gamma(X)$ is separable and consists of continuous functions. Since $k_\gamma(x,x)=1$, the reproducing property gives
\[
    \|f\|_\infty\leq\|f\|_{\mathcal{H}_\gamma(X)}, \quad f\in\mathcal{H}_\gamma(X).
\]
For a detailed analysis of $\mathcal{H}_\gamma(X)$, see \cite[Section 4.4]{Steinwart2008Support}.

\subsection{Function spaces on manifolds}\label{sec:function-spaces-on-manifolds}

We now introduce the function spaces used to measure regularity on manifolds. The Sobolev and Besov spaces provide a regularity framework for stating our approximation bounds. Let $\Delta_X=\operatorname{div}\nabla$ be the Laplace--Beltrami operator on $X$, with $\nabla$ denoting the Levi-Civita covariant derivative. For $r\in\mathbb R$, define Sobolev spaces $H^r(X)$ by completion of $C^\infty(X)$ in the spectral norm
\[
    \|f\|_{H^r(X)}:=\|(I-\Delta_X)^{r/2}f\|_{L_2(X)}.
\]
Thus $H^0(X)=L_2(X)$, and for $r>0$, $H^{-r}(X)$ is the dual of $H^r(X)$ under the $L_2(X)$ pairing. For integer $m \geq 0$, the spectral norm is equivalent to the covariant derivative norm:
\[
    \|f\|_{H^m(X)}^2 \asymp \sum_{\ell=0}^m\int_X |\nabla^\ell f(x)|^2\,dV(x),
\]
where all pointwise tensor norms are induced by the Riemannian metric on $X$. See \cite[Theorem~7.4.5(ii)]{Triebel1992Theory}.

For $f\in L_2(X)+H^m(X)$, define the Peetre $K$-functional by
\[
    K(t,f;L_2(X),H^m(X)) := \inf_{\substack{f=f_0+f_1\\ f_0\in L_2(X),\ f_1\in H^m(X)}} \bigl\{\|f_0\|_{L_2(X)}+t\|f_1\|_{H^m(X)}\bigr\}.
\]
Given $\alpha>0$, choose any integer $m>\alpha$ and define
\begin{equation}\label{eq:besov-interpolation}
    \begin{aligned}
        B^\alpha_{2,\infty}(X) &:= (L_2(X),H^m(X))_{\alpha/m,\infty}, \\
        \|f\|_{B^\alpha_{2,\infty}(X)} &\asymp \sup_{t>0}t^{-\alpha/m}K(t,f;L_2(X),H^m(X)).
    \end{aligned}
\end{equation}
The real-interpolation identity in \eqref{eq:besov-interpolation} and the independence, up to equivalence of norms, of the choice of $m>\alpha$ are standard. See \cite[Chapter 3]{Bergh1976Interpolation} and \cite[Chapter 7]{Triebel1992Theory}.

For $j\in\mathbb N_0$, let $P_j$ be the spectral projection of $(I-\Delta_X)^{1/2}$ onto $[2^j,2^{j+1})$. Equivalently, $P_j$ selects eigenmodes of $-\Delta_X$ with eigenvalues in $[2^{2j}-1,2^{2(j+1)}-1)$. Orthogonality yields
\begin{equation}\label{eq:sobolev-spectral-blocks}
    \|f\|_{H^r(X)}^2\asymp\sum_{j\geq0}2^{2jr}\|P_jf\|_{L_2(X)}^2,\quad r\in\mathbb R,
\end{equation}
and the Besov norm has the equivalent spectral form
\begin{equation}\label{eq:besov-spectral}
    \|f\|_{B^\alpha_{2,\infty}(X)}
    \asymp\sup_{j\geq0}2^{j\alpha}\|P_jf\|_{L_2(X)},\quad \alpha>0,
\end{equation}
See \cite[Theorem 1.1]{Pesenson2015Approximations}. The spectral bounds give $B^\alpha_{2,\infty}(X)\hookrightarrow H^r(X)$ for $r<\alpha$. When $r>\varrho/2$, Sobolev embedding gives $H^r(X)\hookrightarrow C^{0,\theta}(X)$ for $0<\theta<\min\{1,r-\varrho/2\}$, and in particular point evaluation is continuous on $H^r(X)$.

We also recall the chartwise definition of the classical Hölder spaces $C^{k,\beta}(X)$. Fix a finite smooth atlas $\{(U_\nu,\phi_\nu)\}_{\nu=1}^J$ and a smooth subordinate partition of unity $\{\chi_\nu\}_{\nu=1}^J$ such that $\operatorname{supp}\chi_\nu\Subset U_\nu$. For $k\in\mathbb{N}_0$ and $0<\beta\leq1$, set
\[
    \|f\|_{C^{k,\beta}(X)} := \sum_{\nu=1}^J \bigl\|(\chi_\nu f)\circ\phi_\nu^{-1} \bigr\|_{C^{k,\beta}(\mathbb{R}^\varrho)},
\]
where each localized function is extended by zero from $\phi_\nu(U_\nu)$ to $\mathbb{R}^\varrho$ and
\[
    \|u\|_{C^{k,\beta}(\mathbb{R}^\varrho)} := \sum_{|a|\leq k}\|D^a u\|_\infty +\sum_{|a|=k} \sup_{z\neq z'}\frac{|D^a u(z)-D^a u(z')|}{|z-z'|^\beta}.
\]
Thus $C^{k,\beta}(X)$ consists of the functions for which this norm is finite. Different finite atlases and subordinate partitions give equivalent norms.

There is also a chartwise difference characterization of the Besov space $B^\alpha_{\infty,\infty}(X)$ at every order $\alpha>0$. Let
\[
    k:=\lceil\alpha\rceil-1, \quad \beta:=\alpha-k\in(0,1],
\]
and, for $u\in C^k(\mathbb{R}^\varrho)$, set
\[
    \begin{aligned}
        \|u\|_{B^\alpha_{\infty,\infty}(\mathbb{R}^\varrho)} &:= \sum_{|a|\leq k}\|D^a u\|_\infty +\sum_{|a|=k} \sup_{h\neq0}|h|^{-\beta} \|\Delta_h^2D^a u\|_\infty, \\
        \Delta_h^2u(z)&:=u(z+2h)-2u(z+h)+u(z).
    \end{aligned}
\]
Then, using the same atlas and partition of unity, define
\[
    \|f\|_{B^\alpha_{\infty,\infty}(X)}:= \sum_{\nu=1}^J \bigl\|(\chi_\nu f)\circ\phi_\nu^{-1}\bigr\|_{B^\alpha_{\infty,\infty}(\mathbb{R}^\varrho)},
\]
with zero extension. This defines $B^\alpha_{\infty,\infty}(X)$. Up to equivalence of norms, it is independent of the atlas and partition. See \cite[Section 1.2.2, Definition 7.2.2, Theorem 7.2.3, and Theorem 7.3.1(ii)]{Triebel1992Theory}. If $\alpha=k+\beta$ with $0<\beta<1$, this is the classical $C^{k,\beta}(X)$ space with an equivalent norm. If $\alpha=k+1$, however, $B^{k+1}_{\infty,\infty}(X)$ is the Zygmund class which contains $C^{k,1}(X)$ continuously and is generally larger.

Using $B^\alpha_{\infty,\infty}(X)$ as a bridge, we can prove that the Besov class $B^\alpha_{2,\infty}(X)$ is indeed very rich and contains the Hölder class $C^{k,\beta}(X)$ continuously.

\begin{proposition}\label{prop:holder-besov-embedding}
    Let $k\in\mathbb{N}_0$, $\beta\in(0,1]$, and $\alpha=k+\beta$. Then
    \[
        C^{k,\beta}(X)\hookrightarrow B^\alpha_{2,\infty}(X).
    \]
    More precisely, there is a constant $C=C(X,k,\beta)$ such that
    \[
        \|f\|_{B^\alpha_{2,\infty}(X)} \leq C\|f\|_{C^{k,\beta}(X)} \quad \text{for all }f\in C^{k,\beta}(X).
    \]
\end{proposition}

\begin{remark}\label{rem:tang-holder-class}
    Covariant derivatives at two different points lie in different tensor spaces, so their difference is not well-defined until those spaces are identified. The work \cite{Tang2026Adaptive} uses parallel transport to define an atlas-free Hölder norm, cf. \cite[Section 2.4 and Definition 1]{Tang2026Adaptive}. In their notation, if $\xi$ is the injectivity radius and $P_{x,y}:T_xX\to T_yX$ denotes parallel transport along the unique minimizing geodesic from $x$ to $y$, where $d_X(x,y)<\xi$, their highest-order seminorm is
    \[
        L_{k,\beta}(f) := \sup_{x\in X} \sup_{\substack{0<d_X(x,y)<\xi}} \sup_{\substack{v\in T_xX\\|v|=1}} \frac{ \bigl|\nabla_v^k f(x) -\nabla_{P_{x,y}v}^k f(y)\bigr| }{d_X(x,y)^\beta},
    \]
    and their Hölder norm is
    \[
        \sup_{x\in X}|f(x)|
        +\sum_{\ell=1}^k \sup_{x\in X}\sup_{\substack{v\in T_xX\\|v|=1}}
          |\nabla_v^\ell f(x)|
        +L_{k,\beta}(f).
    \]
    Here $\nabla_v^\ell f(x)$ denotes the covariant derivative $(\nabla^\ell f)_x$ evaluated on $\ell$ copies of the same unit vector $v$.
    On a fixed compact smooth manifold, this covariant norm and the chartwise norm in Proposition~\ref{prop:holder-besov-embedding} are equivalent. The parallel-transport formulation is used in \cite{Tang2026Adaptive} because it is intrinsic and is directly adapted to Taylor expansion along geodesics. In particular, the result in \cite[Lemma D.13]{Tang2026Adaptive} gives, for a unit vector $\theta\in T_xX$ and $0\leq t<\xi$,
    \[
        f(\exp_x(t\theta)) = \sum_{j=0}^k\frac{t^j}{j!}\nabla_\theta^j f(x) +r_{f,x,\theta}(t), \quad |r_{f,x,\theta}(t)| \leq \frac{L_{k,\beta}(f)}{k!}\,t^{k+\beta}.
    \]
\end{remark}

\section{High-order Gaussian approximation on manifolds}\label{sec:manifold-approximation}

We construct Gaussian RKHS approximants that fully exploit the Besov smoothness $\alpha$ while retaining RKHS norm bounds governed by the intrinsic dimension $\varrho$. The construction uses a signed combination of Gaussian integral operators at different bandwidths. 

Define the Gaussian integral operator
\[
    T_\gamma f(x) := \left(\frac{1}{\gamma\sqrt{\pi}}\right)^\varrho \int_X \exp\left(-\frac{\|x-y\|^2_{\mathbb{R}^d}}{\gamma^2}\right)f(y)\,dV(y).
\]
Results in \cite[Lemma 4.1 and Proposition 4.2]{Tang2026Adaptive} use a one-bandwidth expansion in $C^{k,\beta}(X)$ and correct the input function recursively. Our following signed combination cancels the common expansion coefficients across the bandwidths $j\gamma$. For an integer $s\geq1$, set
\[
    c_j:=(-1)^{j+1}\binom{s}{j},\quad 1\leq j\leq s,
\]
and define the multiscale approximation operator
\begin{equation}\label{eq:multiscale-operator}
    S_{\gamma,s}f := \sum_{j=1}^s c_jT_{j\gamma}f.
\end{equation}
The coefficients satisfy the finite-difference identities
\begin{equation}\label{eq:binomial-moments}
    \sum_{j=1}^s c_j=1, \quad \sum_{j=1}^s c_jj^\ell=0, \quad 1\leq\ell<s.
\end{equation}
The following theorem bounds the RKHS norm of $S_{\gamma,s}f$.
\begin{theorem}\label{thm:rkhs-norm}
    For every $f\in L_2(X)$ and $\gamma>0$,
    \[
        \|S_{\gamma,s}f\|_{\mathcal{H}_\gamma(X)} \leq C_{s,X,d}\gamma^{-\varrho/2}\|f\|_{L_2(X)}.
    \]
\end{theorem}

We next establish an operator expansion in which all operator coefficients are independent of the bandwidth. This result is a Sobolev counterpart of \cite[Lemma 4.1]{Tang2026Adaptive}. They establish a uniform expansion for functions in $C^{k,\beta}(X)$ with the remainder controlled in the supremum norm. Here, we establish an $L_2(X)$ remainder bound of order $\gamma^{2N+2}$ controlled by $\|f\|_{H^{2N+2}(X)}$, and identify the differential operator coefficients. This Sobolev formulation enables the interpolation argument used below to obtain approximation bounds on Besov spaces.

\begin{theorem}\label{thm:gaussian-sobolev-expansion}
    Let $\operatorname{Diff}^{m}(X)$ denote the space of linear differential operators on $X$ of order at most $m$ with smooth coefficients. Then, there exist differential operators $A_r$, $r \geq 0$, depending only on $X$, such that $A_0=I$, $A_r\in\operatorname{Diff}^{2r}(X)$, and, for every $N\in\mathbb{N}_0$, there exists a constant $C_{N,X,d}>0$ satisfying
    \begin{equation}\label{eq:uniform-sobolev-expansion}
        \left\| T_\gamma f-\sum_{r=0}^{N}\gamma^{2r}A_rf \right\|_{L_2(X)} \leq C_{N,X,d}\gamma^{2N+2}\|f\|_{H^{2N+2}(X)}
    \end{equation}
    for all $0<\gamma\leq1$ and $f\in H^{2N+2}(X)$. Moreover,
    \[
        A_r = \frac{1}{4^rr!}\Delta_X^r+B_r, \quad B_r\in\operatorname{Diff}^{2r-2}(X), \quad r\geq1.
    \]
\end{theorem}

The next theorem controls the Gaussian operators in Sobolev norms independently of the bandwidth. It allows the expansion to be used for approximation in stronger norms and, by duality, in negative-order norms.

\begin{theorem}\label{thm:gaussian-sobolev-stability}
    For every $r\in\mathbb R$, there is a constant $C_{r,X,d} > 0$ such that $T_\gamma$ extends to a bounded operator on $H^r(X)$ and
    \[
        \|T_\gamma f\|_{H^r(X)}\leq C_{r,X,d}\|f\|_{H^r(X)},\quad
        0<\gamma\leq1,\quad f\in H^r(X).
    \]
\end{theorem}

The identities \eqref{eq:binomial-moments} cancel the lower-order terms in the Sobolev expansion when the operators are combined in $S_{\gamma,s}$. The resulting Sobolev approximation bound and uniform $L_2(X)$ boundedness yield the following bound in a Sobolev norm of any order strictly below the Besov smoothness of the target.

\begin{theorem}\label{thm:besov-approximation}
    Let $\alpha>0$, $r\in\mathbb R$ with $r<\alpha$, and let $s$ be an integer satisfying $s>\alpha+\max\{0,-r\}$. Then there exists a constant $C_{\alpha,r,s,X,d} > 0$ such that for all $f\in B^\alpha_{2,\infty}(X)$,
    \begin{equation}\label{eq:besov-approximation}
        \|S_{\gamma,s}f-f\|_{H^r(X)}
        \leq C_{\alpha,r,s,X,d}\gamma^{\alpha-r}\|f\|_{B^\alpha_{2,\infty}(X)},
        \quad 0<\gamma\leq1.
    \end{equation}
\end{theorem}

For $0\leq r<\alpha$, one can use $s=\lfloor\alpha\rfloor+1$ throughout. In particular, $r=0$ gives the $L_2(X)$ estimate used for classification, and $r=1$ gives the energy norm estimate used for elliptic solution operators.

\section{Application to Gaussian logistic classification}\label{sec:logistic-application}

In this section, we apply the Gaussian approximation bounds to the problem of binary classification with logistic loss. The oracle inequality we establish below applies to separable normed hypothesis spaces $\mathcal{F}$ on a general input space $X$. The manifold assumptions on $X$ are used only in subsequent Gaussian risk bounds.

\subsection{The logistic classification problem}
We consider the binary classification problem where we regard a separable metric space $X$ as the input space and $Y=\{-1,1\}$ as the label space. Let $P$ be a probability distribution on $X \times Y$. We denote by $P_X$ the marginal distribution of $P$ on $X$, and we fix a measurable version of the conditional probability $\eta(x):=P(Y=1\mid X=x)$. The logistic loss function is defined by
\[
    \phi(t):=\log(1+e^{-t}).
\]
We define the logistic risk of a measurable function $f:X\to\mathbb R$ by
\[
    \mathcal R(f):=\mathbb{E}_P\phi(Yf(X)).
\]
With the convention $0\log0=0$, the Bayes logistic risk is
\[
    \mathcal R^* := \inf_{\substack{f:X\to\mathbb R\\ f\ \text{measurable}}}\mathcal R(f) = \int_X \left(-\eta\log\eta-(1-\eta)\log(1-\eta)\right)\,dP_X.
\]
When $0<\eta(x)<1$, the pointwise minimizer is
\[
    f^*(x):=\log\frac{\eta(x)}{1-\eta(x)}.
\]
We do not regard $f^*$ as a real-valued function at points where $\eta$ is $0$ or $1$. Instead, define the measurable Bayes reference loss
\[
    \psi(x,y) :=
    \begin{cases}
        \phi(yf^*(x)),&0<\eta(x)<1,\\
        0,&\eta(x)\in\{0,1\}.
    \end{cases}
\]
Then $\mathbb{E}_P\psi(X, Y)=\mathcal R^*$, including when $P_X\{\eta\in\{0,1\}\}>0$.

To learn a classifier from $n$ i.i.d. samples $D = ((X_1,Y_1),\ldots,(X_n,Y_n))$, we minimize the empirical risk on a hypothesis class $\mathcal{F}$:
\[
    \widehat f_D \in \arg\min_{f\in\mathcal{F}} \frac1n\sum_{i=1}^n \phi(Y_if(X_i)),
\]
or, usually, a regularized version:
\begin{equation}\label{eq:regularized-empirical-risk-minimization}
    \widehat f_{D,\lambda} \in \arg\min_{f\in\mathcal{F}} \left\{ \frac1n\sum_{i=1}^n \phi(Y_if(X_i)) + \lambda\|f\|^2_{\mathcal{F}}\right\},
\end{equation}
where $\lambda > 0$ is a regularization parameter and $\|\cdot\|_{\mathcal{F}}$ is a norm on $\mathcal{F}$. We aim to establish an oracle inequality to bound the excess risk $\mathcal R(\widehat f_{D,\lambda})-\mathcal R^*$ in terms of the sample size $n$, the regularization parameter $\lambda$, and the complexity of the hypothesis class $\mathcal{F}$. In particular, by choosing the regularization parameter $\lambda_n$ and the hypothesis class $\mathcal{F}_n$ appropriately depending on $n$, we can obtain an asymptotic convergence rate for the excess risk as $n\to\infty$, commonly called a learning rate.

The logistic loss serves as a convex surrogate for the $0$--$1$ classification loss to facilitate optimization. For a measurable decision function $f:X\to\mathbb R$, the misclassification risk and the Bayes risk are defined by
\[
    \begin{aligned}
        \mathcal R_{01}(f)
        &:=\mathbb{E}_P\mathbf 1_{\{\operatorname{sgn}(f(X))\neq Y\}},\\
        \mathcal R_{01}^*
        &:=\int_X\min\{\eta(x),1-\eta(x)\}\,dP_X(x).
    \end{aligned}
\]
Here we agree that $\operatorname{sgn}(0) = 1$. $\mathcal R_{01}(f)$ directly reflects the performance of classifier $f$. We therefore use comparison or calibration inequalities to translate bounds on the excess logistic risk into bounds on the excess misclassification risk. Such inequalities are widely studied in statistical learning theory. See \cite{Zhang2004Statistical,Bartlett2006Convexity,Cucker2007Learning,Steinwart2008Support}. For the logistic loss $\phi$, by \cite{Bartlett2006Convexity}, the calibration inequality takes the form
\begin{equation}\label{eq:logistic-calibration-weak}
    \mathcal R_{01}(f)-\mathcal R_{01}^*
    \leq C\sqrt{\bigl(\mathcal R(f)-\mathcal R^*\bigr)}.
\end{equation}
This calibration inequality can be improved when the distribution $P$ satisfies the Tsybakov noise condition \cite{Tsybakov2004Optimal}. See a detailed discussion in  Remark~\ref{rem:logistic-calibration}.

\subsection{Complexity measures of function classes}
The complexity of the hypothesis class $\mathcal{F}$ plays a crucial role in determining the learning rate. To describe the complexity of the function class, we introduce dyadic entropy numbers, covering numbers and empirical Rademacher complexity. For a bounded linear operator $T:E\to F$, its $i$th dyadic entropy number is
\[
    e_i(T) := \inf\left\{ \delta>0:T(B_E)\text{ is covered by }2^{i-1} \text{ balls of radius }\delta\text{ in }F \right\}.
\]
Here $B_E$ is the unit ball in $E$. For an arbitrary subset $\mathcal Q\subset F$, we use the analogous notation
\[
    e_i(\mathcal Q,F) := \inf\left\{ \delta>0:\mathcal Q\text{ is covered by }2^{i-1} \text{ balls of radius }\delta\text{ in }F \right\}.
\]
For an arbitrary subset $\mathcal Q \subset F$, define covering numbers
\[
    N(\delta,\mathcal Q,\|\cdot\|_{F}) := \min\left\{ n\geq1:\exists \, q_1,\ldots,q_n\in F\text{ such that }\mathcal Q\subset \bigcup_{i=1}^n B(q_i,\delta) \right\}.
\]
For a class $\mathcal Q$ of measurable functions on $X \times Y$, define its empirical Rademacher complexity on the sample $D$ by
\[
    \operatorname{Rad}_D(\mathcal Q) := \mathbb{E}_{\boldsymbol{\sigma}}\sup_{q\in\mathcal Q}\left|\frac1n\sum_{i=1}^n\sigma_iq(X_i,Y_i)\right|,
\]
where $\sigma_1,\ldots,\sigma_n$ are independent Rademacher variables, each taking the values $-1$ and $1$ with equal probability, and are independent of $D$.

Our oracle inequality assumes polynomial decay of the expected empirical dyadic entropy numbers on the hypothesis class $\mathcal{F}$. In particular, for each fixed bandwidth $\gamma>0$, the dyadic entropy numbers of the Gaussian RKHS $\mathcal H_\gamma(X)$ exhibit super-polynomial decay. This flexibility allows us to optimize the entropy exponent to obtain the best learning rate for the Gaussian kernel estimator.

\subsection{Oracle inequalities for potentially unbounded Bayes functions}\label{sec:oracle-inequality}

To derive an oracle inequality for regularized empirical risk minimization, typical arguments combine a Bernstein-type variance condition with a capacity assumption for the hypothesis class and concentration inequalities such as Talagrand's inequality. The variance condition controls the second moment of the excess loss in terms of the excess risk. For logistic loss, obtaining such a condition with a useful constant is nontrivial. The results in \cite[Lemma 8]{Farrell2021Deep} and \cite[Proposition 8.1]{Alquier2019Estimation} yield variance bounds with constants that grow exponentially with the uniform bound $\|f\|_\infty$. This dependence can lead to poor statistical guarantees when $\|f\|_\infty$ increases. This is particularly relevant for logistic loss, since the Bayes function $f^*(x) \to -\infty$ as $\eta(x)\to0$ and $f^*(x) \to +\infty$ as $\eta(x)\to1$.

The following lemma gives a sharp Bernstein-type bound with only linear dependence on $\|f\|_\infty$, without requiring the conditional probability $\eta$ to be uniformly bounded away from $0$ and $1$.

\begin{lemma}\label{lem:logistic-variance}
    For every bounded measurable $f:X\to\mathbb R$,
    \[
        \mathbb{E}_P\bigl(\phi(Yf(X))-\psi(X,Y)\bigr)^2 \leq (\|f\|_\infty+2)\bigl(\mathcal R(f)-\mathcal R^*\bigr).
    \]
\end{lemma}

We now state an oracle inequality for regularized empirical risk minimization over a separable normed function space. Using the Bernstein-type bound in Lemma~\ref{lem:logistic-variance}, we do not require the conditional probability $\eta$ to be bounded away from $0$ and $1$. For $D=((X_1,Y_1),\ldots,(X_n,Y_n))$, let $L_2(D)$ be equipped with the empirical norm
\[
    \|g\|_{L_2(D)}^2:=\frac1n\sum_{i=1}^n|g(X_i,Y_i)|^2, \quad g:X\times Y\to\mathbb R.
\]
Similarly, for $D_X=(X_1,\ldots,X_n)$, let $L_2(D_X)$ be equipped with the empirical norm
\[
    \|f\|_{L_2(D_X)}^2:=\frac1n\sum_{i=1}^n|f(X_i)|^2, \quad f:X\to\mathbb R.
\]

\begin{theorem}\label{thm:oracle-inequality}
    Let $\mathcal{F}$ be a separable normed linear space of bounded measurable functions $f:X\to\mathbb R$ such that
    \[
        \|f\|_\infty\leq\|f\|_{\mathcal{F}}, \quad f\in\mathcal{F}.
    \]
    For every $n\geq1$ and $0<\lambda\leq1$, assume that the regularized empirical risk minimization problem \eqref{eq:regularized-empirical-risk-minimization} admits a minimizer for every sample $D$, and fix a choice $\widehat f_{D,\lambda}\in\mathcal{F}$ that is measurable as a function of $D$. Define the approximation error by
    \begin{equation}\label{eq:approximation-error-function}
        a_{\mathcal{F}}(\lambda) := \inf_{f\in \mathcal{F}} \left\{ \lambda\|f\|_{\mathcal{F}}^2+\mathcal R(f)-\mathcal R^* \right\}.
    \end{equation}
    Furthermore, suppose that, for some $p\in(0,1/2]$ and $A\geq1$, for every $n\geq1$,
    \begin{equation}\label{eq:average-entropy-assumption}
        \mathbb{E}_{D_X\sim P_X^n} e_i\bigl(\operatorname{id}:\mathcal{F}\to L_2(D_X)\bigr) \leq A^{1/(2p)}i^{-1/(2p)}, \quad i\geq1.
    \end{equation}
    Then there is a numerical constant $C\geq1$, uniform over $p\in(0,1/2]$, such that, for every $n\geq1$, $0<\lambda\leq1$, and $t\geq1$, with probability at least $1-e^{-t}$,
    \begin{equation}\label{eq:oracle-inequality}
        \mathcal R(\widehat f_{D,\lambda})-\mathcal R^* \leq C\left[ a_{\mathcal{F}}(\lambda)+\lambda +\lambda^{-1}\left(\frac{A}{n}\right)^{2/(1+p)} +\lambda^{-1}\frac{t^2}{n^2} \right].
    \end{equation}
\end{theorem}

The entropy assumption in Theorem~\ref{thm:oracle-inequality} requires control of every entropy number. In our manifold application, the available covering estimate holds only above a positive radius, as stated in Proposition~\ref{prop:manifold-entropy}. The following corollary considers this restriction by requiring the covering estimate only for $b\leq\delta\leq1$.

\begin{corollary}\label{cor:truncated-entropy-oracle}
    Suppose that the assumptions of Theorem~\ref{thm:oracle-inequality} hold with the entropy condition \eqref{eq:average-entropy-assumption} replaced by the following covering number condition: for some $p\in(0,1/2]$, $A\geq1$, and $b\in(0,1/4]$,
    \begin{equation}\label{eq:truncated-covering-assumption}
        \log N(\delta,B_{\mathcal F},\|\cdot\|_\infty)
        \leq A\delta^{-2p},\quad b\leq\delta\leq1.
    \end{equation}
    Then there is a numerical constant $C\geq1$, independent of $p,A,b$, such that, for every $n\geq1$, $0<\lambda\leq1$, and $t\geq1$, with probability at least $1-e^{-t}$,
    \begin{equation}\label{eq:truncated-entropy-oracle}
        \mathcal R(\widehat f_{D,\lambda})-\mathcal R^*
        \leq C\left[a_{\mathcal F}(\lambda)+\lambda
        +\lambda^{-1}\left(\frac{A}{n}\right)^{2/(1+p)}
        +\frac{b^2}{\lambda}+\lambda^{-1}\frac{t^2}{n^2}\right].
    \end{equation}
\end{corollary}

\begin{remark}\label{rem:regularization-dependence}
    The factor $\lambda^{-1}$ in \eqref{eq:oracle-inequality} comes from the dependence of both the localized envelope and the variance bound on the predictor norm. In the proof of Theorem~\ref{thm:oracle-inequality}, functions in a localized class at level $\varepsilon$ have supremum norm bounded by $C(\varepsilon/\lambda)^{1/2}$. Lemma~\ref{lem:logistic-variance} then gives the second-moment bound $C\varepsilon^{3/2}\lambda^{-1/2}$. Together, these estimates yield the entropy contribution $C\lambda^{-1}(A/n)^{2/(1+p)}$ through the local Rademacher bound.

    In the squared-loss settings of \cite{Hamm2021Adaptive, Lei2024Pairwise}, fixed-level clipping gives a bounded loss class and a Bernstein constant independent of $\lambda$.  Hence, for its localized class, the variance bound is of order $\varepsilon$ and gives the entropy contribution $C\lambda^{-p}A/n$.

    This difference in the dependence on $\lambda$ explains why approximation by ambient-space extension can yield an intrinsic-dimensional polynomial learning rate in bounded squared-loss regression, whereas the same norm bounds do not suffice in our logistic analysis. We will discuss this distinction further in Remark~\ref{rem:learning-rate-comparison}.
\end{remark}

\subsection{Approximation error}

For the Gaussian RKHS, its separability and the supremum norm bound verify the function-space assumptions in the oracle inequality. For fixed $n$, $\lambda>0$, and $\gamma>0$, the Gaussian regularized empirical objective is coercive and weakly lower semicontinuous, so it admits a minimizer. Strong convexity makes this minimizer unique, and continuity of the kernel implies continuous dependence on the sample. The resulting estimator $\widehat f_{D,\lambda,\gamma}$ is therefore measurable with respect to $D$. Before we apply the oracle inequality to the Gaussian RKHS, we need to bound the approximation error in \eqref{eq:approximation-error-function} for the Bayes function $f^*$. For the bandwidth-dependent Gaussian RKHS $\mathcal{H}_\gamma(X)$, we write
\begin{equation}\label{eq:bandwidth-approximation-error}
    a_\gamma(\lambda) := \inf_{f\in \mathcal{H}_\gamma(X)} \left\{ \lambda\|f\|_{\mathcal{H}_\gamma(X)}^2 +\mathcal R(f)-\mathcal R^* \right\}.
\end{equation}
Throughout the rest of this section, we assume that $X$ is a manifold satisfying the conditions in Section~\ref{sec:preliminaries}. 

\begin{theorem}\label{thm:logistic-besov-approximation}
    Suppose that $0<\eta<1$ holds $P_X$-almost surely and the Bayes decision function
    \[
        f^*=\log\frac{\eta}{1-\eta}
    \]
    has a measurable extension in $B^\alpha_{2,\infty}(X)$ for some $\alpha>0$. The extension is again denoted by $f^*$. Suppose also that $P_X$ has a density $g\in L_p(X)$ with respect to $dV$ for a fixed $2\leq p\leq\infty$. Then, for every $\lambda>0$ and $0<\gamma\leq1$,
    \begin{equation}\label{eq:logistic-approximation-error}
        a_\gamma(\lambda) \leq C_{\alpha,p,X,d}\gamma^{2\alpha(1-1/p)}\|g\|_{L_p(X)}\|f^*\|_{B^\alpha_{2,\infty}(X)}^{2(1-1/p)}+C_{\alpha,X,d}\lambda\gamma^{-\varrho}\|f^*\|_{L_2(X)}^2.
    \end{equation}
\end{theorem}

The Besov assumption in Theorem~\ref{thm:logistic-besov-approximation} can be verified through regularity and endpoint conditions on $\eta$. The small value bound in \cite[Definition 3.1]{Bos2022Convergence} controls the probability that a conditional class probability is close to zero. For binary classification, it controls both $\eta$ near zero and $\eta$ near one. This condition is combined with low-order H\"older regularity in \cite[Assumption 3.1]{Wen2025Optimal} for kernel logistic regression. Unlike the usual Tsybakov noise condition, which concerns $\eta$ near the decision level $1/2$, it controls the regions where $f^*$ can become large. Motivated by these works, we give two sufficient conditions for the Besov regularity of $f^*$, covering low orders under H\"older assumptions and arbitrary positive orders under stronger smoothness and endpoint conditions. We impose the small value condition with respect to $dV$, since our approximation space is defined relative to manifold volume.

\begin{proposition}\label{prop:logit-besov-regularity}
    Let $\alpha>0$, and suppose that the conditional probability admits a version $\eta:X\to[0,1]$. Each of the following conditions ensures that $0<\eta<1$ holds $dV$-almost everywhere and that the Bayes function $f^*:=\log(\eta/(1-\eta))$, extended arbitrarily on the remaining null set, belongs to $B^\alpha_{2,\infty}(X)$.
    \begin{enumerate}
        \item[(i)] Let $0<\alpha<1$ and $\alpha\leq\beta\leq1$. If $\eta\in C^{0,\beta}(X)$ and
        \begin{equation}\label{eq:logit-negative-moment}
            \int_X[\eta(x)(1-\eta(x))]^{-2\alpha/\beta}\,dV(x)<\infty,
        \end{equation}
        then $f^*\in B^\alpha_{2,\infty}(X)$ and
        \begin{equation}\label{eq:logit-besov-bound}
            \|f^*\|_{B^\alpha_{2,\infty}(X)}\leq C_{X,\alpha,\beta}\bigl(1+\|\eta\|_{C^{0,\beta}(X)}^{\alpha/\beta}\bigr)\left(\int_X[\eta(1-\eta)]^{-2\alpha/\beta}\,dV\right)^{1/2}.
        \end{equation}
        \item[(ii)] Let $m\geq2$ be an integer with $\alpha\leq m$. If $\eta\in C^m(X)$ and
        \begin{equation}\label{eq:logit-high-order-moment}
            \int_X[\eta(x)(1-\eta(x))]^{-m}\,dV(x)<\infty,
        \end{equation}
        then $f^*\in H^m(X)$ and
        \begin{equation}\label{eq:logit-sobolev-bound}
            \|f^*\|_{H^m(X)}
            \leq C\left(\int_X[\eta(1-\eta)]^{-m}\,dV\right)^{1/2},
        \end{equation}
        where $C$ depends only on $X$, $m$, and $\|\eta\|_{C^m(X)}$.
    \end{enumerate}
    In particular, suppose that, for some $C_0>0$ and $\kappa>0$,
    \begin{equation}\label{eq:volume-small-value}
        V\bigl\{x\in X:\min\{\eta(x),1-\eta(x)\}\leq u\bigr\}
        \leq C_0u^\kappa,\quad 0<u\leq\frac12.
    \end{equation}
    Then \eqref{eq:logit-negative-moment} follows when $\kappa>2\alpha/\beta$, and \eqref{eq:logit-high-order-moment} follows when $\kappa>m$.
\end{proposition}

It is worth noting that the small value condition in \cite[Definition 3.1]{Bos2022Convergence} is imposed under $P_X$. This does not by itself imply \eqref{eq:volume-small-value}. If one additionally assumes that the density $g \geq c>0$ almost everywhere, the corresponding bound under $P_X$ implies \eqref{eq:volume-small-value}.

\subsection{Learning rates}\label{subsec:learning-rates}

We now combine the Gaussian approximation bounds with the oracle inequalities to derive learning rates on the manifold. The remaining ingredient is a covering estimate for the Gaussian RKHS whose dependence on the bandwidth is governed by the intrinsic dimension.

\begin{proposition}\label{prop:manifold-entropy}
    Fix an integer $m\geq2$. There are constants $C_m\geq1$ and $\gamma_m\in(0,1)$, depending only on $m$ and $X$, such that $\gamma_m^{m-1}<1/4$ and, for every $0<\gamma\leq\gamma_m$,
    \begin{equation}\label{eq:manifold-covering-entropy}
        \log N(\delta,B_{\mathcal H_\gamma(X)},\|\cdot\|_\infty)
        \leq C_m\gamma^{-\varrho}\left(\log\frac2\delta\right)^{\varrho+1},
        \quad \gamma^{m-1}<\delta\leq1.
    \end{equation}
    This implies, uniformly for the entropy exponent $q\in(0,1/2]$,
    \begin{equation}\label{eq:manifold-polynomial-covering}
        \log N(\delta,B_{\mathcal H_\gamma(X)},\|\cdot\|_\infty)
        \leq C_m q^{-(\varrho+1)}\gamma^{-\varrho}\delta^{-2q},
        \quad \gamma^{m-1}<\delta\leq1.
    \end{equation}
\end{proposition}

The proof of Proposition~\ref{prop:manifold-entropy} is based on \cite[Lemma 5.4]{Yang2016Bayesian} at the fixed smoothness order $m$. The manifold improvement in \cite[Theorem A.10]{Hamm2021Adaptive} also uses \cite[Lemma 5.4]{Yang2016Bayesian}. Here we retain the strict lower limit $\delta>\gamma^{m-1}$ in that covering estimate and apply Corollary~\ref{cor:truncated-entropy-oracle} at a covering scale above this limit. The resulting additional term is of order $\lambda^{-1}\gamma^{2(m-1)}$. By choosing a sufficiently large fixed order $m$, we can absorb this term and obtain the following learning rate.

\begin{theorem}\label{thm:besov-learning-rate}
    Under the assumptions of Theorem~\ref{thm:logistic-besov-approximation}, write $\widehat f_{n,\lambda,\gamma}$ for the regularized empirical minimizer over $\mathcal{H}_\gamma(X)$. Set $\beta:=2\alpha(1-1/p)$, and define
    \[
        \gamma_n:=\left(\frac{(\log n)^{\varrho+1}}{n}\right)^{\frac{2}{2\beta+3\varrho}}, \quad\lambda_n:=\left(\frac{(\log n)^{\varrho+1}}{n}\right)^{\frac{2(\beta+\varrho)}{2\beta+3\varrho}}.
    \]
    For every fixed $t\geq1$ and all sufficiently large $n$, with probability at least $1-e^{-t}$,
    \begin{equation}\label{eq:besov-learning-rate}
        \mathcal R(\widehat f_{n,\lambda_n,\gamma_n})-\mathcal R^* \leq C \left( \frac{(\log n)^{\varrho+1}}{n} \right)^{\frac{4\alpha(1-1/p)}{4\alpha(1-1/p)+3\varrho}}.
    \end{equation}
    Here $C$ may depend on $t,\alpha,p,X$, $\|g\|_{L_p(X)}$, and $\|f^*\|_{B^\alpha_{2,\infty}(X)}$, but is independent of $n$.
\end{theorem}

\begin{remark}\label{rem:logistic-calibration}
    Recall the calibration inequality \eqref{eq:logistic-calibration-weak}. This calibration inequality can be improved when the distribution $P$ satisfies the Tsybakov noise condition \cite{Tsybakov2004Optimal}. Specifically, assume that there exist $0\leq q<\infty$ and $C_q>0$ such that
    \[
        P_X\{x\in X:|2\eta(x)-1|\leq u\}\leq C_q u^q, \quad u>0.
    \]
    For $q=\infty$, we instead assume that there exists $u_0>0$ such that $P_X\{x\in X:|2\eta(x)-1|\leq u_0\}=0$. Then, by \cite[(2.4)]{Xiang2011Logistic}, the excess misclassification risk satisfies
    \begin{equation}\label{eq:logistic-calibration}
        \mathcal R_{01}(f)-\mathcal R_{01}^*
        \leq C\bigl(\mathcal R(f)-\mathcal R^*\bigr)^{(q+1)/(q+2)}
    \end{equation}
    for every measurable $f:X\to\mathbb R$. For finite $q$, the constant $C$ depends only on $q$ and $C_q$. For $q=\infty$, the exponent $(q+1)/(q+2)$ is interpreted as $1$, and $C$ depends on $u_0$.

    Under the assumptions and with the parameter choices of Theorem~\ref{thm:besov-learning-rate}, if the above noise condition holds, then, for every fixed $t\geq1$ and all sufficiently large $n$, the calibration inequality \eqref{eq:logistic-calibration} gives, with probability at least $1-e^{-t}$,
    \[
        \mathcal R_{01}(\widehat f_{n,\lambda_n,\gamma_n})-\mathcal R_{01}^*
        \leq C\left(\frac{(\log n)^{\varrho+1}}{n}\right)^{
        \frac{4\alpha(1-1/p)(q+1)}{(4\alpha(1-1/p)+3\varrho)(q+2)}}.
    \]
\end{remark}

\begin{remark}\label{rem:learning-rate-comparison}
    Throughout this remark, we suppress logarithmic factors, write $\beta:=2\alpha(1-1/p)$, and use $q$ for the entropy exponent. As we have mentioned in Remark~\ref{rem:regularization-dependence}, in the bounded squared-loss settings of \cite{Hamm2021Adaptive, Lei2024Pairwise}, fixed-level clipping gives an envelope and a Bernstein constant independent of $\lambda$. At localized risk level $\varepsilon$, the second moment is of order $\varepsilon$, and the leading Rademacher term has the form
    \[
        \begin{aligned}
            \left(\frac{\varepsilon}{\lambda}\right)^{q/2}
            \varepsilon^{(1-q)/2}\left(\frac{A}{n}\right)^{1/2}
            &=\varepsilon^{1/2}\lambda^{-q/2}\left(\frac{A}{n}\right)^{1/2}.
        \end{aligned}
    \]
    For logistic loss, the second-moment bound instead gives
    \[
        \begin{aligned}
            \left(\frac{\varepsilon}{\lambda}\right)^{q/2}
            \left(\varepsilon^{3/2}\lambda^{-1/2}\right)^{(1-q)/2}
            \left(\frac{A}{n}\right)^{1/2}
            &=\varepsilon^{(3-q)/4}\lambda^{-(1+q)/4}
            \left(\frac{A}{n}\right)^{1/2}.
        \end{aligned}
    \]
    Requiring each term to be bounded by a sufficiently small $\varepsilon$ gives $\varepsilon \geq C\lambda^{-q}A/n$ for squared loss, and $\varepsilon \geq C\lambda^{-1}(A/n)^{2/(1+q)}$ for logistic loss.

    This distinction matters when the squared RKHS norm of an ambient-space Gaussian approximant is bounded by $C\gamma^{-d}$, where $\gamma$ is the Gaussian bandwidth. For any fixed $\kappa>0$ and sufficiently large $n$, consider
    \[
        \begin{aligned}
            \lambda_n&=n^{-\kappa}, & q_n&=\frac1{\log n},\\
            \lambda_n^{-q_n}&=e^\kappa, & \lambda_n^{-1}&=n^\kappa.
        \end{aligned}
    \]
    In the squared-loss bounds of \cite{Hamm2021Adaptive,Lei2024Pairwise}, the remaining dependence on $q_n$ contributes only logarithmic factors. One can therefore choose $\kappa$ large enough to control the norm penalty $\lambda\gamma^{-d}$ of an ambient-space approximant without introducing $d$ into the polynomial exponent of the stochastic term. With intrinsic-dimensional covering bounds, balancing the squared-loss risk bias $\gamma^{2\alpha}$ against the stochastic term $\gamma^{-\varrho}/n$ gives the rate $n^{-2\alpha/(2\alpha+\varrho)}$, whose exponent matches the minimax rate in \cite{Stone1982Optimal}.

    However, for logistic loss, Theorem~\ref{thm:logistic-besov-approximation} gives a risk bias of order $\gamma^\beta$. The Gaussian covering estimate makes the stochastic contribution of polynomial order $\lambda^{-1}\gamma^{-2\varrho}/n^2$. With the same risk bias, an ambient-space squared RKHS norm bound of order $\gamma^{-d}$ and the intrinsic bound of Theorem~\ref{thm:rkhs-norm} lead to the respective balances
    \[
        \inf_{\lambda>0}\left\{\lambda\gamma^{-d} +\lambda^{-1}\frac{\gamma^{-2\varrho}}{n^2}\right\} =\frac{2\gamma^{-(d+2\varrho)/2}}{n}, \quad \inf_{\lambda>0}\left\{\lambda\gamma^{-\varrho} +\lambda^{-1}\frac{\gamma^{-2\varrho}}{n^2}\right\} =\frac{2\gamma^{-3\varrho/2}}{n}.
    \]
    Balancing each term with $\gamma^\beta$ gives the ambient rate $n^{-2\beta/(2\beta+2\varrho+d)}$ and the intrinsic rate $n^{-2\beta/(2\beta+3\varrho)}$ in Theorem~\ref{thm:besov-learning-rate}. Thus the intrinsic norm bound removes the ambient dimension from the polynomial exponent. This comparison explains the role of the intrinsic RKHS norm bound within the present oracle analysis.

    The resulting rate remains slower than the squared-loss regression rate $n^{-2\alpha/(2\alpha+\varrho)}$ in \cite{Hamm2021Adaptive}, whose dependence on smoothness and dimension matches the minimax rate in \cite{Stone1982Optimal}. Two differences in the analysis explain this gap. The first is, as discussed above, the norm-dependent variance bound in Lemma~\ref{lem:logistic-variance}. The second is the density-dependent risk bias in Theorem~\ref{thm:logistic-besov-approximation}, which is of order $\gamma^{2\alpha(1-1/p)}$ for $g\in L_p(X)$.
\end{remark}

In view of Remark~\ref{rem:learning-rate-comparison}, we now seek to close the gap between the logistic learning rate and the squared-loss regression rate by clipping the predictions at a level that increases with the sample size. The idea of clipping is to control the supremum norm and the Bernstein constant through the clipping level $M$. As we will show in the proof of Theorem~\ref{thm:clipped-learning-rate}, clipping only increases the logistic risk by at most $e^{-M}$. Therefore, a clipping level that grows logarithmically with $n$ can make the clipping bias negligible. The following theorem gives a faster rate for the clipped predictions under the same Besov and density assumptions, with a new choice of bandwidth and regularization parameter. For bounded densities, it recovers the optimal polynomial exponent of the regression rate $n^{-2\alpha/(2\alpha+\varrho)}$. Since clipping preserves the classification rule, it also improves the misclassification bound for the original estimator.

\begin{theorem}\label{thm:clipped-learning-rate}
    Under the assumptions of Theorem~\ref{thm:logistic-besov-approximation}, define the clipping operator
    \[
        \pi_M(u)=\max\{-M,\min\{u,M\}\},\quad M>0.
    \]
    Set $\beta:=2\alpha(1-1/p)$, and define
    \[
        \gamma_n:=\left(\frac{(\log n)^{\varrho+2}}{n}\right)^{1/(\beta+\varrho)}, \quad \lambda_n:=\frac{(\log n)^{\varrho+2}}{n}, \quad M_n:=3\log n.
    \]
    For every fixed $t\geq1$ and all sufficiently large $n$, with probability at least $1-e^{-t}$,
    \begin{equation}\label{eq:clipped-learning-rate}
        \begin{aligned}
            \mathcal R(\pi_{M_n}\widehat f_{n,\lambda_n,\gamma_n})-\mathcal R^*
            &\leq C\left(\frac{(\log n)^{\varrho+2}}{n}\right)^{\beta/(\beta+\varrho)},\\
            \mathcal R_{01}(\widehat f_{n,\lambda_n,\gamma_n})-\mathcal R_{01}^*
            &\leq C\left(\frac{(\log n)^{\varrho+2}}{n}\right)^{\beta/(2(\beta+\varrho))}.
        \end{aligned}
    \end{equation}
    Here $C$ may depend on $t,\alpha,p,X$, $\|g\|_{L_p(X)}$, and $\|f^*\|_{B^\alpha_{2,\infty}(X)}$, but is independent of $n$.
\end{theorem}

For $p=\infty$, the misclassification bound in \eqref{eq:clipped-learning-rate} has polynomial order $n^{-\alpha/(2\alpha+\varrho)}$. In the classical setting of binary classification with an $\alpha$-H\"older conditional class probability on a $d$-dimensional Euclidean domain, under standard density assumptions and without an additional margin condition, the minimax rate of expected excess misclassification risk is $n^{-\alpha/(2\alpha+d)}$. See \cite[(1.5)]{Audibert2007Fast}. Thus, up to logarithmic factors, our bound also matches the minimax optimal rate for binary classification, with the ambient dimension $d$ replaced by the intrinsic dimension $\varrho$.

\section{Application to elliptic solution operators from random samples}\label{sec:elliptic-operator-learning}

We approximate the solution operator of an elliptic equation from random points on the manifold and source values at those points. Theorem~\ref{thm:gaussian-sobolev-expansion} gives a high-order localized approximation of the Laplace--Beltrami operator. Theorems~\ref{thm:rkhs-norm} and \ref{thm:besov-approximation} then control the approximation error of Gaussian trial spaces in $H^1(X)$. On the manifold $X$ satisfying the conditions in Section~\ref{sec:preliminaries}, consider the elliptic problem
\begin{equation}\label{eq:pde-elliptic-problem}
    (\kappa I-\Delta_X)u = f,
\end{equation}
where $\kappa>0$. For a given source $f\in L_2(X)$, the weak formulation of \eqref{eq:pde-elliptic-problem} is to find $u\in H^1(X)$ such that
\[
    a_\kappa(u,v)=\langle f,v\rangle_{L_2(X)},
    \quad \forall v\in H^1(X),
\]
where the bilinear form $a_\kappa$ is defined by
\[
    a_\kappa(w,v):=\kappa\langle w,v\rangle_{L_2(X)}
    +\int_X\langle\nabla w,\nabla v\rangle\,dV,
    \quad w,v\in H^1(X).
\]
Let $c_\kappa:=\min\{\kappa,1\}$ and $C_\kappa:=\max\{\kappa,1\}$. Then the bilinear form $a_\kappa$ is coercive and bounded on $H^1(X)$:
\begin{equation}\label{eq:pde-coercivity}
    \begin{aligned}
        c_\kappa\|u\|_{H^1(X)}^2
        &\leq a_\kappa(u,u),\\
        |a_\kappa(u,v)|
        &\leq C_\kappa\|u\|_{H^1(X)}\|v\|_{H^1(X)}.
    \end{aligned}
\end{equation}
Since $|\langle f,v\rangle_{L_2(X)}|\leq\|f\|_{L_2(X)}\|v\|_{H^1(X)}$, the Lax--Milgram theorem gives a unique weak solution for each $f\in L_2(X)$. We denote the resulting bounded linear solution operator by $G:L_2(X)\to H^1(X)$.
We aim to approximate $G$ from randomly sampled points on $X$. This requires that point evaluations of the sources are well-defined. Therefore, we fix an integer $s>\varrho/2$ and assume $f \in H^s(X) \hookrightarrow C(X)$. Let $q$ be a known density with respect to $dV$  satisfying $0 < q_0\leq q(x)<\infty$ for every $x\in X$. We observe $2n$ i.i.d. sample points from $q\,dV$:
\[
    \{X_1,\ldots,X_n,Y_1,\ldots,Y_n\}.
\]
For each new source $f\in H^s(X)$, we also observe its exact values $f(X_i)$. The points $Y_i$ are paired with $X_i$ to approximate the double integrals in the Gaussian approximation and to construct the stiffness matrix. Our goal is to construct a random linear approximation $\widehat G:H^s(X)\to H^1(X)$ that maps these source values to an approximate solution. The uniform approximation error over the unit ball of $H^s(X)$ is defined as
\[
    \|\widehat G-G\|_{H^s(X)\to H^1(X)}
    :=\sup_{\|f\|_{H^s(X)}\leq1}\|\widehat Gf-Gf\|_{H^1(X)}.
\]
For Gaussian trial spaces, we consider sources $f\in B^\alpha_{2,\infty}(X)$ with $\alpha>\varrho/2$, so the embedding $B^\alpha_{2,\infty}(X)\hookrightarrow C(X)$ ensures that the observations $f(X_i)$ are well defined. The corresponding operator norm is
\[
    \|\widehat G-G\|_{B^\alpha_{2,\infty}(X)\to H^1(X)}
    :=\sup_{\|f\|_{B^\alpha_{2,\infty}(X)}\leq1}\|\widehat Gf-Gf\|_{H^1(X)}.
\]

\subsection{High-order Gaussian approximation of the Laplacian}

We first construct a high-order Gaussian approximation of $\Delta_X$, then use its symmetric form to define $\widehat G$. We remove the zeroth-order coefficient of $T_\gamma$ by centering
\[
    C_\gamma u:=T_\gamma u-uT_\gamma\mathbf1
    =\int_X \left(\frac{1}{\gamma\sqrt{\pi}}\right)^\varrho
    \exp\left(-\frac{\|\cdot - y\|_{\mathbb R^d}^2}{\gamma^2}\right)\bigl(u(y)-u(\cdot)\bigr)\,dV(y).
\]
Fix an integer $N\geq1$. By solving a Vandermonde linear system, we can always choose $a_1,\ldots,a_N$ such that
\begin{equation}\label{eq:pde-laplacian-moments}
    \sum_{j=1}^N a_jj^{2r}
    =\begin{cases}4,&r=1,\\0,&2\leq r\leq N.\end{cases}
\end{equation}
Define
\[
    \begin{aligned}
        L_{\gamma,N}&:=\gamma^{-2}\sum_{j=1}^N a_j C_{j\gamma}, \\
        J_{\gamma,N}(x,y)&:=\gamma^{-2}\sum_{j=1}^N a_j\left(\frac{1}{j\gamma\sqrt{\pi}}\right)^\varrho \exp\left(-\frac{\|x - y\|_{\mathbb R^d}^2}{(j\gamma)^2}\right).
    \end{aligned}
\]

The following lemma shows that $L_{\gamma,N}$ is a high-order approximation of the Laplacian $\Delta_X$ and that it has a symmetric form with kernel $J_{\gamma,N}$.
\begin{lemma}\label{lem:pde-centered-consistency}
    For every $N\geq1$, there exists a constant $C_{N,X,d} > 0$ such that, for $0<\gamma\leq1/N$ and $u, v \in C^\infty(X)$,
    \begin{equation}\label{eq:pde-centered-bias}
        \|L_{\gamma,N}u-\Delta_Xu\|_{L_2(X)}
        \leq C_{N,X,d}\gamma^{2N}
        \bigl(\|u\|_{H^{2N+2}(X)}+\|u\|_\infty\bigr),
    \end{equation}
    \begin{equation}\label{eq:pde-gaussian-energy}
        -\langle L_{\gamma,N}u,v\rangle_{L_2(X)}
        =\frac12\int_X\int_X J_{\gamma,N}(x,y)
        \bigl(u(x)-u(y)\bigr)\bigl(v(x)-v(y)\bigr)\,dV(x)dV(y).
    \end{equation}
\end{lemma}

The right-hand side of \eqref{eq:pde-gaussian-energy} is symmetric in $u$ and $v$, so $L_{\gamma,N}$ is symmetric in $L_2(X)$. For $N=1$, the kernel is nonnegative, and taking $v=u$ gives $\langle L_{\gamma,1}u,u\rangle_{L_2(X)}\leq0$. For $N>1$, the kernel $J_{\gamma,N}$ may change sign, so the sign of the quadratic form is no longer evident from the integrand. On a fixed finite-dimensional trial space, Theorem~\ref{thm:pde-random-solution-operator} combines the consistency estimate \eqref{eq:pde-centered-bias} with a sampling bound to control the perturbation of the Galerkin matrix for $a_\kappa$ with high probability. The empirical matrix remains positive definite whenever the operator norm of this perturbation is smaller than the smallest eigenvalue of the exact matrix.

\subsection{High-order Gaussian approximation of the operator from sample points}

Choose a fixed trial space
\[
    W=\operatorname{span}\{\varphi_1,\ldots,\varphi_J\}\subset C^\infty(X),
    \quad \Phi(x)=(\varphi_1(x),\ldots,\varphi_J(x))^{\top},
\]
where the basis functions are linearly independent on $X$ and chosen independently of the sampled points. An example is a finite collection of Gaussian radial functions with a fixed bandwidth and distinct centers on $X$ and we will consider it in Section~\ref{subsec:pde-besov-trial-spaces}. The deterministic trial space error is defined as
\[
    \eta_W:=\sup_{\|f\|_{H^s(X)}\leq1}\inf_{v\in W}\|Gf-v\|_{H^1(X)}.
\]

Using sample points $\{X_1,\ldots,X_n,Y_1,\ldots,Y_n\}$, define
\begin{equation}\label{eq:pde-empirical-mass-load}
    \widehat{M}
    :=\frac1n\sum_{i=1}^n\frac{\Phi(X_i)\Phi(X_i)^{\top}}{q(X_i)},
    \quad
    \widehat b(f)
    :=\frac1n\sum_{i=1}^n\frac{\Phi(X_i)f(X_i)}{q(X_i)},
\end{equation}
the symmetric stiffness matrix
\begin{equation}\label{eq:pde-empirical-stiffness}
    \widehat{K}
    :=\frac1{2n}\sum_{i=1}^n
    \frac{J_{\gamma,N}(X_i,Y_i)}{q(X_i)q(Y_i)}
    \bigl(\Phi(X_i)-\Phi(Y_i)\bigr)
    \bigl(\Phi(X_i)-\Phi(Y_i)\bigr)^{\top}, \quad \widehat{B}:=\kappa\widehat{M}+\widehat{K},
\end{equation}
and the random operator
\[
    \widehat G_{n,\gamma,W}f
    :=\begin{cases}
        \displaystyle\sum_{j=1}^J
        \bigl(\widehat{B}^{-1}\widehat b(f)\bigr)_j\varphi_j,
        &\widehat{B}\text{ is positive definite},\\
        0,&\text{otherwise}.
    \end{cases}
\]
The following result proves that $\widehat{B}$ is positive definite with high probability and gives a uniform bound on the approximation error on the same event.
\begin{theorem}\label{thm:pde-random-solution-operator}
    Fix an integer $s>\varrho/2$, an integer $N\geq1$, and the trial space $W$. There are constants $C,c>0$, depending on $s$, $N$, $W$, $X$, $d$, $\kappa$, and $q_0$, such that the following holds. For $0<\delta<1$ and $0<\gamma\leq1/N$, set
    \[
        \varepsilon_{n,\gamma,\delta}
        :=\gamma^{2N}+\sqrt{\frac{\log(2/\delta)}{n\gamma^\varrho}}.
    \]
    If $\varepsilon_{n,\gamma,\delta}\leq c$, then, with probability at least $1-\delta$, the matrix $\widehat{B}$ is positive definite and
    \begin{equation}\label{eq:pde-random-operator-bound}
        \|\widehat G_{n,\gamma,W}-G\|_{H^s(X)\to H^1(X)}\leq\frac{C_\kappa}{c_\kappa}\eta_W+C\varepsilon_{n,\gamma,\delta}.
    \end{equation}
\end{theorem}

\begin{corollary}\label{cor:pde-operator-rate}
    Under the assumptions of Theorem~\ref{thm:pde-random-solution-operator}, set $\gamma=\bigl(\log(2/\delta)/n\bigr)^{1/(4N+\varrho)}$.
    For all sufficiently large $n/\log(2/\delta)$, with probability at least $1-\delta$,
    \begin{equation}\label{eq:pde-operator-rate}
        \|\widehat G_{n,\gamma,W}-G\|_{H^s(X)\to H^1(X)}
        \leq\frac{C_\kappa}{c_\kappa}\eta_W
        +C\left(\frac{\log(2/\delta)}{n}\right)^{\frac{2N}{4N+\varrho}}.
    \end{equation}
    On the same event, every probability distribution $\mu$ on $H^s(X)$ with a finite second moment satisfies
    \begin{equation}\label{eq:pde-new-source-risk}
        \int_{H^s(X)}\|\widehat G_{n,\gamma,W}f-Gf\|_{H^1(X)}^2\,d\mu(f)
        \leq\left(\frac{C_\kappa}{c_\kappa}\eta_W
        +C\left(\frac{\log(2/\delta)}{n}\right)^{\frac{2N}{4N+\varrho}}\right)^2
        \int_{H^s(X)}\|f\|_{H^s(X)}^2\,d\mu(f).
    \end{equation}
\end{corollary}

Theorem~\ref{thm:pde-random-solution-operator} separates the trial-space error, the high-order Gaussian bias, and the sampling error. For fixed $W$, the bandwidth in Corollary~\ref{cor:pde-operator-rate} balances the bias and the sampling error. 

\subsection{Gaussian trial spaces for Besov sources}\label{subsec:pde-besov-trial-spaces}

We now bound the trial-space term under Gaussian trial spaces. To this end, we use $\sigma$ for the bandwidth of the trial functions and retain $\gamma$ for the bandwidth in $L_{\gamma,N}$.

We assemble the trial space by choosing a finite set of centers and taking the span of the corresponding Gaussian kernels. Let $\Xi\subset X$ be a finite set of distinct centers chosen independently of the sample points. Define the Gaussian trial space and its fill distance by
\[
    W_{\sigma,\Xi}:=\operatorname{span}\{k_{\sigma}(\cdot,\xi):\xi\in\Xi\}, \quad h:=\sup_{x\in X}\min_{\xi\in\Xi}d_X(x,\xi).
\]
For $\alpha>\varrho/2$, the embeddings in Section~\ref{sec:function-spaces-on-manifolds} ensure that the point observations in \eqref{eq:pde-empirical-mass-load} are well defined for sources in $B^\alpha_{2,\infty}(X)$.

\begin{corollary}\label{cor:pde-besov-trial-bound}
    Fix $\alpha>\varrho/2$ and an integer $s>\max\{1,\varrho/2\}$. If $h$ is sufficiently small, then for $0<\sigma\leq1$,
    \begin{equation}\label{eq:pde-besov-trial-bound}
        \sup_{\|f\|_{B^\alpha_{2,\infty}(X)}\leq1}
        \inf_{v\in W_{\sigma,\Xi}}\|Gf-v\|_{H^1(X)}
        \leq C\left(\sigma^{\alpha+1}+h^{s-1}\sigma^{-s-\varrho/2}\right),
    \end{equation}
    where the constant $C$ is independent of $\sigma$ and $\Xi$. Fix the trial space $W=W_{\sigma,\Xi}$ and $N\geq1$. There are constants $c_W,C_W>0$ such that, if $0<\gamma\leq1/N$, $0<\delta<1$, and $\varepsilon_{n,\gamma,\delta}\leq c_W$, then with probability at least $1-\delta$, $\widehat B$ is positive definite and
    \begin{equation}\label{eq:pde-besov-operator-bound}
        \|\widehat G_{n,\gamma,W}-G\|_{B^\alpha_{2,\infty}(X)\to H^1(X)} \leq C\left(\sigma^{\alpha+1}+h^{s-1}\sigma^{-s-\varrho/2}\right) +C_W\varepsilon_{n,\gamma,\delta}.
    \end{equation}
\end{corollary}

\begin{remark}\label{rem:pde-quasi-uniform-centers}
    Taking $\sigma=h^{(s-1)/(s+\alpha+1+\varrho/2)}$ balances the two trial-space terms in \eqref{eq:pde-besov-operator-bound}. For quasi-uniform centers with $J=|\Xi|$, the fill distance is of order $J^{-1/\varrho}$ by volume comparison. With this choice of $\sigma$ and the choice of $\gamma$ in Corollary~\ref{cor:pde-operator-rate}, we obtain
    \begin{equation}\label{eq:pde-besov-fixed-resolution-rate}
        \|\widehat G_{n,\gamma,W}-G\|_{B^\alpha_{2,\infty}(X)\to H^1(X)}
        \leq C J^{-\frac{(\alpha+1)(s-1)}{\varrho(s+\alpha+1+\varrho/2)}}
             + C_W'\left(\frac{\log(2/\delta)}{n}\right)^{\frac{2N}{4N+\varrho}}
    \end{equation}
    with probability at least $1-\delta$ for all sufficiently large $n/\log(2/\delta)$ at fixed $W$. The first term quantifies the trial-space resolution. Its exponent approaches $(\alpha+1)/\varrho$ as $s$ increases, but at fixed $h<1$, a larger $s$ reduces the bandwidth $\sigma$. The resulting narrower basis functions can be harder to resolve with a fixed set of sample points when assembling the empirical Galerkin matrices. The constant $C$ in the trial-space term may also grow with $s$. The second term bounds the combined Gaussian bias and sampling error. Its exponent approaches $1/2$ as $N$ increases, but a larger $N$ requires more Gaussian kernel evaluations per sample pair when assembling the stiffness matrix.

    The sampling rate in \eqref{eq:pde-besov-fixed-resolution-rate} holds for a fixed Gaussian trial space $W$. To obtain a rate when the number of centers $J$ grows with the sample size $n$, one would need an upper bound for $C_W'$ and a lower bound for the stability threshold $c_W$ in terms of $J$. We leave this as an open problem for future work.
\end{remark}

\section{Conclusions}\label{sec:conclusions}

In this paper, we use signed combinations of Gaussian integral operators at different bandwidths to yield high-order approximation of intrinsic Besov functions while preserving an RKHS norm bound governed by the manifold dimension. The Sobolev expansion identifies the differential bias terms that can be removed by this combination. Uniform stability extends the approximation estimate to the Sobolev scale, with rate $\gamma^{\alpha-r}$ for a Besov target of smoothness $\alpha$ measured in $H^r(X)$. Potential directions for future work include considering more general RKHSs, such as those with rapidly decaying or compactly supported kernels, and extending the analysis to manifolds with boundary or lower regularity.

For regularized logistic classification, the $L_2(X)$ approximation bound and the RKHS norm bound, together with a localized oracle inequality, yield rates for the excess logistic risk that depend on the intrinsic dimension of $X$ even when the Bayes function is unbounded. Open questions include how to select the bandwidth and regularization parameter adaptively from the data and whether the dependence of these rates on the integrability of the marginal density is optimal. For elliptic operator learning, the Gaussian expansion controls the bias in the stiffness matrix assembled from random point pairs. Combining this estimate with the sampling bound and the $H^1(X)$ approximation error of the Gaussian trial space gives a high-probability bound on the $H^1(X)$ solution error, uniform over all source functions in the Besov unit ball. Open questions include quantifying a joint rate in the number of Gaussian centers and the sample size and extending these results to more general types of solution operators.

\begin{appendix}
\section{Proofs}\label{sec:proofs}

\subsection{Proof of Proposition~\ref{prop:holder-besov-embedding}}\label{subsec:proofs-preliminaries}

\begin{proof}

If $0<\beta<1$, the Euclidean Hölder--Zygmund theorem \cite[Theorem 1.2.2]{Triebel1992Theory}, applied in a finite atlas after multiplication by a subordinate partition of unity whose supports are compactly contained in the corresponding charts, gives $C^{k,\beta}(X)=B^\alpha_{\infty,\infty}(X)$ with equivalent norms. If $\beta=1$, the same argument gives the continuous inclusion $C^{k,1}(X)\hookrightarrow B^{k+1}_{\infty,\infty}(X)$.

It remains to pass from $L_\infty(X)$ to $L_2(X)$. Choose $0<r<\operatorname{inj}(X)$, $\varepsilon>0$ sufficiently small, and take the local-means order $N\in\mathbb{N}$ with $N>\max\{\alpha,\varrho+6\}$. Then \cite[Theorem 7.3.2]{Triebel1992Theory} applies, with the same local means $k_0(\varepsilon,f)$ and $k(t,f)$, both for $(p,q)=(\infty,\infty)$ and for $(p,q)=(2,\infty)$. Compactness gives
\[
    \|u\|_{L_2(X)} \leq V(X)^{1/2}\|u\|_{L_\infty(X)}.
\]
Applying this inequality to every local mean in the two equivalent quasi-norms yields
\[
    \|f\|_{B^\alpha_{2,\infty}(X)} \leq C_{\alpha,X}\|f\|_{B^\alpha_{\infty,\infty}(X)} \leq C\|f\|_{C^{k,\beta}(X)},
\]
which completes the proof.

\end{proof}

\subsection{Proofs for Section~\ref{sec:manifold-approximation}}

\begin{proof}[Proof of Theorem~\ref{thm:rkhs-norm}]
    The compact embedded manifold $X$  satisfies the uniform extrinsic volume-growth bound \cite{Karp1989Volume}
    \[
        \sup_{y\in X}V\bigl(X\cap B_{\mathbb R^d}(y,r)\bigr) \leq C_Xr^\varrho,\quad 0<r\leq1.
    \]
    This gives the uniform integral bound
    \[
        \sup_{y\in X} \int_X \exp\left(-\frac{\|x-y\|^2_{\mathbb{R}^d}}{\gamma^2}\right)dV(x) \leq C_X\gamma^\varrho.
    \]
    By the reproducing property,
    \[
        \begin{aligned}
            \|T_\gamma f\|_{\mathcal{H}_\gamma(X)}^2 &=(\gamma\sqrt{\pi})^{-2\varrho}\int_X\int_X f(x)f(y)\langle k_\gamma(\cdot,x),k_\gamma(\cdot,y)\rangle_{\mathcal{H}_\gamma(X)}
            \,dV(x)dV(y)\\
            \\
            &= (\gamma\sqrt{\pi})^{-2\varrho} \int_X\int_X \exp\left(-\frac{\|x-y\|^2_{\mathbb{R}^d}}{\gamma^2}\right)f(x)f(y)\,dV(x)dV(y)\\
            &\leq \frac{1}{2(\gamma\sqrt{\pi})^{2\varrho}}\int_X\int_X \left(\vert f(x)\vert^2+\vert f(y)\vert^2\right)\exp\left(-\frac{\|x-y\|^2_{\mathbb{R}^d}}{\gamma^2}\right)\,dV(x)dV(y)\\
            &\leq C_X\gamma^{-\varrho}\|f\|_{L_2(X)}^2.
        \end{aligned}
    \]
    The Gaussian RKHS inclusion criterion \cite[Proposition 4.46]{Steinwart2008Support} gives
    \[
        \mathcal{H}_{j\gamma}(X)\subset \mathcal{H}_\gamma(X), \quad \|u\|_{\mathcal{H}_\gamma(X)} \leq j^{d/2}\|u\|_{\mathcal{H}_{j\gamma}(X)}, \quad j \geq 1.
    \]
    Applying these two estimates to every term in \eqref{eq:multiscale-operator} proves the result.
\end{proof}

\begin{proof}[Proof of Theorem~\ref{thm:gaussian-sobolev-expansion}]
    We first prove the estimate for $f\in C^\infty(X)$ and then extend it by density. Write $L=2N+2$. Choose $0<\delta<\operatorname{inj}(X)/4$ and $\chi_0\in C_c^\infty([0,\infty))$ with $0\leq\chi_0\leq1$ such that $\chi_0=1$ on $[0,\delta^2]$ and $\chi_0=0$ on $[4\delta^2,\infty)$. Define
    \[
        \begin{aligned}
            \chi(x,y) &:=\chi_0\bigl(d_X(x,y)^2\bigr),\\
            T_\gamma &=T_\gamma^{\mathrm{loc}}+T_\gamma^{\mathrm{far}}. \\
            T_\gamma^{\mathrm{loc}}f(x) &:=(\gamma\sqrt{\pi})^{-\varrho}\int_X \exp\left(-\frac{\|x-y\|_{\mathbb R^d}^2}{\gamma^2}\right)\chi(x,y)f(y)\,dV(y),\\
            T_\gamma^{\mathrm{far}}f(x)&:=(\gamma\sqrt{\pi})^{-\varrho}\int_X \exp\left(-\frac{\|x-y\|_{\mathbb R^d}^2}{\gamma^2}\right)\bigl(1-\chi(x,y)\bigr)f(y)\,dV(y).
        \end{aligned}
    \]

    Since $X$ is compact and its smooth embedding in $\mathbb R^d$ induces the Riemannian metric, there is a $c_\delta>0$ such that
    \[
        d_X(x,y)\geq\delta \Longrightarrow\|x-y\|_{\mathbb{R}^d}\geq c_\delta.
    \]
    Denote the integral kernel of $T_\gamma^{\mathrm{far}}$ by
    \[
        K_\gamma^{\mathrm{far}}(x,y):=(\gamma\sqrt{\pi})^{-\varrho}
        k_\gamma(x,y)\bigl(1-\chi(x,y)\bigr).
    \]
    Since $1-\chi(x,y)$ vanishes for $d_X(x,y)\leq\delta$ and $0\leq\chi\leq1$, we obtain
    \[
        \sup_{x\in X}\int_X |K_\gamma^{\mathrm{far}}(x,y)|\,dV(y)
        \leq (\gamma\sqrt{\pi})^{-\varrho}V(X)\exp(-c_\delta^2/\gamma^2).
    \]
    By symmetry, the same bound holds with $x$ and $y$ interchanged. The Schur test therefore gives
    \begin{equation}\label{eq:far-gaussian-tail}
        \|T_\gamma^{\mathrm{far}}\|_{L_2(X)\to L_2(X)} \leq C_{X,\delta}\gamma^{-\varrho}\exp(-c_\delta^2/\gamma^2).
    \end{equation}

    For the local term, let $\mathcal{E}_x: T_xX \to X$ denote the exponential map at $x$. Since $\chi(x,y)$ vanishes for $d_X(x,y) \geq 2 \delta$ and $\delta < \operatorname{inj}(X)/4$, we can put $y=\mathcal{E}_x(u)$, and write
    \[
        q_x(u):=\|x-\mathcal{E}_x(u)\|^2_{\mathbb{R}^d}, \quad dV(\mathcal{E}_x(u))=J_x(u)\,du_x.
    \]
    By \cite[Lemma D.12]{Tang2026Adaptive}, uniformly in $x$ and for $\|u\|<2\delta$,
    \begin{equation}\label{eq:normal-coordinate-estimates}
        \begin{aligned}
            q_x(u)&=\|u\|^2+O(\|u\|^4),\\
            J_x(u)&=1+O(\|u\|^2),\\
            c\|u\|^2&\leq q_x(u)\leq C\|u\|^2.
        \end{aligned}
    \end{equation}
    Rescale the tangent coordinates by the bandwidth via $u=\gamma z$. We then define
    \[
        \Psi_x(t,z) :=
        \begin{cases}
            q_x(tz)/t^2,&t\neq0, \|tz\|<3\delta,\\
            \|z\|^2,&t=0,
        \end{cases}
    \]
    and extend the product
    \[
        \exp(-\Psi_x(t,z))J_x(tz)\chi_0(t^2\|z\|^2)
    \]
    by zero when $\|tz\| \geq 3\delta$. Since $q_x$ and its first $u$-derivative vanish at the origin, applying Taylor's formula to the scalar function $\theta\mapsto q_x(\theta tz)$ and dividing by $t^2$ for $t\neq0$ gives
    \begin{equation}\label{eq:psi-integral-representation}
        \begin{aligned}
            \Psi_x(t,z)&=\int_0^1(1-\theta)D_u^2q_x(\theta tz)[z,z]\,d\theta.
        \end{aligned}
    \end{equation}
    For fixed $x,z$, the right-hand side is smooth in $t$ near zero and takes the value $\|z\|^2$ there, so this representation also holds at zero.
    The cutoff restricts the nonzero terms to $\|tz\|<2\delta$. On this region, $\Psi_x(t,z)\geq c\|z\|^2$, and its integral representation bounds each $t$-derivative by a polynomial in $\|z\|$. The same holds for the derivatives of $J_x(tz)$ and the cutoff. The chain rule and Leibniz's rule therefore give, for $\vert t \vert \leq1$ and $0 \leq \ell \leq L$,
    \begin{equation}\label{eq:amplitude-domination}
        \left\vert \partial_t^\ell \left[ \exp(-\Psi_x(t,z))J_x(tz)\chi_0(t^2\|z\|^2) \right] \right\vert \leq C_\ell(1+\|z\|^2)^{M_\ell/2}\exp(-c\|z\|^2).
    \end{equation}
    Here, $C_\ell>0$, $M_\ell\in\mathbb N_0$, and $c>0$ are independent of $x,t,z$. Thus each derivative is integrable over $T_xX$. Let $\mathcal{I}_t f(x)$ denote the rescaled integral
    \[
        \mathcal{I}_t f(x) = \pi^{-\varrho/2} \int_{T_xX} \exp(-\Psi_x(t,z))J_x(tz)\chi_0(t^2\|z\|^2) f(\mathcal{E}_x(tz))\,dz_x.
    \]
    Then $\mathcal{I}_t=T_{|t|}^{\mathrm{loc}}$ for $t\neq0$ and $\mathcal{I}_0=I$.

    We now apply Taylor's formula at the origin to the integrand as a function of $t$, retaining terms up to order $L-1$. For fixed $x$ and $z$, write the full integrand as the scalar function
    \[
        \begin{aligned}
            F_{x,z}(s) &= \exp(-\Psi_x(s,z))J_x(sz)\chi_0(s^2\|z\|^2)f(\mathcal E_x(sz)).
        \end{aligned}
    \]
    Let $F_{x,z}^{(j)}$ denote differentiation with respect to $s$. For the factor $f(\mathcal E_x(tz))$, differentiation along the radial geodesic gives, for $1\leq\ell\leq L$ and $\|tz\|<2\delta$,
    \[
        \frac{d^\ell}{dt^\ell}f(\mathcal E_x(tz)) =(\nabla^\ell f)_{\mathcal E_x(tz)}\bigl(\underbrace{P_{x,\mathcal E_x(tz)}z,\ldots,P_{x,\mathcal E_x(tz)}z}_{\ell\text{ arguments}}\bigr).
    \]
    Since parallel transport preserves lengths, this yields
    \begin{equation}\label{eq:f-derivative-domination}
        \left|\frac{d^\ell}{dt^\ell}f(\mathcal E_x(tz))\right|
        \leq \|z\|^\ell\vert\nabla^\ell f\vert(\mathcal E_x(tz)).
    \end{equation}
    By Leibniz's rule, each term in $F_{x,z}^{(L)}(s)$ has $\ell$ derivatives on $f(\mathcal E_x(sz))$ and $L-\ell$ derivatives on the other factor. Combining \eqref{eq:f-derivative-domination} with \eqref{eq:amplitude-domination}, and using $\|z\|^\ell\leq(1+\|z\|^2)^{\ell/2}$, gives
    \begin{equation}\label{eq:F-derivative-bound}
        \begin{aligned}
            \vert F_{x,z}^{(L)}(s)\vert &\leq C(1+\|z\|^2)^{M/2}\exp(-c\|z\|^2) \mathbf 1_{\{\|sz\|\leq2\delta\}} \sum_{\ell=0}^{L}|\nabla^\ell f|(\mathcal E_x(sz)), \\
            M &:= \max_{0\leq\ell\leq L}(M_{L-\ell}+\ell).
        \end{aligned}
    \end{equation}
    Applying Taylor's formula to $F_{x,z}$ at zero and then integrating over $z$ gives the following expansion, with remainder operator $\mathcal R_{L,t}$:
    \[
        \begin{aligned}
            \mathcal I_t f(x) &= \pi^{-\varrho/2}\sum_{j=0}^{L-1}\frac{t^j}{j!} \int_{T_xX}F_{x,z}^{(j)}(0)\,dz_x +\mathcal R_{L,t}f(x),\\
            \mathcal R_{L,t}f(x) &:= \frac{\pi^{-\varrho/2}t^L}{(L-1)!} \int_0^1(1-\theta)^{L-1} \int_{T_xX}F_{x,z}^{(L)}(\theta t)\,dz_x\,d\theta.
        \end{aligned}
    \]
    Applying \eqref{eq:F-derivative-bound}, we obtain
    \begin{equation}\label{eq:remainder-bound}
        \begin{aligned}
            \left\vert \mathcal R_{L,t}f(x)\right\vert
            &\leq C\vert t \vert^L \sum_{\ell=0}^{L} \int_0^1(1-\theta)^{L-1}
            \int_{T_xX}(1+\|z\|^2)^{M/2}\exp(-c\|z\|^2)\\
            &\qquad\times\mathbf 1_{\{\|\theta tz\|\leq2\delta\}}
            \vert \nabla^\ell f\vert(\mathcal E_x(\theta tz))\,dz_x\,d\theta.
        \end{aligned}
    \end{equation}
    Choose finitely many pairs of coordinate neighborhoods $U_a\Subset V_a$ such that a partition of unity is supported in the $U_a$, each $V_a$ carries a smooth orthonormal frame, and $\varphi_a(V_a)$ is convex (for example, an open Euclidean ball). For each $a$, let $e_1,\ldots,e_\varrho$ denote the frame on $V_a$. For fixed $\theta\in[0,1]$, $\vert t \vert\leq1$, and $z\in\mathbb R^\varrho$ with $\|\theta tz\|\leq2\delta$, define
    \[
        \begin{aligned}
            \Theta_{\theta,t,z}(x) &= \mathcal{E}_x\left(\theta t\sum_{i=1}^{\varrho}z_ie_i(x)\right),\quad x\in V_a.
        \end{aligned}
    \]
    After decreasing $\delta$, we have $\Theta_{\theta,t,z}(U_a)\subset V_a$, and the coordinate representation $\varphi_a\circ\Theta_{\theta,t,z}\circ\varphi_a^{-1}$ is uniformly $C^1$-close to the identity on a convex neighborhood of $\varphi_a(\overline{U_a})$. Consequently, $\Theta_{\theta,t,z}$ is injective on $U_a$ and satisfies
    \[
        0<c\leq |\det D_x\Theta_{\theta,t,z}| \leq C, \quad x \in U_a.
    \]
    Here $|\det D_x\Theta_{\theta,t,z}|$ denotes the Jacobian with respect to Riemannian volume on the domain and image. The constants are independent of $a,x,\theta,t,z$. For every $0\leq\ell\leq L$, the lower Jacobian bound and change of variables under the injective map $\Theta_{\theta,t,z}$ give
    \[
        \begin{aligned}
            \int_{U_a}|\nabla^\ell f|^2(\Theta_{\theta,t,z}(x))\,dV(x)
            &\leq c^{-1}\int_{U_a}|\nabla^\ell f|^2(\Theta_{\theta,t,z}(x))
            |\det D_x\Theta_{\theta,t,z}|\,dV(x)\\
            &= c^{-1}\int_{\Theta_{\theta,t,z}(U_a)}|\nabla^\ell f|^2(y)\,dV(y)\\
            &\leq c^{-1}\int_{V_a}|\nabla^\ell f|^2(y)\,dV(y).
        \end{aligned}
    \]
    Taking square roots yields
    \begin{equation}\label{eq:derivative-l2-bound}
        \bigl\|\,|\nabla^\ell f|\circ\Theta_{\theta,t,z}\,\bigr\|_{L_2(U_a)} \leq C\|\nabla^\ell f\|_{L_2(V_a)}.
    \end{equation}
    On each $U_a$, the orthonormal frame identifies integration over $T_xX$ with Lebesgue integration over $\mathbb R^\varrho$. In these coordinates, the Gaussian weight and the condition $\|\theta tz\|\leq2\delta$ are independent of $x$. Applying Minkowski's integral inequality to the pointwise remainder bound moves the $L_2(U_a)$ norm inside the integrals in $\theta$ and $z$. The estimates \eqref{eq:remainder-bound} and \eqref{eq:derivative-l2-bound} then give
    \[
        \begin{aligned}
            &\|\mathcal R_{L,t}f\|_{L_2(U_a)}\\
            \leq{}& C|t|^L\sum_{\ell=0}^{L}\int_0^1(1-\theta)^{L-1}
            \int_{\{z\in\mathbb R^\varrho:\|\theta tz\|\leq2\delta\}}
            (1+\|z\|^2)^{M/2}\exp(-c\|z\|^2)\\
            &\qquad\times\bigl\|\,|\nabla^\ell f|\circ\Theta_{\theta,t,z}\,\bigr\|_{L_2(U_a)}\,dz\,d\theta\\
            \leq{}& C|t|^L\sum_{\ell=0}^{L}\|\nabla^\ell f\|_{L_2(V_a)}
            \int_0^1(1-\theta)^{L-1}\int_{\mathbb R^\varrho}(1+\|z\|^2)^{M/2}\exp(-c\|z\|^2)\,dz\,d\theta.
        \end{aligned}
    \]
    Both scalar integrals are finite and independent of $t$, so they can be absorbed into $C$. Since the finitely many sets $U_a$ cover $X$, summing these local bounds yields
    \begin{equation}\label{eq:local-remainder}
        \|\mathcal R_{L,t}f\|_{L_2(X)}\leq\sum_a\|\mathcal R_{L,t}f\|_{L_2(U_a)} \leq C|t|^L\sum_{\ell=0}^{L}\|\nabla^\ell f\|_{L_2(X)} \leq C|t|^L\|f\|_{H^L(X)}.
    \end{equation}

    Note that $\mathcal{I}_t=T_{|t|}^{\mathrm{loc}}$ is even in $t$. Thus its odd Taylor coefficients vanish. Define
    \[
        A_rf := \frac1{(2r)!} \left.\partial_t^{2r}\mathcal I_tf\right|_{t=0}.
    \]
    This definition is independent of the truncation order $N$. It is also independent of the auxiliary cutoff as $\chi_0$ is identically one near zero.

    To compute the term where all $2r$ derivatives fall on $f(\mathcal{E}_x(tz))$, fix $x$ and use geodesic normal coordinates $u$. The radial derivative formula identifies $(\nabla^{2r}f)_x(z,\ldots,z)$ with the $2r$-th directional derivative of $f\circ\mathcal E_x$ at the origin in direction $z$. When we expand this derivative, all terms with an odd power of any $z_i$ integrate to zero. The remaining terms are evaluated using
    \[
        \begin{aligned}
            \pi^{-1/2}\int_{\mathbb R}\exp(-s^2)s^{2k}\,ds
            &=\frac{(2k)!}{4^k k!},\quad k\geq0.
        \end{aligned}
    \]
    Thus, summing over nonnegative integers $k_1,\ldots,k_\varrho$ with total $r$ and then applying the multinomial formula, we obtain
    \[
        \begin{aligned}
            \frac{\pi^{-\varrho/2}}{(2r)!} \int_{\mathbb{R}^\varrho} \exp(-\|z\|^2) (\nabla^{2r}f)_x
            \underbrace{(z,\ldots,z)}_{2r\text{ arguments}}\,dz
            &=\frac1{4^r}\sum_{k_1+\cdots+k_\varrho=r}
            \frac{\partial_{u_1}^{2k_1}\cdots\partial_{u_\varrho}^{2k_\varrho}(f\circ\mathcal E_x)(0)}{k_1!\cdots k_\varrho!}\\
            &=\frac1{4^r r!}\left(\sum_{i=1}^{\varrho}\partial_{u_i}^2\right)^r(f\circ\mathcal E_x)(0).
        \end{aligned}
    \]
    By the coordinate formula for $\Delta_X$ and the normal-coordinate identities in \cite[Proposition 2.46 and Proposition 5.24]{Lee2018Introduction}, its second-order coefficients agree with those of the Euclidean Laplacian to first order at the origin, while its first-order coefficients vanish there. Consequently, the terms of order $2r$ agree and those of order $2r-1$ vanish at the origin when taking the $r$-th power, so for $r\geq1$ we obtain
    \[
        \begin{aligned}
            \frac1{4^r r!}\left(\sum_{i=1}^{\varrho}\partial_{u_i}^2\right)^r(f\circ\mathcal E_x)(0)
            &=\frac1{4^r r!}(\Delta_X^rf)(x) + (\widetilde{B_r}f)(x), \quad \widetilde{B_r} \in \operatorname{Diff}^{2r-2}(X).
        \end{aligned}
    \]

    We now consider the remaining Leibniz terms, in which at least one derivative falls on
    \[
        \exp(-\Psi_x(t,z))J_x(tz)\chi_0(t^2\|z\|^2).
    \]
    Since $q_x$ and $J_x$ are smooth, the estimates \eqref{eq:normal-coordinate-estimates} imply, for every $x\in X$,
    \[
        D_uJ_x(0)=0, D_u^3q_x(0)=0.
    \]
    Differentiating \eqref{eq:psi-integral-representation} with respect to $t$ and evaluating at zero gives, for each fixed $x,z$,
    \[
        \partial_t\Psi_x(0,z) =\int_0^1\theta(1-\theta)D_u^3q_x(0)[z,z,z]\,d\theta = 0.
    \]
    Together with the vanishing of $D_uJ_x(0)$ and the constancy of the cutoff near zero, this gives
    \[
        \begin{aligned}
            \left.\partial_t\left[\exp(-\Psi_x(t,z))J_x(tz)\chi_0(t^2\|z\|^2)\right]\right|_0
            &=0.
        \end{aligned}
    \]
    Thus the terms with exactly one derivative on this factor vanish. Every nonzero remaining term has at least two derivatives on this factor and hence at most $2r-2$ derivatives on $f$. This proves that $A_r$ has the stated form for $r\geq1$.

    Now we are ready to prove the final estimate \eqref{eq:uniform-sobolev-expansion}. For $f\in C^\infty(X)$ and $0<\gamma\leq1$, we start from the decomposition
    \[
        T_\gamma f =T_\gamma^{\mathrm{far}}f+T_\gamma^{\mathrm{loc}}f =T_\gamma^{\mathrm{far}}f+\mathcal I_\gamma f.
    \]
    The Taylor expansion of $\mathcal I_t f$ above retains terms through degree $2N+1$. Since its odd coefficients vanish, the definition of $A_r$ gives
    \[
        \begin{aligned}
            \mathcal I_\gamma f
            &=\sum_{r=0}^{N}\gamma^{2r}A_rf+\mathcal R_{L,\gamma}f.
        \end{aligned}
    \]
    Substituting this identity into the decomposition yields
    \[
        \begin{aligned}
            T_\gamma f-\sum_{r=0}^{N}\gamma^{2r}A_rf
            &=T_\gamma^{\mathrm{far}}f+\mathcal R_{L,\gamma}f.
        \end{aligned}
    \]
    By \eqref{eq:far-gaussian-tail} and the fact that $\gamma^{-\varrho-L}\exp(-c_\delta^2/\gamma^2)$ is bounded for $0<\gamma\leq1$, we have
    \[
        \begin{aligned}
            \|T_\gamma^{\mathrm{far}}f\|_{L_2(X)}
            &\leq C_{X,\delta}\gamma^{-\varrho}\exp(-c_\delta^2/\gamma^2)\|f\|_{L_2(X)}\\
            &\leq C\gamma^L\|f\|_{L_2(X)}.
        \end{aligned}
    \]
    The local remainder estimate \eqref{eq:local-remainder}, evaluated at $\gamma$, gives
    \[
        \begin{aligned}
            \|\mathcal R_{L,\gamma}f\|_{L_2(X)}
            &\leq C\gamma^L\|f\|_{H^L(X)}.
        \end{aligned}
    \]
    Thus,
    \[
        \begin{aligned}
            \left\|T_\gamma f-\sum_{r=0}^{N}\gamma^{2r}A_rf\right\|_{L_2(X)}
            &\leq\|T_\gamma^{\mathrm{far}}f\|_{L_2(X)}+\|\mathcal R_{L,\gamma}f\|_{L_2(X)}\\
            &\leq C\gamma^L\bigl(\|f\|_{L_2(X)}+\|f\|_{H^L(X)}\bigr)\\
            &\leq C_{N,X,d}\gamma^{2N+2}\|f\|_{H^{2N+2}(X)}.
        \end{aligned}
    \]
    This proves \eqref{eq:uniform-sobolev-expansion} for smooth $f$. For $f\in H^L(X)$, choose $f_m\in C^\infty(X)$ with $f_m\to f$ in $H^L(X)$. This convergence also holds in $L_2(X)$. For each fixed $\gamma$, the boundedness of $T_\gamma:L_2(X)\to L_2(X)$ therefore gives $T_\gamma f_m\to T_\gamma f$ in $L_2(X)$. Moreover, each $A_r$ has order at most $2r\leq L$, so it is continuous from $H^L(X)$ to $L_2(X)$ and $A_rf_m\to A_rf$ in $L_2(X)$. Applying the estimate just proved to $f_m$ and passing to the limit proves \eqref{eq:uniform-sobolev-expansion} for every $f\in H^L(X)$. We thus complete the proof.
\end{proof}

\begin{proof}[Proof of Theorem~\ref{thm:gaussian-sobolev-stability}]
    Fix $r\in\mathbb N_0$. We first consider $f\in C^\infty(X)$ and use the cutoff $\chi$ and decomposition from the proof of Theorem~\ref{thm:gaussian-sobolev-expansion}:
    \[
        \begin{aligned}
            T_{\gamma}&=T_{\gamma}^{\mathrm{loc}}+T_{\gamma}^{\mathrm{far}}.
        \end{aligned}
    \]
    On the support of $1-\chi$, the separation estimate used in \eqref{eq:far-gaussian-tail} gives $\|x-y\|_{\mathbb R^d}\geq c_\delta$. Hence, for some constants $c, C_r>0$ and $M_r\geq0$,
    \[
        \begin{aligned}
            \left|\nabla_x^j K_\gamma^{\mathrm{far}}(x,y)\right|
            &\leq C_r\gamma^{-M_r}\exp(-c/\gamma^2),
            \quad 0\leq j\leq r,\quad 0<\gamma\leq1.
        \end{aligned}
    \]
    Applying the Schur test to each differentiated kernel gives
    \[
        \begin{aligned}
            \|T_{\gamma}^{\mathrm{far}}f\|_{H^r(X)}
            &\leq C_r\gamma^{-M_r}\exp(-c/\gamma^2)\|f\|_{L_2(X)}
            \leq C_r\|f\|_{H^r(X)}.
        \end{aligned}
    \]

    For the local part, use the same finite coordinate neighborhoods $U_a\Subset V_a$ and smooth orthonormal frames $e_1,\ldots,e_\varrho$ as in the proof of Theorem~\ref{thm:gaussian-sobolev-expansion}. For $x\in U_a$ and $\|z\|<3\delta$, define the squared ambient distance and the volume density in this frame by
    \[
        \begin{aligned}
            Q_a(x,z)&:=\left\|x-\mathcal E_x\!\left(\sum_{j=1}^{\varrho}z_j e_j(x)\right)\right\|_{\mathbb R^d}^2,\\
            J_a(x,z)&:=J_x\!\left(\sum_{j=1}^{\varrho}z_j e_j(x)\right).
        \end{aligned}
    \]
    Both functions are smooth in $x$ and $z$. In the local integral, write $u=\gamma\sum_{j=1}^{\varrho}z_j e_j(x)$, so that $\mathcal E_x(u)=\Theta_{1,\gamma,z}(x)$ and $du_x=\gamma^\varrho\,dz$. When differentiating in $x$, we hold the coefficients $z\in\mathbb R^\varrho$ fixed in the moving frame. Since $Q_a(x,0)=0$ and $D_zQ_a(x,0)=0$, Taylor's formula yields
    \[
        \Psi_{a,\gamma}(x,z)=\gamma^{-2}Q_a(x,\gamma z)=\int_0^1(1-t)D_z^2Q_a(x,t\gamma z)[z,z]\,dt.
    \]
    On the support of the cutoff, $\|\gamma z\|\leq2\delta$. The integral representation shows that every fixed-order coordinate derivative in $x$ of $\Psi_{a,\gamma}$ is bounded by a constant times $\|z\|^2$, uniformly in $\gamma$. Moreover, \eqref{eq:normal-coordinate-estimates} gives the lower bound $\Psi_{a,\gamma}(x,z)\geq c\|z\|^2$. Thus, for the amplitude
    \[
        \begin{aligned}
            A_{a,\gamma}(x,z)
            &=\pi^{-\varrho/2}\exp(-\Psi_{a,\gamma}(x,z))
            J_a(x,\gamma z)\chi_0(\gamma^2\|z\|^2),
        \end{aligned}
    \]
    extended by zero outside $\|\gamma z\|<3\delta$, the chain rule gives an integrable function $G_r$ on $\mathbb R^\varrho$ such that
    \[
        \begin{aligned}
            |D_x^\beta A_{a,\gamma}(x,z)|
            &\leq G_r(z)\mathbf1_{\{\|\gamma z\|\leq2\delta\}},
            \quad |\beta|\leq r,\quad x\in U_a.
        \end{aligned}
    \]
    We may take $G_r(z)$ to be a polynomial in $\|z\|$ multiplied by $\exp(-c\|z\|^2)$, uniformly in $a$ and $\gamma$. The cutoff is independent of $x$ in an orthonormal frame.

    The local operator has the representation
    \[
        \begin{aligned}
            T_{\gamma}^{\mathrm{loc}}f(x)
            &=\int_{\mathbb R^\varrho}A_{a,\gamma}(x,z)
            f(\Theta_{1,\gamma,z}(x))\,dz,\quad x\in U_a.
        \end{aligned}
    \]
    On $\|\gamma z\|\leq2\delta$, all coordinate derivatives of $\Theta_{1,\gamma,z}$ through order $r$ are uniformly bounded. This follows from smoothness of the exponential map and the frame, since the tangent vector has uniformly bounded length. The chain rule therefore bounds every derivative of $f\circ\Theta_{1,\gamma,z}$ through order $r$ by
    \[
        \begin{aligned}
            |D_x^\beta(f\circ\Theta_{1,\gamma,z})(x)|
            &\leq C_r\sum_{\ell=0}^r
            |\nabla^\ell f|(\Theta_{1,\gamma,z}(x)),\quad |\beta|\leq r.
        \end{aligned}
    \]
    After decreasing the fixed cutoff radius if necessary, $\Theta_{1,\gamma,z}$ maps $U_a$ injectively into $V_a$ with Jacobian uniformly bounded away from zero, exactly as in \eqref{eq:derivative-l2-bound}. Consequently,
    \[
        \begin{aligned}
            \bigl\|\,|\nabla^\ell f|\circ\Theta_{1,\gamma,z}\,\bigr\|_{L_2(U_a)}
            &\leq C\|\nabla^\ell f\|_{L_2(V_a)},
            \quad 0\leq\ell\leq r,\quad \|\gamma z\|\leq2\delta.
        \end{aligned}
    \]
    Differentiating the local representation, applying Leibniz's rule and Minkowski's inequality, and then using these bounds gives
    \[
        \begin{aligned}
            \sum_{|\beta|\leq r}\|D_x^\beta T_{\gamma}^{\mathrm{loc}}f\|_{L_2(U_a)}
            &\leq C_r\int_{\mathbb R^\varrho}G_r(z)\,dz
            \sum_{\ell=0}^r\|\nabla^\ell f\|_{L_2(V_a)}\\
            &\leq C_r\|f\|_{H^r(X)}.
        \end{aligned}
    \]
    The finite covering and equivalence of coordinate and covariant Sobolev norms now yield the same bound for $T_{\gamma}^{\mathrm{loc}}$ in $H^r(X)$. Combining the local and far bounds and using density proves the estimate for nonnegative integer $r$. For $r=m+\theta$ with $m\in\mathbb N_0$ and $0<\theta<1$, interpolation of the spectral Sobolev spaces gives
    \[
        \begin{aligned}
            H^r(X)&=[H^m(X),H^{m+1}(X)]_\theta,\\
            \|T_\gamma\|_{H^r(X)\to H^r(X)}
            &\leq \|T_\gamma\|_{H^m(X)\to H^m(X)}^{1-\theta}
                  \|T_\gamma\|_{H^{m+1}(X)\to H^{m+1}(X)}^\theta
             \leq C_m^{1-\theta}C_{m+1}^\theta .
        \end{aligned}
    \]
    The kernel of $T_\gamma$ is symmetric with respect to $dV$, so $T_\gamma$ is self-adjoint on $L_2(X)$. Duality under the $L_2(X)$ pairing gives the extension and uniform bound on $H^{-r}(X)$ for $r>0$.
\end{proof}

\begin{proof}[Proof of Theorem~\ref{thm:besov-approximation}]
    Theorem~\ref{thm:gaussian-sobolev-stability} bounds each operator $T_{j\gamma}$ in \eqref{eq:multiscale-operator} with $j\gamma\leq1$ on the Sobolev scale. For $1\leq j\gamma\leq s$, the kernel derivatives are uniformly bounded on $X\times X$. This gives an $L_2(X)$ to $H^m(X)$ bound for every nonnegative integer $m$. Symmetry of the kernel and duality give the bounds for negative orders. Interpolation therefore gives, for each $t\in\mathbb R$,
    \[
        \|S_{\gamma,s}-I\|_{H^t(X)\to H^t(X)}\leq C,\quad 0<\gamma\leq1.
    \]
    Fix an integer $m>\alpha$ and let $f\in B^\alpha_{2,\infty}(X)$. For each decomposition $f=f_0+f_1$ with $f_0\in L_2(X)$ and $f_1\in H^m(X)$, 
    \[
        S_{\gamma,s}f-f=(S_{\gamma,s}f_0-f_0)+(S_{\gamma,s}f_1-f_1).
    \]
    Applying the preceding operator bound at orders $0$ and $m$ to these two terms gives, for every $t>0$,
    \[
        \begin{aligned}
            K(t,S_{\gamma,s}f-f;L_2(X),H^m(X))
            &\leq \|S_{\gamma,s}f_0-f_0\|_{L_2(X)}
                  +t\|S_{\gamma,s}f_1-f_1\|_{H^m(X)}\\
            &\leq C\bigl(\|f_0\|_{L_2(X)}+t\|f_1\|_{H^m(X)}\bigr),
        \end{aligned}
    \]
    with $C$ independent of $t$ and $\gamma$. Taking the infimum over all such decompositions bounds the left-hand side by $C K(t,f;L_2(X),H^m(X))$. Multiplying by $t^{-\alpha/m}$ and taking the supremum over $t>0$ now gives, by \eqref{eq:besov-interpolation},
    \begin{equation}\label{eq:besov-gaussian-stability}
        \|S_{\gamma,s}f-f\|_{B^\alpha_{2,\infty}(X)}\leq C\|f\|_{B^\alpha_{2,\infty}(X)}.
    \end{equation}

    Set $\beta:=\max\{0,-r\}$ and $L:=2\lfloor(\alpha+\beta)/2\rfloor+2$. Thus $L$ is the smallest even integer greater than $\alpha+\beta$, and $L-2\leq\alpha+\beta<s$. We first establish an estimate for $g\in H^L(X)$. For $s\gamma\leq1$, Theorem~\ref{thm:gaussian-sobolev-expansion} at order $L/2-1$ and the identities \eqref{eq:binomial-moments} give
    \[
        \begin{aligned}
            S_{\gamma,s}g-g
            &=\sum_{j=1}^s c_j\left[T_{j\gamma}g-
               \sum_{k=0}^{L/2-1}(j\gamma)^{2k}A_kg\right],\\
            \|S_{\gamma,s}g-g\|_{L_2(X)}
            &\leq C\gamma^L\|g\|_{H^L(X)}.
        \end{aligned}
    \]
    For $1/s<\gamma\leq1$, the uniform $L_2(X)$ bound gives the same estimate because $\gamma^L\geq s^{-L}$. By self-adjointness of $S_{\gamma,s}-I$, its norm from $L_2(X)$ to $H^{-L}(X)$ is also bounded by $C\gamma^L$. For $\beta>0$, we interpolate these two operator bounds with parameter $\beta/L$. For $\beta=0$, the first bound applies directly. In both cases,
    \begin{equation}\label{eq:gaussian-approximation-shifted}
        \|S_{\gamma,s}g-g\|_{H^{-\beta}(X)}\leq C\gamma^L\|g\|_{H^{L-\beta}(X)},\quad g\in H^{L-\beta}(X).
    \end{equation}

    Choose $J_0\in\mathbb N_0$ so that $2^{J_0}\leq\gamma^{-1}<2^{J_0+1}$. Write $f_{\mathrm{lo}}:=\sum_{j\leq J_0}P_jf$ and $f_{\mathrm{hi}}:=f-f_{\mathrm{lo}}$. By \eqref{eq:sobolev-spectral-blocks} and \eqref{eq:besov-spectral},
    \[
        \begin{aligned}
            \|f_{\mathrm{lo}}\|_{H^{L-\beta}(X)}
            &\leq C 2^{J_0(L-\beta-\alpha)}\|f\|_{B^\alpha_{2,\infty}(X)},\\
            \|f_{\mathrm{hi}}\|_{H^{-\beta}(X)}
            &\leq C 2^{-J_0(\alpha+\beta)}\|f\|_{B^\alpha_{2,\infty}(X)}.
        \end{aligned}
    \]
    Since $f_{\mathrm{lo}}$ is a finite linear combination of smooth eigenfunctions of $-\Delta_X$, we may apply \eqref{eq:gaussian-approximation-shifted} to $f_{\mathrm{lo}}$. For $f_{\mathrm{hi}}$, we use the uniform $H^{-\beta}(X)$ bound for $S_{\gamma,s}-I$. Adding the two estimates yields
    \[
        \begin{aligned}
            \|S_{\gamma,s}f-f\|_{H^{-\beta}(X)}
            &\leq C\left(\gamma^L 2^{J_0(L-\beta-\alpha)}+2^{-J_0(\alpha+\beta)}\right)
                \|f\|_{B^\alpha_{2,\infty}(X)}\\
            &\leq C\gamma^{\alpha+\beta}\|f\|_{B^\alpha_{2,\infty}(X)}.
        \end{aligned}
    \]
    This is the desired estimate when $r\leq0$.

    For $0<r<\alpha$ we have $\beta=0$, so the last estimate gives $\|S_{\gamma,s}f-f\|_{L_2(X)}\leq C\gamma^\alpha\|f\|_{B^\alpha_{2,\infty}(X)}$. To pass to $H^r(X)$, take any $v\in B^\alpha_{2,\infty}(X)$ and split the sum in \eqref{eq:sobolev-spectral-blocks} at $J_0$. For $j\leq J_0$, we have $2^{jr}\leq 2^{J_0r}$. Orthogonality of the spectral projections in $L_2(X)$ then gives
    \[
        \sum_{j=0}^{J_0}2^{2jr}\|P_jv\|_{L_2(X)}^2
        \leq 2^{2J_0r}\sum_{j=0}^{J_0}\|P_jv\|_{L_2(X)}^2
        \leq 2^{2J_0r}\|v\|_{L_2(X)}^2.
    \]
    For $j>J_0$, \eqref{eq:besov-spectral} gives $\|P_jv\|_{L_2(X)}\leq C2^{-j\alpha}\|v\|_{B^\alpha_{2,\infty}(X)}$. Since $\alpha-r>0$, we have
    \[
        \begin{aligned}
            \sum_{j>J_0}2^{2jr}\|P_jv\|_{L_2(X)}^2
            &\leq C\|v\|_{B^\alpha_{2,\infty}(X)}^2
                    \sum_{j>J_0}2^{-2j(\alpha-r)}\\
            &\leq C 2^{-2J_0(\alpha-r)}\|v\|_{B^\alpha_{2,\infty}(X)}^2.
        \end{aligned}
    \]
    Adding the two bounds and taking square roots yields
    \[
        \|v\|_{H^r(X)}
        \leq C\left(2^{J_0r}\|v\|_{L_2(X)}
              +2^{-J_0(\alpha-r)}\|v\|_{B^\alpha_{2,\infty}(X)}\right).
    \]
    Now take $v=S_{\gamma,s}f-f$. The $L_2(X)$ error estimate supplies a factor $\gamma^\alpha$ in the first term, while \eqref{eq:besov-gaussian-stability} bounds the Besov norm in the second. Since $2^{J_0}\leq\gamma^{-1}<2^{J_0+1}$, we have $2^{J_0r}\leq\gamma^{-r}$ and $2^{-J_0}<2\gamma$. Hence
    \[
        \begin{aligned}
            \|S_{\gamma,s}f-f\|_{H^r(X)}
            &\leq C\bigl(2^{J_0r}\gamma^\alpha+2^{-J_0(\alpha-r)}\bigr)
                    \|f\|_{B^\alpha_{2,\infty}(X)}\\
            &\leq C\gamma^{\alpha-r}\|f\|_{B^\alpha_{2,\infty}(X)},
        \end{aligned}
    \]
    which proves \eqref{eq:besov-approximation}.
\end{proof}

\subsection{Proofs for Section~\ref{sec:logistic-application}}

In the classification proofs, for a $P$-integrable measurable function $g:X\times Y\to\mathbb R$, we write
\[
    \begin{aligned}
        \mathbb{E}_P g &:= \int_{X\times Y}g(x,y)\,dP(x,y),\\
        \mathbb{E}_D g &:= \frac1n\sum_{i=1}^n g(X_i,Y_i).
    \end{aligned}
\]
Thus $\mathbb{E}_D$ denotes the average over the observed samples. Expectations over the random samples are written as $\mathbb{E}_{D\sim P^n}$.

\begin{proof}[Proof of Lemma~\ref{lem:logistic-variance}]
    At a point where $\eta\in\{0,1\}$, since $0\leq\phi(t)\leq |t|+\log2\leq |t|+2$, one has $\phi(t)^2\leq(|t|+2)\phi(t)$, which proves the inequality at either endpoint. Fix now $\eta\in(0,1)$ and $t\in\mathbb R$, and set
    \[
        A:=\phi(t)+\log\eta, \quad B:=\phi(-t)+\log(1-\eta).
    \]
    It is enough to prove
    \begin{equation}\label{eq:pointwise-logistic-variance}
        \eta A^2+(1-\eta)B^2 \leq (|t|+2)\bigl(\eta A+(1-\eta)B\bigr).
    \end{equation}
    Let
    \[
        F(\eta,t) := (|t|+2)\bigl(\eta A+(1-\eta)B\bigr)-\bigl(\eta A^2+(1-\eta)B^2\bigr).
    \]
    Direct differentiation gives
    \[
        \partial_\eta F(\eta,t) = \bigl(|t|-A-B\bigr)(A-B).
    \]
    Note that
    \[
        |t|-\phi(t)-\phi(-t)=-2\log(1+e^{-|t|})\geq -2\log2 \geq  \log\eta+\log(1-\eta).
    \]
    Hence $|t|-A-B\geq0$. Moreover, $A-B$ is strictly increasing in $\eta$ and vanishes exactly at $\eta^*=e^t/(1+e^t)$. At this point $A=B=0$, so $F(\eta^*,t)=0$. Thus $F(\eta,t)\geq0$, proving \eqref{eq:pointwise-logistic-variance}. Applying this inequality with $t=f(x)$ and integrating over $P_X$ completes the proof.
\end{proof}

\begin{proof}[Proof of Theorem~\ref{thm:oracle-inequality}]
    By the definition of the infimum in \eqref{eq:approximation-error-function} and comparison with the zero function, choose $f_\lambda\in\mathcal{F}$ such that
    \[
        a:=\lambda\|f_\lambda\|_{\mathcal{F}}^2+\mathcal R(f_\lambda)-\mathcal R^*
        \leq\min\bigl\{a_{\mathcal{F}}(\lambda)+\lambda,\log2-\mathcal R^*\bigr\}.
    \]
    For $f \in \mathcal{F}$, set
    \[
        g_f(z) := \lambda\|f\|_{\mathcal{F}}^2+\phi(yf(x)) -\lambda\|f_\lambda\|_{\mathcal{F}}^2-\phi(yf_\lambda(x)), \quad z=(x,y).
    \]
    We see that
    \[
        \mathbb{E}_P g_f = \lambda\|f\|_{\mathcal{F}}^2+\mathcal R(f)-\lambda\|f_\lambda\|_{\mathcal{F}}^2-\mathcal R(f_\lambda) = \lambda\|f\|_{\mathcal{F}}^2+ \mathcal{R}(f) - \mathcal{R}^* - a.
    \]
    Comparison of the empirical objective with its value at zero gives $\lambda\|\widehat f_{D,\lambda}\|_{\mathcal{F}}^2\leq\log2$. The choice of $f_\lambda$ and the nonnegativity of the excess risk give $\lambda\|f_\lambda\|_{\mathcal{F}}^2\leq a\leq\log2$. Thus both $f_\lambda$ and $\widehat f_{D,\lambda}$ belong to $\lambda^{-1/2}B_{\mathcal{F}}$.

    We only consider $f\in\lambda^{-1/2}B_{\mathcal{F}}$, and define the slice
    \[
        \mathcal F_{\lambda,\varepsilon} := \{f\in\lambda^{-1/2}B_{\mathcal{F}}:\mathbb{E}_P g_f=\varepsilon\}.
    \]
    The map $f\mapsto\mathbb{E}_P g_f$ is continuous with respect to the norm $\|\cdot\|_{\mathcal F}$ because the logistic loss is $1$-Lipschitz and $\|f\|_\infty\leq\|f\|_{\mathcal{F}}$. Hence the slice $\mathcal F_{\lambda,\varepsilon}$ is a closed separable subset of $\mathcal{F}$.

    For each fixed sample set $D$, the map $f\mapsto\mathbb{E}_D g_f$ is convex. Indeed, $f\mapsto\|f\|_{\mathcal F}^2$ is convex, and each $f\mapsto\phi(Y_if(X_i))$ is convex as the composition of the convex loss $\phi$ with a linear functional.

    Let $\varepsilon>0$ and suppose that $f\in\lambda^{-1/2}B_{\mathcal F}$ satisfies $\mathbb{E}_D g_f\leq0$ and $\mathbb{E}_P g_f\geq\varepsilon$. For $0\leq\theta\leq1$, set $f_\theta=(1-\theta)f_\lambda+\theta f$. Since both endpoints belong to the convex ball $\lambda^{-1/2}B_{\mathcal F}$, so does every $f_\theta$. The convexity of $f\mapsto\mathbb{E}_D g_f$ and $g_{f_\lambda}=0$ give
    \[
        \mathbb{E}_D g_{f_\theta} \leq (1-\theta)\mathbb{E}_D g_{f_\lambda}+\theta\mathbb{E}_D g_f =\theta\mathbb{E}_D g_f\leq0.
    \]
    Moreover, the continuous map $\theta\mapsto\mathbb{E}_P g_{f_\theta}$ takes the value $0$ at $\theta=0$ and the value $\mathbb{E}_P g_f\geq\varepsilon$ at $\theta=1$. The intermediate value theorem therefore gives a $\theta\in[0,1]$ for which $f_\theta\in\mathcal F_{\lambda,\varepsilon}$ and $\mathbb{E}_D g_{f_\theta}\leq0$. To apply this reduction to the estimator, note that $\mathbb{E}_D g_{\widehat f_{D,\lambda}}\leq0$ and $\widehat f_{D,\lambda}\in\lambda^{-1/2}B_{\mathcal F}$. The preceding argument shows that whenever $\mathbb{E}_P g_{\widehat f_{D,\lambda}}\geq\varepsilon$, there is an $f\in\mathcal F_{\lambda,\varepsilon}$ with $\mathbb{E}_D g_f\leq0$. Consequently,
    \[
        \mathbb P_{D\sim P^n}\left\{\mathbb{E}_P g_{\widehat f_{D,\lambda}}\geq\varepsilon\right\} \leq \mathbb P_{D\sim P^n}\left\{\mathbb{E}_D g_f\leq0\text{ for some }f\in\mathcal F_{\lambda,\varepsilon}\right\}.
    \]
    Moreover, the definitions of $a$ and $g_f$ give
    \[
        \mathcal R(\widehat f_{D,\lambda})-\mathcal R^* =a+\mathbb{E}_P g_{\widehat f_{D,\lambda}}-\lambda\|\widehat f_{D,\lambda}\|_{\mathcal F}^2 \leq a+\mathbb{E}_P g_{\widehat f_{D,\lambda}}.
    \]
    Thus, bounding the probability on the right by $e^{-t}$ ensures that the excess risk is at most $a+\varepsilon$ with probability at least $1-e^{-t}$. It therefore suffices to control the event on the slice $\mathcal F_{\lambda,\varepsilon}$.

    Assume $\varepsilon\geq a+\lambda$. For every $f\in\mathcal F_{\lambda,\varepsilon}$,
    \begin{equation}\label{eq:localized-radii}
        \|f\|_{\mathcal{F}}^2\leq\frac{a+\varepsilon}{\lambda} \leq\frac{2\varepsilon}{\lambda}, \quad \|f_\lambda\|_{\mathcal{F}}^2\leq\frac{a}{\lambda} \leq\frac{\varepsilon}{\lambda}.
    \end{equation}
    Define the loss difference
    \[
        q_f(x,y):=\phi(yf(x))-\phi(yf_\lambda(x)),
    \]
    and the auxiliary class
    \[
        \mathcal Q_{\lambda,\varepsilon} :=\{q_f:f\in\mathcal F_{\lambda,\varepsilon}\}.
    \]
    Since $f,f_\lambda\in\lambda^{-1/2}B_{\mathcal{F}}$ and $0<\lambda\leq1$,
    \[
        \varepsilon=\mathbb{E}_P g_f \leq \lambda\|f\|_{\mathcal{F}}^2 + \mathcal{R}(f) \leq 1+ \|f\|_\infty + \log 2 \leq 1 + \lambda^{-1/2} + \log 2 \leq3/\lambda.
    \]
    We first establish boundedness in the supremum norm. The $1$-Lipschitz property of $\phi$ and \eqref{eq:localized-radii} give
    \begin{equation}\label{eq:localized-boundedness}
        \|q_f\|_\infty\leq\|f-f_\lambda\|_\infty
        \leq\|f\|_{\mathcal F}+\|f_\lambda\|_{\mathcal F}
        \leq C\varepsilon^{1/2}\lambda^{-1/2}.
    \end{equation}
    Next, we use the Bernstein condition in Lemma~\ref{lem:logistic-variance} to bound the second moment. It gives
    \begin{equation}\label{eq:localized-variance}
        \begin{aligned}
            \mathbb{E}_P(q_f^2) &\leq 2\mathbb{E}_P\bigl[(\phi(Yf)-\psi)^2\bigr] +2\mathbb{E}_P\bigl[(\phi(Yf_\lambda)-\psi)^2\bigr]\\
            &\leq 2(\|f\|_\infty+2)(\mathcal R(f)-\mathcal R^*) +2(\|f_\lambda\|_\infty+2) (\mathcal R(f_\lambda)-\mathcal R^*)\\
            &\leq C\varepsilon^{3/2}\lambda^{-1/2}.
        \end{aligned}
    \end{equation}
    The last inequality uses \eqref{eq:localized-radii} and the bounds $\mathcal R(f)-\mathcal R^*\leq a+\varepsilon\leq2\varepsilon$ and $\mathcal R(f_\lambda)-\mathcal R^*\leq a\leq\varepsilon$.

    The regularization terms are constants and cancel in $(\mathbb{E}_P-\mathbb{E}_D)g_f$, so that $(\mathbb{E}_P-\mathbb{E}_D)g_f=(\mathbb{E}_P-\mathbb{E}_D)q_f$. Hence we only need to estimate the complexity of the auxiliary class $\mathcal Q_{\lambda,\varepsilon}$. Fix a sample set $D$ and define $\Phi_D:L_2(D_X)\to L_2(D)$ on the sample points by
    \[
        (\Phi_D u)(X_j,Y_j):=\phi(Y_ju(X_j))-\phi(Y_jf_\lambda(X_j)), \quad 1\leq j\leq n.
    \]
    The $1$-Lipschitz property of $\phi$ implies that, for any $u,v\in L_2(D_X)$,
    \[
        \begin{aligned}
            \|\Phi_D u-\Phi_D v\|_{L_2(D)}^2
            &=\frac1n\sum_{j=1}^n\bigl|\phi(Y_ju(X_j))-\phi(Y_jv(X_j))\bigr|^2\\
            &\leq\frac1n\sum_{j=1}^n|u(X_j)-v(X_j)|^2\\
            &=\|u-v\|_{L_2(D_X)}^2.
        \end{aligned}
    \]
    In particular, $\Phi_D$ does not increase distances. Hence, by \eqref{eq:localized-radii} and \eqref{eq:average-entropy-assumption},
    \[
        e_i\bigl(\mathcal Q_{\lambda,\varepsilon},L_2(D)\bigr) \leq e_i\bigl(\mathcal F_{\lambda,\varepsilon},L_2(D_X)\bigr) \leq\left(\frac{2\varepsilon}{\lambda}\right)^{1/2}e_i\bigl(\operatorname{id}:\mathcal F\to L_2(D_X)\bigr).
    \]
    Consequently, \eqref{eq:average-entropy-assumption} shows that
    \begin{equation}\label{eq:localized-entropy}
        \mathbb{E}_{D\sim P^n} e_i\bigl(\mathcal Q_{\lambda,\varepsilon},L_2(D)\bigr) \leq C\left(\frac{\varepsilon}{\lambda}\right)^{1/2} A^{1/(2p)}i^{-1/(2p)}.
    \end{equation}
    Since the map $(z,f)\mapsto q_f(z)$ is measurable in $z$ and continuous in $f$, the class $\mathcal Q_{\lambda,\varepsilon}$ is a separable Carathéodory class. Applying \cite[Theorem 7.16]{Steinwart2008Support} with \eqref{eq:localized-boundedness}, \eqref{eq:localized-variance} and \eqref{eq:localized-entropy} gives the local Rademacher bound
    \begin{align}\label{eq:localized-rademacher}
        \mathbb{E}_{D\sim P^n}\operatorname{Rad}_D (\mathcal Q_{\lambda,\varepsilon}) \leq C\bigg[& \varepsilon^{(3-p)/4}\lambda^{-(1+p)/4} \left(\frac{A}{n}\right)^{1/2} +\varepsilon^{1/2}\lambda^{-1/2} \left(\frac{A}{n}\right)^{1/(1+p)} \bigg].
    \end{align}
    The constant $C$ is uniform for $p\in(0,1/2]$.

    Let $h_f:=\mathbb{E}_P q_f-q_f$. Since $(\mathbb{E}_P-\mathbb{E}_D)g_f=(\mathbb{E}_P-\mathbb{E}_D)q_f$, one has
    \[
        \mathbb P_{D\sim P^n}\left\{ \mathbb{E}_D g_f\leq0\text{ for some } f\in\mathcal F_{\lambda,\varepsilon} \right\} \leq \mathbb P_{D\sim P^n}\left\{ \sup_{f\in\mathcal F_{\lambda,\varepsilon}}\mathbb{E}_D h_f \geq\varepsilon \right\}.
    \]
    By symmetrization \cite[Proposition 7.10]{Steinwart2008Support},
    \[
        \mathbb{E}_{D\sim P^n}\sup_{f\in\mathcal F_{\lambda,\varepsilon}}\mathbb{E}_D h_f \leq 2\mathbb{E}_{D\sim P^n}\operatorname{Rad}_D (\mathcal Q_{\lambda,\varepsilon}).
    \]
    For every $f\in\mathcal F_{\lambda,\varepsilon}$, we have $\mathbb{E}_P h_f=0$, $\|h_f\|_\infty\leq2\|q_f\|_\infty$, and $\mathbb{E}_P(h_f^2)\leq\mathbb{E}_P(q_f^2)$. Thus \eqref{eq:localized-boundedness} and \eqref{eq:localized-variance} provide the supremum and variance bounds required by the simplified Talagrand's inequality \cite[Theorem 7.5]{Steinwart2008Support}. Together with the preceding symmetrization bound, it gives, for every $t\geq1$, with probability at least $1-e^{-t}$,
    \[
            \sup_{f\in\mathcal F_{\lambda,\varepsilon}}\mathbb{E}_D h_f
            <C\bigg[\mathbb{E}_{D\sim P^n}\operatorname{Rad}_D(\mathcal Q_{\lambda,\varepsilon})+\varepsilon^{3/4}\lambda^{-1/4}\left(\frac{t}{n}\right)^{1/2}+\varepsilon^{1/2}\lambda^{-1/2}\frac{t}{n}\bigg].
    \]
    Substituting \eqref{eq:localized-rademacher}, we obtain
    \[
        \mathbb P_{D\sim P^n}\left\{\sup_{f\in\mathcal F_{\lambda,\varepsilon}}\mathbb{E}_D h_f\geq\varepsilon\right\}
        \leq e^{-t},
    \]
    provided that $\varepsilon\geq a+\lambda$, as required for the localized estimates, and
    \[
        \begin{aligned}
            \varepsilon \geq{}& C\bigg[\varepsilon^{(3-p)/4}\lambda^{-(1+p)/4}\left(\frac{A}{n}\right)^{1/2}+\varepsilon^{1/2}\lambda^{-1/2}\left(\frac{A}{n}\right)^{1/(1+p)} \\
            &\qquad +\varepsilon^{3/4}\lambda^{-1/4}\left(\frac{t}{n}\right)^{1/2}+\varepsilon^{1/2}\lambda^{-1/2}\frac{t}{n}\bigg].
        \end{aligned}
    \]
    Choose
    \[
        \varepsilon :=C_0\left[ a+\lambda +\lambda^{-1}\left(\frac{A}{n}\right)^{2/(1+p)} +\lambda^{-1}\frac{t^2}{n^2} \right]
    \]
    with $C_0$ a sufficiently large numerical constant, so that both conditions hold. If $\mathcal F_{\lambda,\varepsilon}$ is empty, the earlier event inclusion gives $\mathbb{E}_P g_{\widehat f_{D,\lambda}}<\varepsilon$ for every $D$. Otherwise, the probability bound above and the same event inclusion show that $\mathbb{E}_P g_{\widehat f_{D,\lambda}}<\varepsilon$ with probability at least $1-e^{-t}$.

    Finally, note that $a \leq a_{\mathcal F}(\lambda)+\lambda$. We therefore have, when $\mathbb{E}_P g_{\widehat f_{D,\lambda}}<\varepsilon$,
    \[
        \begin{aligned}
            \mathcal R(\widehat f_{D,\lambda})-\mathcal R^*
            &=a+\mathbb{E}_P g_{\widehat f_{D,\lambda}}-\lambda\|\widehat f_{D,\lambda}\|_{\mathcal F}^2\\
            &\leq a+\varepsilon\\
            &\leq C\left[a_{\mathcal F}(\lambda)+\lambda
            +\lambda^{-1}\left(\frac{A}{n}\right)^{2/(1+p)}
            +\lambda^{-1}\frac{t^2}{n^2}\right].
        \end{aligned}
    \]
    This completes the proof of \eqref{eq:oracle-inequality}.
\end{proof}

\begin{proof}[Proof of Corollary~\ref{cor:truncated-entropy-oracle}]
    Use the function $f_\lambda$, the number $a$, and the classes $\mathcal F_{\lambda,\varepsilon}$ and $\mathcal Q_{\lambda,\varepsilon}$ from the proof of Theorem~\ref{thm:oracle-inequality}. The localization argument and the bounds \eqref{eq:localized-radii}, \eqref{eq:localized-boundedness}, and \eqref{eq:localized-variance} do not use the entropy assumption. Fix $\varepsilon\geq a+\lambda$ and first suppose that the slice $\mathcal{F}_{\lambda,\varepsilon}$ is nonempty. By \eqref{eq:localized-radii}, the slice is contained in $\left(2\varepsilon/\lambda\right)^{1/2}B_{\mathcal F}$. The map $f\mapsto q_f$ is $1$-Lipschitz in the supremum norm. Thus \eqref{eq:truncated-covering-assumption} gives
    \[
        \log N\left(\delta\left(\frac{2\varepsilon}{\lambda}\right)^{1/2},\mathcal Q_{\lambda,\varepsilon},\|\cdot\|_\infty\right)
        \leq A\delta^{-2p},\quad b\leq\delta\leq1.
    \]
    Choose such a cover at radius $b\left(2\varepsilon/\lambda\right)^{1/2}$. Since the centers of the covering balls need not belong to $\mathcal Q_{\lambda,\varepsilon}$, we select one member from each ball that intersects it. These representatives form a finite set $\mathcal N\subset\mathcal Q_{\lambda,\varepsilon}$ such that
    \begin{equation}\label{eq:finite-loss-net}
        \sup_{q\in\mathcal Q_{\lambda,\varepsilon}}\min_{v\in\mathcal N}\|q-v\|_\infty\leq2b\left(\frac{2\varepsilon}{\lambda}\right)^{1/2},
        \quad \log|\mathcal N|\leq Ab^{-2p}.
    \end{equation}

    Set $\mathcal N_0:=\mathcal N\cup\{0\}$ and consider a $\delta(2\varepsilon/\lambda)^{1/2}$-cover of it. For $b\leq\delta\leq1$, we can cover $\mathcal N$ using the corresponding $\delta\left(2\varepsilon/\lambda\right)^{1/2}$-cover of $\mathcal Q_{\lambda,\varepsilon}$ and add one ball for zero. For $0<\delta<b$, use one ball for each member of $\mathcal N_0$. Both cases give
    \[
        \log N\left(\delta\left(\frac{2\varepsilon}{\lambda}\right)^{1/2},\mathcal N_0,\|\cdot\|_\infty\right)
        \leq2A\delta^{-2p},\quad 0<\delta\leq1.
    \]
    By \eqref{eq:localized-boundedness}, every member of $\mathcal N_0$ has supremum norm at most $2\left(2\varepsilon/\lambda\right)^{1/2}$. For $1<\delta\leq2$, the $\left(2\varepsilon/\lambda\right)^{1/2}$-cover gives the same estimate with $2A$ replaced by $4A$. Inverting this covering bound and using $\|\cdot\|_{L_2(D)}\leq\|\cdot\|_\infty$ yields, for a numerical constant $C_0\geq1$,
    \[
        e_i(\mathcal N_0,L_2(D))
        \leq\left(\frac{2\varepsilon}{\lambda}\right)^{1/2}(C_0A)^{1/(2p)}i^{-1/(2p)},\quad i\geq1.
    \]

    Apply \cite[Theorem 7.16]{Steinwart2008Support} to the finite class $\mathcal N_0$, using \eqref{eq:localized-boundedness} and \eqref{eq:localized-variance}. As in \eqref{eq:localized-rademacher}, this gives
    \[
        \mathbb E_{D\sim P^n}\operatorname{Rad}_D(\mathcal N_0)
        \leq C\left[
        \varepsilon^{(3-p)/4}\lambda^{-(1+p)/4}\left(\frac{A}{n}\right)^{1/2}
        +\varepsilon^{1/2}\lambda^{-1/2}\left(\frac{A}{n}\right)^{1/(1+p)}\right],
    \]
    with $C$ uniform for $p\in(0,1/2]$. For every sample set $D$, \eqref{eq:finite-loss-net} also implies
    \[
        \operatorname{Rad}_D(\mathcal Q_{\lambda,\varepsilon})
        \leq\operatorname{Rad}_D(\mathcal N_0)+2b\left(\frac{2\varepsilon}{\lambda}\right)^{1/2}.
    \]
    Consequently, the estimate \eqref{eq:localized-rademacher} remains valid with an additional term $Cb(\varepsilon/\lambda)^{1/2}$ on the right-hand side. Then, applying symmetrization and the simplified Talagrand's inequality and following the argument in the proof of Theorem~\ref{thm:oracle-inequality} yields \eqref{eq:truncated-entropy-oracle}, with the additional term contributing $Cb^2/\lambda$ to the bound.
\end{proof}

\begin{proof}[Proof of Theorem~\ref{thm:logistic-besov-approximation}]
    For $P_X$-almost every $x$, define the conditional logistic risk by
    \[
        \ell_x(t):=\eta(x)\log(1+e^{-t})+(1-\eta(x))\log(1+e^t).
    \]
    Since $f^*(x)$ is its minimizer and is finite, we have
    \[
        \ell_x'(f^*(x))=0,\quad
        \ell_x''(t)=\frac{e^t}{(1+e^t)^2}\leq\frac14.
    \]
    Taylor's formula therefore gives a quadratic bound on the conditional excess risk. Combining it with the $1$-Lipschitz property of the logistic loss, for every $t\in\mathbb R$ and $1 \leq r\leq2$ we obtain
    \[
        \begin{aligned}
            0\leq\ell_x(t)-\ell_x(f^*(x))
            &\leq\min\left\{|t-f^*(x)|,\frac18|t-f^*(x)|^2\right\}\\
            &\leq8^{1-r}|t-f^*(x)|^r.
        \end{aligned}
    \]
    Let $p'$ be the conjugate exponent of $p$ and take $r:=2/p'=2(1-1/p)$. Then $1\leq p'\leq2$, $1\leq r\leq2$, and $rp'=2$. For any measurable $f$ with $f-f^*\in L_2(X)$, H\"older's inequality gives
    \begin{equation}\label{eq:lp-logistic-risk-bias}
        \begin{aligned}
            \mathcal R(f)-\mathcal R^*
            &\leq8^{1-r}\int_X|f-f^*|^r g\,dV\\
            &\leq8^{1-r}\|g\|_{L_p(X)}
            \left(\int_X|f-f^*|^{rp'}\,dV\right)^{1/p'}\\
            &=8^{1-r}\|g\|_{L_p(X)}\|f-f^*\|_{L_2(X)}^r.
        \end{aligned}
    \end{equation}

    Take $s=\lfloor\alpha\rfloor+1$ and the approximant $S_{\gamma,s}f^*\in\mathcal H_\gamma(X)$. Applying Theorem~\ref{thm:besov-approximation} with $H^0(X) = L_2(X)$ norm in \eqref{eq:lp-logistic-risk-bias} yields
    \[
        \begin{aligned}
            \mathcal R(S_{\gamma,s}f^*)-\mathcal R^*
            &\leq8^{1-r}\|g\|_{L_p(X)}\|S_{\gamma,s}f^*-f^*\|_{L_2(X)}^r\\
            &\leq C_{\alpha,p,X,d}\gamma^{2\alpha(1-1/p)}\|g\|_{L_p(X)}
            \|f^*\|_{B^\alpha_{2,\infty}(X)}^{2(1-1/p)}.
        \end{aligned}
    \]
    Moreover, Theorem~\ref{thm:rkhs-norm} gives
    \[
        \lambda\|S_{\gamma,s}f^*\|_{\mathcal{H}_\gamma(X)}^2 \leq C_{\alpha,X,d}\lambda\gamma^{-\varrho} \|f^*\|_{L_2(X)}^2.
    \]
    Substitution into \eqref{eq:bandwidth-approximation-error} proves \eqref{eq:logistic-approximation-error}.
\end{proof}

\begin{proof}[Proof of Proposition~\ref{prop:logit-besov-regularity}]
    \emph{(i)} Set $\theta:=\alpha/\beta\in(0,1]$. For $a,b>0$, the inequality $\log(1+s)\leq C_\theta s^\theta$ for $s\geq0$ gives
    \begin{equation}\label{eq:log-difference-inequality}
        |\log a-\log b| \leq C_\theta|a-b|^\theta(a^{-\theta}+b^{-\theta}).
    \end{equation}
    Condition \eqref{eq:logit-negative-moment} implies that $\eta$ takes values in $(0,1)$ outside a set of volume zero. At every $x \in X$ where $0<\eta(x)<1$,
    \[
        |f^*(x)|\leq-\log\bigl(\eta(x)(1-\eta(x))\bigr)
        \leq C_\theta[\eta(x)(1-\eta(x))]^{-\theta},
    \]
    so $f^*\in L_2(X)$. Applying \eqref{eq:log-difference-inequality} to $\eta$ and $1-\eta$, and using the H\"older regularity of $\eta$, yields
    \begin{equation}\label{eq:logit-pointwise-difference}
        \begin{aligned}
            &|f^*(x)-f^*(y)| \\
            \leq{}& C\|\eta\|_{C^{0,\beta}(X)}^\theta d_X(x,y)^\alpha\bigl([\eta(x)(1-\eta(x))]^{-\theta}+[\eta(y)(1-\eta(y))]^{-\theta}\bigr).
        \end{aligned}
    \end{equation}

    Use the fixed finite atlas $\{(U_\nu,\phi_\nu)\}_{\nu=1}^J$ and partition of unity $\{\chi_\nu\}_{\nu=1}^J$ fixed above. Write
    \[
        F_\nu:=(\chi_\nu f^*)\circ\phi_\nu^{-1},
    \]
    extended by zero to $\mathbb R^\varrho$. Fix a compact neighborhood of $\operatorname{supp}\chi_\nu$ inside $U_\nu$. For sufficiently small $\|h\|$, whenever either cutoff value at $u$ or $u+h$ is nonzero, both corresponding manifold points lie in this neighborhood. There, geodesic distances are bounded by a constant times coordinate distances, and the volume density is bounded above and below, see \eqref{eq:normal-coordinate-estimates}. For such $u$, write $x:=\phi_\nu^{-1}(u)$ and $y:=\phi_\nu^{-1}(u+h)$. Then
    \[
        \begin{aligned}
            F_\nu(u+h)-F_\nu(u)
            &=\chi_\nu(y)f^*(y)-\chi_\nu(x)f^*(x)\\
            &=\chi_\nu(y)\bigl(f^*(y)-f^*(x)\bigr)+\bigl(\chi_\nu(y)-\chi_\nu(x)\bigr)f^*(x).
        \end{aligned}
    \]
    We apply \eqref{eq:logit-pointwise-difference} to the first term and use $|\chi_\nu(y)-\chi_\nu(x)|\leq C\|h\|$ for the second. Taking $L_2(\mathbb R^\varrho)$ norms and using the coordinate change then gives
    \[
        \begin{aligned}
            \|F_\nu(\cdot+h)-F_\nu\|_{L_2(\mathbb R^\varrho)}
            &\leq C\|\eta\|_{C^{0,\beta}(X)}^\theta\|h\|^\alpha
            \left(\int_X[\eta(1-\eta)]^{-2\theta}\,dV\right)^{\frac{1}{2}}+C\|h\|\|f^*\|_{L_2(X)}.
        \end{aligned}
    \]
    Since $0<\alpha<1$, the second term is also bounded by a constant times $\|h\|^\alpha\|f^*\|_{L_2(X)}$ for $\|h\|\leq1$. Larger shifts are controlled by $2\|F_\nu\|_{L_2(\mathbb R^\varrho)}$. The Euclidean first-difference characterization in \cite[Theorem 2.6.1]{Triebel1992Theory} applies to each $F_\nu$ because $0<\alpha<1$. On the compact manifold $X$, the corresponding chartwise norm is equivalent to the interpolation norm in \eqref{eq:besov-interpolation} by interpolating the localization and reconstruction maps over a finite atlas, see \cite[proof of Theorem 7.2.3, Step 2, and Section 7.3.1]{Triebel1992Theory}. The preceding estimates therefore give \eqref{eq:logit-besov-bound}.

    \emph{(ii)} Under the assumptions of the second case, we first establish the gradient estimates
    \begin{equation}\label{eq:eta-gradient-endpoint}
        |\nabla\eta(x)|^2\leq C\eta(x), \quad|\nabla\eta(x)|^2\leq C(1-\eta(x)).
    \end{equation}
    To prove these estimates, let $M:=1+\|\nabla^2\eta\|_{L_\infty(X)}$. The estimates hold where the gradient vanishes. At a point where $\nabla\eta(x)\neq0$, take $v:=\nabla\eta(x)/|\nabla\eta(x)|$. Compactness makes $X$ geodesically complete, so Taylor's formula along the geodesic $t\mapsto\mathcal{E}_x(-tv)$ gives, for every $t\geq0$,
    \[
        0\leq\eta(\mathcal{E}_x(-tv))
        \leq\eta(x)-t|\nabla\eta(x)|+\frac{M}{2}t^2.
    \]
    Choosing $t=|\nabla\eta(x)|/M$ yields $|\nabla\eta(x)|^2\leq2M\eta(x)$. Applying the same argument to $1-\eta$ proves the second estimate in \eqref{eq:eta-gradient-endpoint}.

    To justify differentiation despite possible endpoint values of $\eta$, define
    \[
        f_\varepsilon:=\log(\eta+\varepsilon)-\log(1-\eta+\varepsilon), \quad 0<\varepsilon\leq1.
    \]
    These functions belong to $C^m(X)$. Use the fixed finite atlas $\{(U_\nu,\phi_\nu)\}_{\nu=1}^J$ and partition of unity $\{\chi_\nu\}_{\nu=1}^J$. For each $\nu$, choose an open neighborhood $V_\nu$ of $\operatorname{supp}\chi_\nu$ such that $V_\nu\Subset U_\nu$. For $u=(u_1,\ldots,u_\varrho)\in\phi_\nu(V_\nu)$ and a multi-index $a=(a_1,\ldots,a_\varrho)\in\mathbb N_0^\varrho$ with $1\leq|a|\leq m$, the chain rule expresses
    \[
        \partial_u^a\bigl[\log(\eta\circ\phi_\nu^{-1}+\varepsilon)\bigr](u)
    \]
    as a finite linear combination of terms of the form
    \[
        \begin{aligned}
            &\bigl(\eta(\phi_\nu^{-1}(u))+\varepsilon\bigr)^{-k}\prod_{i=1}^k\bigl[\partial_u^{b_i}(\eta\circ\phi_\nu^{-1})\bigr](u), \quad b_1+\cdots+b_k=a,\quad b_i\in\mathbb N_0^\varrho,\quad |b_i|\geq1.
        \end{aligned}
    \]
    If $\ell$ of the $k$ factors are exactly the first derivatives, then
    \[
        |a|=\sum_{i=1}^k|b_i|\geq\ell+2(k-\ell)=2k-\ell.
    \]
    By \eqref{eq:eta-gradient-endpoint} and the boundedness of the coordinate vector fields on $V_\nu$, each first coordinate derivative is bounded by $C\sqrt{\eta(\phi_\nu^{-1}(u))+\varepsilon}$, while the higher coordinate derivatives are uniformly bounded by $C\|\eta\|_{C^m(X)}$, with constants depending on the fixed chart. Multiplying these bounds, we find that the $\ell$ first derivative factors contribute a power $\ell/2$, while the bounds for the remaining factors are absorbed into $C$. Hence
    \[
        \begin{aligned}
            &\left|\bigl(\eta(\phi_\nu^{-1}(u))+\varepsilon\bigr)^{-k}\prod_{i=1}^k\bigl[\partial_u^{b_i}(\eta\circ\phi_\nu^{-1})\bigr](u)
            \right|\\
            \leq{}& C\bigl(\eta(\phi_\nu^{-1}(u))+\varepsilon\bigr)^{-k}\bigl(\eta(\phi_\nu^{-1}(u))+\varepsilon\bigr)^{\ell/2}\\
            ={}&C\bigl(\eta(\phi_\nu^{-1}(u))+\varepsilon\bigr)^{-k+\ell/2}\\
            ={}&C\bigl(\eta(\phi_\nu^{-1}(u))+\varepsilon\bigr)^{-|a|/2}\bigl(\eta(\phi_\nu^{-1}(u))+\varepsilon\bigr)^{|a|/2-k+\ell/2}.
        \end{aligned}
    \]
    Since $k-\ell/2\leq|a|/2$ and $0\leq\ell\leq k$, the exponent in the last factor satisfies
    \[
        0\leq\frac{|a|}{2}-k+\frac{\ell}{2}\leq\frac{|a|}{2}\leq\frac m2.
    \]
    Together with $0<\eta(\phi_\nu^{-1}(u))+\varepsilon\leq2$, this gives
    \[
        \bigl(\eta(\phi_\nu^{-1}(u))+\varepsilon\bigr)^{|a|/2-k+\ell/2}\leq2^{m/2}.
    \]
    Absorbing $2^{m/2}$ into $C$, we obtain the bound $C(\eta(\phi_\nu^{-1}(u))+\varepsilon)^{-|a|/2}$ for each term, uniformly in $\varepsilon$. Applying the same reasoning to $(1-\eta)\circ\phi_\nu^{-1}$ gives
    \[
        \begin{aligned}
            \bigl|\partial_u^a(f_\varepsilon\circ\phi_\nu^{-1})(u)\bigr|
            &\leq C\Bigl[
            \bigl(\eta(\phi_\nu^{-1}(u))+\varepsilon\bigr)^{-|a|/2} +\bigl(1-\eta(\phi_\nu^{-1}(u))+\varepsilon\bigr)^{-|a|/2}
            \Bigr]\\
            &\leq C\Bigl[
            \eta(\phi_\nu^{-1}(u))
            \bigl(1-\eta(\phi_\nu^{-1}(u))\bigr)
            \Bigr]^{-|a|/2}.
        \end{aligned}
    \]
    The constants are independent of $\varepsilon$ and can be chosen uniformly over the finitely many charts. Since the sets $V_\nu$ cover $X$, the relation between coordinate and covariant derivatives therefore yields
    \begin{equation}\label{eq:logit-regularized-derivatives}
        |\nabla^{|a|} f_\varepsilon|
        \leq C[\eta(1-\eta)]^{-|a|/2},
        \quad 1\leq|a|\leq m.
    \end{equation}
    Also, adding the same positive number to $\eta$ and $1-\eta$ moves their ratio towards one, so
    \[
        |f_\varepsilon|\leq|f^*|
        \leq-\log(\eta(1-\eta))
        \leq C_m[\eta(1-\eta)]^{-m/2}.
    \]
    Since $\eta(1-\eta) \leq 1/4 < 1$, \eqref{eq:logit-regularized-derivatives} and \eqref{eq:logit-high-order-moment} give the uniform bound
    \[
        \|f_\varepsilon\|_{H^m(X)}
        \leq C\left(\int_X[\eta(1-\eta)]^{-m}\,dV\right)^{1/2}.
    \]
    Note that the endpoint set has zero volume. As $\varepsilon$ decreases to zero, $f_\varepsilon$ converges to $f^*$ almost everywhere. By the dominated convergence theorem, $f_\varepsilon \to f^*$ in $L_2(X)$. By weak compactness in $H^m(X)$ and the strong convergence in $L_2(X)$, we obtain $f^*\in H^m(X)$. Weak lower semicontinuity of the $H^m(X)$ norm then yields \eqref{eq:logit-sobolev-bound}. By \cite[Theorem 7.4.5 and Theorem 7.3.1]{Triebel1992Theory}, we have the identity $H^m(X)=B^m_{2,2}(X)$. The embeddings
    \[
        B^m_{2,2}(X)\hookrightarrow B^m_{2,\infty}(X)
        \hookrightarrow B^\alpha_{2,\infty}(X), \quad 0<\alpha\leq m,
    \]
    by \cite[Theorem 7.4.2]{Triebel1992Theory} give the Besov conclusion.

    Finally, we verify the two claims concerning \eqref{eq:volume-small-value}. Put $w(x):=\min\{\eta(x),1-\eta(x)\}$ and let $r>0$. The small value bound implies that $w>0$ almost everywhere. Since $\eta(1-\eta)=w(1-w)\geq w/2$, we have
    \[
        \begin{aligned}
            \int_X[\eta(1-\eta)]^{-r}\,dV
            &\leq2^r\int_Xw^{-r}\,dV\\
            &=2^r\left[2^rV(X)
            +r\int_0^{1/2}u^{-r-1}V\{w<u\}\,du\right]\\
            &\leq2^r\left[2^rV(X)
            +rC_0\int_0^{1/2}u^{\kappa-r-1}\,du\right]<\infty
        \end{aligned}
    \]
    whenever $\kappa>r$. Taking $r=2\alpha/\beta$ in (i) and $r=m$ in (ii) proves the respective assertions.
\end{proof}

\begin{proof}[Proof of Proposition~\ref{prop:manifold-entropy}]
    Applying \cite[Lemma 5.4]{Yang2016Bayesian} at the fixed smoothness order $m$ gives \eqref{eq:manifold-covering-entropy} for sufficiently small $\gamma>0$ and $\gamma^{m-1}<\delta<1/2$. Choose $\gamma_m\in(0,1)$ so that this estimate holds for every $0<\gamma\leq\gamma_m$ and $\gamma_m^{m-1}<1/4$. For $1/4\leq\delta\leq1$, monotonicity of covering numbers and the bound at $\delta=1/4$ give the same estimate after increasing $C_m$.

    For the polynomial estimate, observe that, for $0<q\leq1/2$,
    \[
        \begin{aligned}
            \delta^{2q}\left(\log\frac2\delta\right)^{\varrho+1}
            &=2^{2q}\exp\left(-2q\log\frac2\delta\right)\left(\log\frac2\delta\right)^{\varrho+1}\\
            &\leq2\sup_{t>0}\exp(-2qt)t^{\varrho+1}\\
            &=2\left(\frac{\varrho+1}{2eq}\right)^{\varrho+1}.
        \end{aligned}
    \]
    Substitution into \eqref{eq:manifold-covering-entropy} proves \eqref{eq:manifold-polynomial-covering} with an enlarged constant $C_m$.
\end{proof}

\begin{proof}[Proof of Theorem~\ref{thm:besov-learning-rate}]
    Fix an integer $m\geq2$ such that
    \[
        m-1>\beta+\frac{\varrho}{2}.
    \]
    Let $C_m$ and $\gamma_m$ be as in Proposition~\ref{prop:manifold-entropy}. For all sufficiently large $n$, we have $0<\gamma_n\leq\gamma_m$, $0<\lambda_n\leq1$, and
    \[
        q_n:=\frac{\log2}{2\log n}\in(0,1/2],\quad
        b_n:=2\gamma_n^{m-1}\leq\frac14,\quad
        A_n:=C_m q_n^{-(\varrho+1)}\gamma_n^{-\varrho}\geq1.
    \]
    The estimate \eqref{eq:manifold-polynomial-covering} verifies \eqref{eq:truncated-covering-assumption} for $\mathcal F=\mathcal H_{\gamma_n}(X)$ with entropy exponent $q_n$, $A=A_n$, and $b=b_n$. Apply Corollary~\ref{cor:truncated-entropy-oracle} and Theorem~\ref{thm:logistic-besov-approximation}. Then with probability at least $1-e^{-t}$, the excess risk is bounded, up to constants independent of $n$, by
    \[
        \gamma_n^\beta+\lambda_n\gamma_n^{-\varrho} +\lambda_n^{-1}\left(\frac{A_n}{n}\right)^{2/(1+q_n)} +\lambda_n+\lambda_n^{-1}b_n^2 +\lambda_n^{-1}\frac{t^2}{n^2}.
    \]
    The choice of $q_n$ gives
    \[
        \begin{aligned}
            \frac{A_n}{n}&\asymp\frac{(\log n)^{\varrho+1}}{n}\gamma_n^{-\varrho}
            =\left(\frac{(\log n)^{\varrho+1}}{n}\right)^{(2\beta+\varrho)/(2\beta+3\varrho)},\\
            \left(\frac{A_n}{n}\right)^{2/(1+q_n)}
            &=\left(\frac{A_n}{n}\right)^2\exp\left(\frac{2q_n}{1+q_n}\log\frac{n}{A_n}\right)
            \asymp \left(\frac{A_n}{n}\right)^2.
        \end{aligned}
    \]
    It follows that the first three terms have the same order:
    \[
        \gamma_n^\beta \asymp\lambda_n\gamma_n^{-\varrho} \asymp\lambda_n^{-1}\left(\frac{A_n}{n}\right)^{2/(1+q_n)} \asymp\left(\frac{(\log n)^{\varrho+1}}{n}\right)^{2\beta/(2\beta+3\varrho)}.
    \]
    To control the error from the finite covering scale, note that $\lambda_n=\gamma_n^{\beta+\varrho}$ and hence
    \[
        \begin{aligned}
            \frac{\lambda_n^{-1}b_n^2}{\gamma_n^\beta}
            &=4\gamma_n^{2(m-1)-2\beta-\varrho}\to 0,
        \end{aligned}
    \]
    by the choice of $m$. The remaining terms satisfy
    \[
        \begin{aligned}
            \frac{\lambda_n}{\gamma_n^\beta}
            &=\gamma_n^\varrho\to 0,\\
            \lambda_n^{-1}\frac{t^2}{n^2}
            &\leq t^2\left(\frac{(\log n)^{\varrho+1}}{n}\right)^{(2\beta+4\varrho)/(2\beta+3\varrho)}
            =o\left(\left(\frac{(\log n)^{\varrho+1}}{n}\right)^{2\beta/(2\beta+3\varrho)}\right)
        \end{aligned}
    \]
    for fixed $t$. Since $2\beta=4\alpha(1-1/p)$, this proves \eqref{eq:besov-learning-rate}.
\end{proof}

\begin{proof}[Proof of Theorem~\ref{thm:clipped-learning-rate}]
    We first establish a comparison bound for clipped predictions. For every $u\in\mathbb R$, $y\in\{-1,1\}$, and $M>0$,
    \begin{equation}\label{eq:clipping-loss-increase}
        \phi(y\pi_M(u))\leq\phi(yu)+e^{-M}.
    \end{equation}
    Indeed, clipping decreases the loss when $yu<-M$ and leaves it unchanged when $|u|\leq M$. When $yu>M$, the increase is at most $\phi(M)\leq e^{-M}$.

    Fix $0<\lambda\leq1$, $\gamma>0$, $M\geq1$, and a deterministic comparison function $f_0\in\mathcal H_\gamma(X)$. By the 1-Lipschitz property of $\phi$,
    \[
        0\leq\bigl(\phi(Yf_0(X))-\phi(Y\pi_M f_0(X))\bigr)_+
        \leq\|f_0\|_\infty.
    \]
    Moreover, \eqref{eq:clipping-loss-increase} bounds the negative part of this loss difference by $e^{-M}$. Hence
    \[
        \begin{aligned}
            \mathbb E_P\bigl(\phi(Yf_0(X))-\phi(Y\pi_M f_0(X))\bigr)_+
            &\leq\mathcal R(f_0)-\mathcal R^*+e^{-M},\\
            \mathbb E_P\bigl(\phi(Yf_0(X))-\phi(Y\pi_M f_0(X))\bigr)_+^2
            &\leq\|f_0\|_\infty\bigl(\mathcal R(f_0)-\mathcal R^*+e^{-M}\bigr).
        \end{aligned}
    \]
    Bernstein's inequality shows that, with probability at least $1-e^{-t}/2$,
    \begin{equation}\label{eq:clipped-comparator-concentration}
        \begin{aligned}
            &\mathbb E_D\bigl(\phi(Yf_0(X))-\phi(Y\pi_M f_0(X))\bigr)_+\\
            \leq{}& 2\bigl(\mathcal R(f_0)-\mathcal R^*+e^{-M}\bigr)+\frac{C\|f_0\|_\infty(t+\log2)}{n}.
        \end{aligned}
    \end{equation}
    \eqref{eq:clipping-loss-increase} implies
    \begin{equation}\label{eq:clipped-empirical-comparison}
        \begin{aligned}
            &\mathbb E_D\bigl(\phi(Y\pi_M\widehat f_{n,\lambda,\gamma}(X))-\phi(Y\pi_M f_0(X))\bigr)\\
            \leq{}&\mathbb E_D\bigl(\phi(Y\widehat f_{n,\lambda,\gamma}(X))-\phi(Yf_0(X))\bigr)\\
            &\quad +\mathbb E_D\bigl(\phi(Yf_0(X))-\phi(Y\pi_M f_0(X))\bigr)+e^{-M}\\
            \leq{}&\lambda\|f_0\|_{\mathcal H_\gamma(X)}^2+\mathbb E_D\bigl(\phi(Yf_0(X))-\phi(Y\pi_M f_0(X))\bigr)_++e^{-M}.
        \end{aligned}
    \end{equation}

    Comparison of the empirical objective with zero gives $\|\widehat f_{n,\lambda,\gamma}\|_{\mathcal H_\gamma(X)}\leq\lambda^{-1/2}$. Put
    \[
        \begin{aligned}
            \mathcal V&=\{\pi_M f:f\in\lambda^{-1/2}B_{\mathcal H_\gamma(X)}\}.
        \end{aligned}
    \]
    Suppose that $\mathcal V$ has a finite $\delta$-net $\{v_1,\ldots,v_N\}\subset\mathcal V$ in the supremum norm. For each net member, both $v_j$ and $\pi_M f_0$ are bounded by $M$, so their loss difference is bounded in absolute value by $2M$. By Lemma~\ref{lem:logistic-variance},
    \[
        \begin{aligned}
            &\mathbb E_P\bigl(\phi(Yv_j(X))-\phi(Y\pi_M f_0(X))\bigr)^2\\
            \leq{}&2\mathbb E_P\bigl(\phi(Yv_j(X))-\psi(X,Y)\bigr)^2
                +2\mathbb E_P\bigl(\phi(Y\pi_M f_0(X))-\psi(X,Y)\bigr)^2\\
            \leq{}&2(M+2)\bigl(\mathcal R(v_j)-\mathcal R^*+\mathcal R(\pi_M f_0)-\mathcal R^*\bigr).
        \end{aligned}
    \]
    Apply Bernstein's inequality to these $N$ loss differences and use a union bound with parameter $\log N+t+\log2$. Then, with probability at least $1-e^{-t}/2$, simultaneously for every $j$,
    \[
        \begin{aligned}
            &\mathcal R(v_j)-\mathcal R(\pi_M f_0) \\
            \leq{}&\mathbb E_D\bigl(\phi(Yv_j(X))-\phi(Y\pi_M f_0(X))\bigr)\\
            &\quad+\frac12\bigl(\mathcal R(v_j)-\mathcal R^*+\mathcal R(\pi_M f_0)-\mathcal R^*\bigr)  +\frac{C(M+2)(\log N+t+\log2)}{n}.
        \end{aligned}
    \]
    For each $w\in\mathcal V$, choose a net member with $\|w-v_j\|_\infty\leq\delta$. The Lipschitz property of the loss controls both the population and empirical loss differences by $\delta$. Rearranging the preceding estimate therefore yields, on the same event,
    \[
        \begin{aligned}
            \mathcal R(w)-\mathcal R^*
            &\leq2\mathbb E_D\bigl(\phi(Yw(X))-\phi(Y\pi_M f_0(X))\bigr)+3\bigl(\mathcal R(\pi_M f_0)-\mathcal R^*\bigr)\\
            &\quad+\frac{C(M+2)(\log N+t+\log2)}{n}+3\delta.
        \end{aligned}
    \]
    In particular, this holds for $w=\pi_M\widehat f_{n,\lambda,\gamma}$. Since $\mathcal R(\pi_M f_0)-\mathcal R^*\leq\mathcal R(f_0)-\mathcal R^*+e^{-M}$ by \eqref{eq:clipping-loss-increase}, combining the two events with \eqref{eq:clipped-comparator-concentration} and \eqref{eq:clipped-empirical-comparison} proves that, with probability at least $1-e^{-t}$,
    \begin{equation}\label{eq:clipped-comparison-bound}
        \begin{aligned}
            \mathcal R(\pi_M\widehat f_{n,\lambda,\gamma})-\mathcal R^* &\leq C\bigg[\mathcal R(f_0)-\mathcal R^*
            +\lambda\|f_0\|_{\mathcal H_\gamma(X)}^2 +\frac{(M+2)(\log N+t+\log2)}{n}\\
            &\qquad+\frac{\|f_0\|_\infty(t+\log2)}{n}+e^{-M}+\delta\bigg].
        \end{aligned}
    \end{equation}

    We now apply this bound with the parameters in the theorem and choose
    \[
        f_0=S_{\gamma_n,s}f^*,\quad s=\lfloor\alpha\rfloor+1, \quad \delta=\gamma_n^\beta.
    \]
    Theorem~\ref{thm:rkhs-norm} and the proof of Theorem~\ref{thm:logistic-besov-approximation} give
    \begin{equation}\label{eq:clipped-comparator-bounds}
        \mathcal R(f_0)-\mathcal R^*\leq C\gamma_n^\beta, \quad \|f_0\|_{\mathcal H_{\gamma_n}(X)}^2\leq C\gamma_n^{-\varrho}, \quad \|f_0\|_\infty\leq\|f_0\|_{\mathcal H_{\gamma_n}(X)}\leq C\gamma_n^{-\varrho/2}.
    \end{equation}
    Fix an integer $m\geq2$ such that
    \[
        m-1>\frac{3\beta+\varrho}{2}.
    \]
    Since $\lambda_n=\gamma_n^{\beta+\varrho}$, the normalized covering radius satisfies
    \[
        \begin{aligned}
            \frac{\delta\sqrt{\lambda_n}}{2}
            &=\frac12\gamma_n^{(3\beta+\varrho)/2}
            >\gamma_n^{m-1}
        \end{aligned}
    \]
    for all sufficiently large $n$. It is also at most one, and $\gamma_n\leq\gamma_m$ eventually. Thus Proposition~\ref{prop:manifold-entropy} applies at this radius. Clipping does not increase supremum distances. Covering $\lambda_n^{-1/2}B_{\mathcal H_{\gamma_n}(X)}$ at radius $\delta/2$, clipping the centers, and choosing one member of $\mathcal V$ from each nonempty covering ball produces a $\delta$-net in $\mathcal V$, with
    \[
        \log N\leq C_m\gamma_n^{-\varrho}\left(\log\frac{4}{\delta\sqrt{\lambda_n}}\right)^{\varrho+1}\leq C\gamma_n^{-\varrho}(\log n)^{\varrho+1}.
    \]
    The constants are independent of $n$.

    Substituting this estimate and \eqref{eq:clipped-comparator-bounds} into \eqref{eq:clipped-comparison-bound}, and using $M_n=3\log n$, gives
    \[
        \begin{aligned}
            &\mathcal R(\pi_{M_n}\widehat f_{n,\lambda_n,\gamma_n})-\mathcal R^* \\
            \leq{}& C\bigg[\gamma_n^\beta+\lambda_n\gamma_n^{-\varrho}+\frac{\gamma_n^{-\varrho}(\log n)^{\varrho+2}}{n}+\frac{(\log n+\gamma_n^{-\varrho/2})(1+t)}{n}+n^{-3}\bigg].
        \end{aligned}
    \]
    The parameter choices imply that the main terms are
    \[
        \gamma_n^\beta = \lambda_n\gamma_n^{-\varrho}=\frac{\gamma_n^{-\varrho}(\log n)^{\varrho+2}}{n}=\left(\frac{(\log n)^{\varrho+2}}{n}\right)^{\beta/(\beta+\varrho)}.
    \]
    The remaining terms are of smaller order for fixed $t$. Absorbing the factor $1+t$ into $C$ for fixed $t$ proves the logistic bound in \eqref{eq:clipped-learning-rate}. For the misclassification bound, note that
    \[
        \operatorname{sgn}(\pi_{M_n}\widehat f_{n,\lambda_n,\gamma_n})=\operatorname{sgn}(\widehat f_{n,\lambda_n,\gamma_n}).
    \]
    The calibration inequality \eqref{eq:logistic-calibration-weak} then gives the misclassification bound on the same event.
\end{proof}

\subsection{Proofs for Section~\ref{sec:elliptic-operator-learning}}

\begin{proof}[Proof of Lemma~\ref{lem:pde-centered-consistency}]
    Apply Theorem~\ref{thm:gaussian-sobolev-expansion} at order $N$ to both $u$ and $\mathbf1$. Let $R_\gamma$ denote the corresponding remainder operator, then we have
    \[
        C_\gamma u=\sum_{r=0}^N \gamma^{2r}
        \bigl(A_ru-uA_r\mathbf1\bigr)+R_\gamma u-uR_\gamma\mathbf1 = \sum_{r=1}^N \gamma^{2r}
        \bigl(A_ru-uA_r\mathbf1\bigr)+R_\gamma u-uR_\gamma\mathbf1,
    \]
    and
    \[
        \|R_\gamma u-uR_\gamma\mathbf1\|_{L_2(X)}
        \leq C_{N,X,d}\gamma^{2N+2}
        \bigl(\|u\|_{H^{2N+2}(X)}+\|u\|_\infty\bigr),\quad 0<\gamma\leq1.
    \]
    For the inequality, we have used H\"older inequality $\|uR_\gamma\mathbf1\|_{L_2(X)} \leq \|u\|_\infty \|R_\gamma\mathbf1\|_{L_2(X)}$. By Theorem~\ref{thm:gaussian-sobolev-expansion}, $A_1 = \Delta_X/4 + B_1$, and $B_1 \in \operatorname{Diff}^{0}(X)$ must be a multiplication operator. Applying $A_1$ to $u$ and $\mathbf1$ gives
    \[
        A_1u-uA_1\mathbf1 = \frac14\Delta_Xu + B_1u - uB_1\mathbf1 = \frac14\Delta_Xu.
    \]
    This gives
    \[
        C_\gamma u = \frac{\gamma^2}{4}\Delta_Xu + \sum_{r=2}^N \gamma^{2r}
        \bigl(A_ru-uA_r\mathbf1\bigr)+ R_\gamma u-uR_\gamma\mathbf1.
    \]
    Apply this expansion at each scale $j\gamma$, multiply by $a_j\gamma^{-2}$, and sum over $j$. \eqref{eq:pde-laplacian-moments} cancels all main terms except $\Delta_Xu$, and the remainders give \eqref{eq:pde-centered-bias}.

    Note that $J_{\gamma,N}(x,y)=J_{\gamma,N}(y,x)$. Interchanging $x$ and $y$ in the integral defining $\langle L_{\gamma,N}u,v\rangle_{L_2(X)}$ and averaging the two expressions proves \eqref{eq:pde-gaussian-energy}.
\end{proof}

\begin{proof}[Proof of Theorem~\ref{thm:pde-random-solution-operator}]
    Define
    \[
        \begin{aligned}
            S&:\mathbb R^J\to W, &Sc&=\sum_jc_j\varphi_j, \\
            M_{kl}&=\langle\varphi_k,\varphi_l\rangle_{L_2(X)},
            &K_{kl}&=\int_X\langle\nabla\varphi_k,\nabla\varphi_l\rangle\,dV,\\
            B&=\kappa M+K,
            &(b(f))_k&=\langle\varphi_k,f\rangle_{L_2(X)}.
        \end{aligned}
    \]
    By \eqref{eq:pde-coercivity} and the linear independence of the basis functions, for every $0 \neq c\in\mathbb R^J$,
    \[
        c^{\top}Bc
        =a_\kappa(Sc,Sc)>0.
    \]

    We first bound the error of the mean stiffness matrix $K_\gamma:=\mathbb E\widehat{K}$. By \eqref{eq:pde-gaussian-energy} and the independence of $X_i$ and $Y_i$, we have
    \[
        (K_\gamma)_{kl}
        =-\langle L_{\gamma,N}\varphi_l,\varphi_k\rangle_{L_2(X)}.
    \]
    Since $X$ has no boundary, integration by parts gives $K_{kl}=-\langle\Delta_X\varphi_l,\varphi_k\rangle_{L_2(X)}$. Apply Lemma~\ref{lem:pde-centered-consistency} to each of the finitely many basis functions to obtain
    \begin{equation}\label{eq:pde-matrix-bias}
        \|K_\gamma-K\|_{\mathrm{op}} \leq \|K_\gamma-K\|_{F} \leq C_W\gamma^{2N}.
    \end{equation}

    We next bound the variance of the empirical quantities. Each basis function admits a smooth extension to $\mathbb R^d$ by Whitney's extension theorem. Boundedness of the gradients of these extensions on the compact convex hull of $X$ gives, for a constant $L_W > 0$,
    \begin{equation}\label{eq:pde-basis-lipschitz}
        \|\Phi(x)-\Phi(y)\|_{\mathbb R^J}
        \leq L_W\|x-y\|_{\mathbb R^d},\quad x,y\in X.
    \end{equation}
    Recall the uniform volume bound used in the proof of Theorem~\ref{thm:rkhs-norm}:
    \[
        \sup_{y\in X}V\bigl(X\cap B_{\mathbb R^d}(y,r)\bigr)
        \leq C_Xr^\varrho,\quad 0<r\leq1,
    \]
    The same bound holds for $r>1$ after increasing $C_X$, since the intersection has volume at most $V(X)$. Fix $y\in X$ and decompose $X$ into annuli
    \[
        A_k(y):=\{x\in X:k\gamma\leq\|x-y\|_{\mathbb R^d}<(k+1)\gamma\},
        \quad k\in\mathbb N_0.
    \]
    On $A_k(y)$, we have
    \[
        \int_{A_k(y)}\exp\left(-\frac{2\|x-y\|_{\mathbb R^d}^2}{\gamma^2}\right)\|x-y\|_{\mathbb R^d}^4\,dV(x) \leq C_X\gamma^{\varrho+4}(k+1)^{\varrho+4}\exp(-2k^2).
    \]
    Summing it over $k$ yields
    \begin{equation}\label{eq:pde-gaussian-fourth-moment}
        \sup_{y\in X}\int_X \exp\left(-\frac{2\|x-y\|_{\mathbb R^d}^2}{\gamma^2}\right)\|x-y\|_{\mathbb R^d}^4\,dV(x) \leq C_X\gamma^{\varrho+4},\quad 0<\gamma\leq1.
    \end{equation}
    Let
    \[
        Z_\gamma(x,y):=
        \frac{J_{\gamma,N}(x,y)}{2q(x)q(y)}
        \bigl(\Phi(x)-\Phi(y)\bigr)
        \bigl(\Phi(x)-\Phi(y)\bigr)^{\top}.
    \]
    To bound a single stiffness matrix sample, put $r=\|x-y\|_{\mathbb R^d}$. Since $\|ww^\top\|_F=\|w\|^2$, the Lipschitz bound \eqref{eq:pde-basis-lipschitz} and $q\geq q_0$ give
    \[
        \begin{aligned}
            \|Z_\gamma(x,y)\|_F
            &\leq C_W\gamma^{-2}\sum_{j=1}^N
            \frac{|a_j|}{(j\gamma)^\varrho}\,r^2 e^{-r^2/(j\gamma)^2}\\
            &\leq C_{W,N}\gamma^{-\varrho},
        \end{aligned}
    \]
    where the last step uses $\sup_{r\geq0}r^2e^{-r^2/(j\gamma)^2}=(j\gamma)^2/e$.
    For the second moment, $(\sum_j z_j)^2\leq N\sum_j z_j^2$ and \eqref{eq:pde-gaussian-fourth-moment} give
    \[
        \begin{aligned}
            \mathbb E\|Z_\gamma(X_1,Y_1)\|_{F}^2
            &\leq C\gamma^{-4}\sum_{j=1}^N
            \frac{a_j^2}{(j\gamma\sqrt{\pi})^{2\varrho}}
            \int_X\int_X
            \exp\left(-\frac{2\|x-y\|_{\mathbb R^d}^2}{(j\gamma)^2}\right)
            \|x-y\|_{\mathbb R^d}^4\,dV(x)dV(y)\\
            &\leq C_{W,N}\gamma^{-\varrho}.
        \end{aligned}
    \]
    By \eqref{eq:pde-empirical-stiffness}, $\widehat K$ is the average of the matrices $Z_\gamma(X_i,Y_i)$, each with mean $K_\gamma$. Let $D_i:=Z_\gamma(X_i,Y_i)-K_\gamma$. These matrices are i.i.d. with mean zero. Writing $\langle\cdot,\cdot\rangle_F$ for the Frobenius inner product, independence gives $\mathbb E\langle D_i,D_j\rangle_F=\langle\mathbb E D_i,\mathbb E D_j\rangle_F=0$ whenever $i\ne j$. Expanding the squared norm of the average therefore gives
    \[
        \begin{aligned}
            \mathbb E\|\widehat K-K_\gamma\|_F^2
            &=\frac{1}{n^2}\sum_{i,j=1}^n\mathbb E\langle D_i,D_j\rangle_F\\
            &=\frac1n\mathbb E\|D_1\|_F^2\\
            &=\frac1n\left(\mathbb E\|Z_\gamma(X_1,Y_1)\|_F^2-\|K_\gamma\|_F^2\right).
        \end{aligned}
    \]
    Using the second moment bound above, we obtain
    \begin{equation}\label{eq:pde-stiffness-variance}
        \mathbb E\|\widehat{K}-K_\gamma\|_{F}^2
        \leq\frac{C_{W,N}}{n\gamma^\varrho}.
    \end{equation}
    Each mass matrix summand has Frobenius norm $q(X_i)^{-1}\|\Phi(X_i)\|^2\leq C_W$. The same variance calculation gives
    \begin{equation}\label{eq:pde-mass-variance}
        \mathbb E\|\widehat{M}-M\|_{F}^2
        \leq\frac{C_W}{n}.
    \end{equation}

    Let $\delta_x$ denote evaluation at $x$ and equip $H^{-s}(X)$ with the dual norm of $H^s(X)$ under the $L_2(X)$ inner product. The Sobolev embedding $H^s(X) \hookrightarrow C(X)$ gives
    \[
        \sup_{x\in X}\|\delta_x\|_{H^{-s}(X)} = \sup_{x\in X} \sup_{\|f\|_{H^s(X)}\leq1}|f(x)| \leq C_s.
    \]
    Define
    \[
        F_x:H^s(X)\longrightarrow\mathbb R^J, \quad F_xf=q(x)^{-1}\Phi(x)f(x).
    \]
    $F_x$ is a Hilbert--Schmidt operator with squared norm
    \[
        \|F_x\|_{\mathrm{HS}(H^s(X),\mathbb R^J)}^2
        =q(x)^{-2}\|\Phi(x)\|^2\|\delta_x\|_{H^{-s}(X)}^2.
    \]
    Choose $0<\theta<\min\{1,s-\varrho/2\}$. The Sobolev embedding $H^s(X)\hookrightarrow C^{0,\theta}(X)$ gives
    \[
        |f(x)-f(y)|\leq C\|f\|_{H^s(X)}d_X(x,y)^\theta,\quad x,y\in X.
    \]
    By the definition of the dual norm,
    \[
        \|\delta_x-\delta_y\|_{H^{-s}(X)} =\sup_{\|f\|_{H^s(X)}\leq1}|f(x)-f(y)| \leq C d_X(x,y)^\theta.
    \]
    Thus $x\mapsto\delta_x$ is norm-continuous in $H^{-s}(X)$. Since $\Phi$ is continuous and $q$ is measurable, $x\mapsto F_x$ is strongly measurable. The uniform bound on $\|\delta_x\|_{H^{-s}(X)}$, boundedness of $\Phi$, and $q\geq q_0$ give $\sup_{x\in X}\|F_x\|_{\mathrm{HS}(H^s(X),\mathbb R^J)}\leq C_{W,s}$ and
    \[
        \mathbb E\|F_{X_1}\|_{\mathrm{HS}(H^s(X),\mathbb R^J)}^2 =\int_X \frac{\|\Phi(x)\|_{\mathbb R^J}^2 \|\delta_x\|_{H^{-s}(X)}^2}{q(x)}\,dV(x)\leq C_{W,s}.
    \]
    In particular, the Bochner expectation of $F_{X_1}$ exists and is given by
    \[
        \begin{aligned}
            (\mathbb E F_{X_1})f
            &=\int_X q(x)^{-1}\Phi(x)f(x)q(x)\,dV(x)\\
            &=\int_X\Phi(x)f(x)\,dV(x)
            =b(f).
        \end{aligned}
    \]
    By \eqref{eq:pde-empirical-mass-load}, $\widehat b$ is the average of the $F_{X_i}$. The centered operators $F_{X_i}-b$ are independent and have mean zero, so the cross terms in the expected squared Hilbert--Schmidt norm vanish. Consequently,
    \begin{equation}\label{eq:pde-load-variance}
        \mathbb E\|\widehat b-b\|_{\mathrm{HS}(H^s(X),\mathbb R^J)}^2=\frac{1}{n}\mathbb E\|F_{X_1}-b\|_{\mathrm{HS}(H^s(X),\mathbb R^J)}^2\leq\frac{1}{n}\mathbb E\|F_{X_1}\|_{\mathrm{HS}(H^s(X),\mathbb R^J)}^2\leq\frac{C_{W,s}}{n}.
    \end{equation}
    For each sample pair $(X_i,Y_i)$, collect its three centered contributions in
    \[
        \xi_i:=\left(Z_\gamma(X_i,Y_i)-K_\gamma,\,
        \frac{\Phi(X_i)\Phi(X_i)^\top}{q(X_i)}-M,\,F_{X_i}-b\right).
    \]
    Equip the two matrix spaces with their Frobenius norms and the load space with its Hilbert--Schmidt norm. The $\xi_i$ are i.i.d. because the sample pairs are independent. Since $\gamma\leq1$, the single-sample bounds above give $\|\xi_i\|\leq C_{W,N,s}\gamma^{-\varrho}$. \eqref{eq:pde-stiffness-variance}, \eqref{eq:pde-mass-variance}, and \eqref{eq:pde-load-variance} with $n=1$ give $\mathbb E\|\xi_i\|^2\leq C_{W,N,s}\gamma^{-\varrho}$. For each integer $p\geq2$, it follows that
    \[
        \mathbb E\|\xi_i\|^p
        \leq (C_{W,N,s}\gamma^{-\varrho})^{p-2}\mathbb E\|\xi_i\|^2.
    \]
    Applying the Hilbert space Bernstein's inequality to $n^{-1}\sum_{i=1}^n\xi_i$ gives an event $\mathcal E$ of probability at least $1-\delta$ on which
    \begin{equation}\label{eq:pde-joint-quadrature-event}
        \|\widehat{K}-K_\gamma\|_{\mathrm{op}}+\|\widehat{M}-M\|_{\mathrm{op}}+\|\widehat b-b\|_{H^s(X)\to\mathbb R^J}\leq C_{W,N,s}\left(\sqrt{\frac{\log(2/\delta)}{n\gamma^\varrho}}+\frac{\log(2/\delta)}{n\gamma^\varrho}\right).
    \end{equation}
    Here each operator norm is bounded by the corresponding Frobenius or Hilbert--Schmidt norm.

    Choose the constant $c$ in the theorem at most one. Under the hypothesis $\varepsilon_{n,\gamma,\delta}\leq c$, the square-root term in \eqref{eq:pde-joint-quadrature-event} is at most one, so it also bounds the linear term. On $\mathcal E$, combining \eqref{eq:pde-matrix-bias} and \eqref{eq:pde-joint-quadrature-event} gives
    \begin{equation}\label{eq:pde-matrix-perturbation}
        \|\widehat{B}-B\|_{\mathrm{op}} \leq C\varepsilon_{n,\gamma,\delta}.
    \end{equation}
    Choose $c$ in the theorem small enough such that $C\varepsilon_{n,\gamma,\delta} < \lambda_{\min}(B)/2$. Then
    \begin{equation}\label{eq:pde-discrete-stability}
        \lambda_{\min}(\widehat{B})\geq\lambda_{\min}(B)/2 > 0, \quad \|\widehat{B}^{-1}\|_{\mathrm{op}}\leq2/\lambda_{\min}(B).
    \end{equation}
    Hence $\widehat{B}$ is positive definite and $\widehat G_{n,\gamma,W} = S\widehat{B}^{-1}\widehat b$.

    It remains to bound the approximation error. Define the exact Galerkin operator $G_W:=SB^{-1}b$. Then
    \[
        \widehat{B}^{-1}\widehat b-B^{-1}b
        =\widehat{B}^{-1}
        \left[(\widehat b-b)+(B-\widehat{B})B^{-1}b\right].
    \]
    Using the load error bound in \eqref{eq:pde-joint-quadrature-event}, the matrix perturbation bound \eqref{eq:pde-matrix-perturbation}, and the inverse bound in \eqref{eq:pde-discrete-stability}, we obtain
    \begin{equation}\label{eq:pde-random-galerkin-perturbation}
        \|\widehat G_{n,\gamma,W}-G_W\|_{H^s(X)\to H^1(X)}
        \leq C\varepsilon_{n,\gamma,\delta}.
    \end{equation}
    Here we also use boundedness of $S$ and $B^{-1}b$. Finally, $u=Gf$ and $u_W=G_Wf$ satisfy Galerkin orthogonality $a_\kappa(u-u_W,w)=0$ for every $w\in W$.
    For $v\in W$, \eqref{eq:pde-coercivity} consequently gives
    \[
        \begin{aligned}
            c_\kappa\|u-u_W\|_{H^1(X)}^2
            &\leq a_\kappa(u-u_W,u-u_W)\\
            &=a_\kappa(u-u_W,u-v)\\
            &\leq C_\kappa\|u-u_W\|_{H^1(X)}\|u-v\|_{H^1(X)}.
        \end{aligned}
    \]
    Taking the infimum over $v$ and the supremum over $f$ with $\|f\|_{H^s(X)}\leq1$ yields
    \[
        \|G-G_W\|_{H^s(X)\to H^1(X)}\leq(C_\kappa/c_\kappa)\eta_W.
    \]
    Combining this estimate with \eqref{eq:pde-random-galerkin-perturbation} proves \eqref{eq:pde-random-operator-bound} and the proof is complete.
\end{proof}

\begin{proof}[Proof of Corollary~\ref{cor:pde-operator-rate}]
    With the stated bandwidth,
    \[
        \gamma^{2N}=\sqrt{\frac{\log(2/\delta)}{n\gamma^\varrho}}
        =\left(\frac{\log(2/\delta)}{n}\right)^{\frac{2N}{4N+\varrho}}.
    \]
    For sufficiently large $n/\log(2/\delta)$, we have $\gamma\leq1/N$ and $\varepsilon_{n,\gamma,\delta}\leq c$, so Theorem~\ref{thm:pde-random-solution-operator} yields \eqref{eq:pde-operator-rate}. Applying this operator norm bound to each $f$, squaring, and integrating with respect to $\mu$ gives \eqref{eq:pde-new-source-risk}.
\end{proof}

\begin{proof}[Proof of Corollary~\ref{cor:pde-besov-trial-bound}]
    Since $G=(\kappa I-\Delta_X)^{-1}$ is a function of $\Delta_X$, its spectral multiplier on the range of $P_j$ is bounded by $C_\kappa'2^{-2j}$ for a constant $C_\kappa'$ depending only on $\kappa$. Thus \eqref{eq:besov-spectral} gives
    \begin{equation}\label{eq:pde-besov-regularity}
        \|Gf\|_{B^{\alpha+2}_{2,\infty}(X)}
        \leq C\|f\|_{B^\alpha_{2,\infty}(X)}.
    \end{equation}

    For $u=Gf$, set $\ell=\lfloor\alpha+2\rfloor+1$ and $g_{\sigma}=S_{\sigma,\ell}u \in \mathcal H_{\sigma}(X)$. Let $P_{\sigma,\Xi}g_{\sigma}$ be its $\mathcal H_{\sigma}(X)$-orthogonal projection onto $W_{\sigma,\Xi}$. The reproducing property implies that the residual $e:=g_{\sigma}-P_{\sigma,\Xi}g_{\sigma}$ vanishes on $\Xi$. The manifold interpolation error estimate \cite[Lemma 10]{Fuselier2012Scattered} gives
    \begin{equation}\label{eq:pde-manifold-interpolation}
        \|e\|_{H^1(X)}\leq Ch^{s-1}\|e\|_{H^s(X)}.
    \end{equation}
    For $w\in\mathcal H_\sigma(X)$, let $\widetilde w\in\mathcal H_\sigma(\mathbb R^d)$ be its minimal-norm extension \cite[Theorem 4.21]{Steinwart2008Support}. By the Gaussian integral representation \cite[(4.43)]{Steinwart2008Support}, we can choose $a\in L_2(\mathbb R^d)$ with $\|a\|_{L_2(\mathbb R^d)}=\|\widetilde w\|_{\mathcal H_\sigma(\mathbb R^d)}$ such that
    \[
        \widetilde w(x)=\frac{2^{d/2}}{\pi^{d/4}\sigma^{d/2}}
        \int_{\mathbb R^d} \exp\left(-\frac{2\|x-y\|^2}{\sigma^2}\right)a(y)\,dy.
    \]
    As in the proof of \cite[Theorem 4.48]{Steinwart2008Support}, we differentiate under the integral and apply Cauchy--Schwarz in $y$ at fixed $x$. For a multi-index $\nu$, differentiation contributes $\sigma^{-|\nu|}$, and the change of variables $z=(x-y)/\sigma$ contributes $\sigma^{d/2}$ to the $L_2(\mathbb R^d)$ norm of the differentiated Gaussian in $y$. This gives
    \[
        |D^\nu\widetilde w(x)|\leq C_{\nu,d}\sigma^{-|\nu|}
        \|\widetilde w\|_{\mathcal H_\sigma(\mathbb R^d)},\quad x\in\mathbb R^d.
    \]
    For $0<\sigma\leq1$, the chain rule in local coordinates, integration over $X$, and this pointwise estimate yield
    \begin{equation}\label{eq:pde-rkhs-sobolev-embedding}
        \begin{aligned}
            \|w\|_{H^s(X)}
            &\leq C\sum_{|\nu|\leq s}\sup_{x\in X}|D^\nu\widetilde w(x)|\\
            &\leq C\sigma^{-s}\|\widetilde w\|_{\mathcal H_\sigma(\mathbb R^d)}\\
            &= C\sigma^{-s}\|w\|_{\mathcal H_\sigma(X)}.
        \end{aligned}
    \end{equation}

    By orthogonality, $\|e\|_{\mathcal H_\sigma(X)}\leq\|g_{\sigma}\|_{\mathcal H_\sigma(X)}$. Combining this bound with \eqref{eq:pde-manifold-interpolation} and \eqref{eq:pde-rkhs-sobolev-embedding}, and then applying Theorem~\ref{thm:rkhs-norm}, gives
    \[
        \begin{aligned}
            \|e\|_{H^1(X)}
            &\leq Ch^{s-1}\sigma^{-s}\|g_{\sigma}\|_{\mathcal H_{\sigma}(X)}\\
            &\leq Ch^{s-1}\sigma^{-s-\varrho/2}\|u\|_{L_2(X)}.
        \end{aligned}
    \]
    The triangle inequality, Theorem~\ref{thm:besov-approximation} with target norm $H^1(X)$ and smoothness $\alpha+2$, and the preceding estimate for $e$ give
    \[
        \begin{aligned}
            \inf_{v\in W_{\sigma,\Xi}}\|Gf-v\|_{H^1(X)}
            &\leq \|u-P_{\sigma,\Xi}g_{\sigma}\|_{H^1(X)}\\
            &\leq \|u-g_{\sigma}\|_{H^1(X)}+\|e\|_{H^1(X)}\\
            &\leq C\sigma^{\alpha+1}\|u\|_{B^{\alpha+2}_{2,\infty}(X)}
                 +Ch^{s-1}\sigma^{-s-\varrho/2}\|u\|_{L_2(X)}.
        \end{aligned}
    \]
    The regularity estimate \eqref{eq:pde-besov-regularity} for $u=Gf$ and the embedding $B^{\alpha+2}_{2,\infty}(X)\hookrightarrow L_2(X)$ bound both norms of $u$ by $C\|f\|_{B^\alpha_{2,\infty}(X)}$. Hence the right-hand side is bounded by $C(\sigma^{\alpha+1}+h^{s-1}\sigma^{-s-\varrho/2})\|f\|_{B^\alpha_{2,\infty}(X)}$. Taking the supremum over $\|f\|_{B^\alpha_{2,\infty}(X)}\leq1$ gives \eqref{eq:pde-besov-trial-bound}.

    To prove \eqref{eq:pde-besov-operator-bound}, fix any $t$ with $\varrho/2<t<\alpha$. Then $B^\alpha_{2,\infty}(X)\hookrightarrow H^t(X)$. The lower bound $t>\varrho/2$ gives $H^t(X)\hookrightarrow C^{0,\theta}(X)$ for $0<\theta<\min\{1,t-\varrho/2\}$. As in the proof of \eqref{eq:pde-load-variance}, this embedding gives strong measurability of $F_x$ and a uniform bound on $\|F_x\|_{\mathrm{HS}(H^t(X),\mathbb R^J)}$. Applying the same joint Bernstein bound \eqref{eq:pde-joint-quadrature-event} with this load space gives \eqref{eq:pde-random-galerkin-perturbation} with $H^t(X)$ as the source space. On the resulting event of probability at least $1-\delta$, every $f\in B^\alpha_{2,\infty}(X)$ satisfies
    \[
        \begin{aligned}
            \|(\widehat G_{n,\gamma,W}-G_W)f\|_{H^1(X)}
            &\leq C_{W,t}\varepsilon_{n,\gamma,\delta}\|f\|_{H^t(X)}\\
            &\leq C_W\varepsilon_{n,\gamma,\delta}\|f\|_{B^\alpha_{2,\infty}(X)}.
        \end{aligned}
    \]
    Galerkin quasi-optimality bounds $\|G_Wf-Gf\|_{H^1(X)}$ by $C\inf_{v\in W}\|Gf-v\|_{H^1(X)}$. Adding the two bounds and taking the supremum over the Besov unit ball gives \eqref{eq:pde-besov-operator-bound} by \eqref{eq:pde-besov-trial-bound}.
\end{proof}

\end{appendix}

\bibliographystyle{plain}
\bibliography{references}

@InProceedings{Suzuki2019Adaptivity,
  author    = {Suzuki, Taiji},
  booktitle = {International {{Conference}} on {{Learning Representations}}},
  title     = {Adaptivity of Deep {{ReLU}} Network for Learning in {{Besov}} and Mixed Smooth {{Besov}} Spaces: Optimal Rate and Curse of Dimensionality},
  year      = {2019},
}

@Book{Triebel1992Theory,
  author    = {Triebel, Hans},
  publisher = {Springer Basel},
  title     = {Theory of {{Function Spaces II}}},
  year      = {1992},
  address   = {Basel},
  doi       = {10.1007/978-3-0346-0419-2},
}

@Article{Audibert2007Fast,
  author  = {Audibert, Jean-Yves and Tsybakov, Alexandre B.},
  journal = {Ann. Stat.},
  title   = {Fast Learning Rates for Plug-in Classifiers},
  year    = {2007},
  number  = {2},
  volume  = {35},
  doi     = {10.1214/009053606000001217},
}

@Article{Zhang2004Statistical,
  author  = {Zhang, Tong},
  journal = {Ann. Stat.},
  title   = {Statistical Behavior and Consistency of Classification Methods Based on Convex Risk Minimization},
  year    = {2004},
  number  = {1},
  pages   = {56--85},
  volume  = {32},
  doi     = {10.1214/aos/1079120130},
}

@InCollection{Steinwart2006new,
  author    = {Steinwart, Ingo and Hush, Don and Scovel, Clint},
  booktitle = {Institute of {{Mathematical Statistics Lecture Notes}} - {{Monograph Series}}},
  publisher = {Institute of Mathematical Statistics},
  title     = {A New Concentration Result for Regularized Risk Minimizers},
  year      = {2006},
  address   = {Beachwood, Ohio, USA},
  pages     = {260--275},
  doi       = {10.1214/074921706000000897},
}

@Article{Yang2016Bayesian,
  author  = {Yang, Yun and Dunson, David B.},
  journal = {Ann. Stat.},
  title   = {Bayesian Manifold Regression},
  year    = {2016},
  number  = {2},
  pages   = {876--905},
  volume  = {44},
  doi     = {10.1214/15-AOS1390},
}

@InProceedings{Suzuki2021Deepa,
  author    = {Suzuki, Taiji and Nitanda, Atsushi},
  booktitle = {Advances in {{Neural Information Processing Systems}}},
  title     = {Deep Learning Is Adaptive to Intrinsic Dimensionality of Model Smoothness in Anisotropic {{Besov}} Space},
  year      = {2021},
  pages     = {3609--3621},
  publisher = {Curran Associates, Inc.},
  volume    = {34},
}

@Article{Stone1982Optimal,
  author    = {Stone, Charles J.},
  journal   = {Ann. Stat.},
  title     = {Optimal Global Rates of Convergence for Nonparametric Regression},
  year      = {1982},
  number    = {4},
  pages     = {1040--1053},
  volume    = {10},
  publisher = {Institute of Mathematical Statistics},
}

@Article{Xiang2011Logistic,
  author   = {Xiang, Dao-Hong},
  journal  = {Comput. Math. Appl.},
  title    = {Logistic Classification with Varying {{Gaussians}}},
  year     = {2011},
  number   = {2},
  pages    = {397--407},
  volume   = {61},
  doi      = {10.1016/j.camwa.2010.11.016},
  fjournal = {Computers & Mathematics with Applications},
}

@Article{Bos2022Convergence,
  author  = {Bos, Thijs and {Schmidt-Hieber}, Johannes},
  journal = {Electron. J. Stat.},
  title   = {Convergence Rates of Deep {{ReLU}} Networks for Multiclass Classification},
  year    = {2022},
  number  = {1},
  volume  = {16},
  doi     = {10.1214/22-EJS2011},
}

@Article{Alquier2019Estimation,
  author    = {Alquier, Pierre and Cottet, Vincent and Lecu{\'e}, Guillaume},
  journal   = {Ann. Stat.},
  title     = {Estimation Bounds and Sharp Oracle Inequalities of Regularized Procedures with {{Lipschitz}} Loss Functions},
  year      = {2019},
  number    = {4},
  pages     = {2117--2144},
  volume    = {47},
  doi       = {10.1214/18-AOS1742},
  publisher = {Institute of Mathematical Statistics},
}

@Article{Wen2025Optimal,
  author  = {Wen, Hongwei and Betken, Annika and Hang, Hanyuan},
  journal = {International Conference on Learning Representations},
  title   = {Optimal Learning of Kernel Logistic Regression for Complex Classification Scenarios},
  year    = {2025},
  pages   = {53918--53958},
}

@Article{Zhang2024Classification,
  author  = {Zhang, Zihan and Shi, Lei and Zhou, Ding-Xuan},
  journal = {J. Mach. Learn. Res.},
  title   = {Classification with Deep Neural Networks and Logistic Loss},
  year    = {2024},
  number  = {125},
  pages   = {1--117},
  volume  = {25},
}

@Article{Bartlett2006Convexity,
  author  = {Bartlett, Peter L and Jordan, Michael I and McAuliffe, Jon D},
  journal = {J. Amer. Stat. Assoc.},
  title   = {Convexity, Classification, and Risk Bounds},
  year    = {2006},
  number  = {473},
  pages   = {138--156},
  volume  = {101},
  doi     = {10.1198/016214505000000907},
}

@Book{Bergh1976Interpolation,
  author     = {Bergh, J{\"o}ran and L{\"o}fstr{\"o}m, J{\"o}rgen},
  publisher  = {Springer},
  title      = {Interpolation {{Spaces}}: {{An Introduction}}},
  year       = {1976},
  address    = {Berlin, Heidelberg},
  series     = {Grundlehren Der Mathematischen {{Wissenschaften}}},
  volume     = {223},
  doi        = {10.1007/978-3-642-66451-9},
  shorttitle = {Interpolation {{Spaces}}},
}

@Article{Steinwart2007Fast,
  author  = {Steinwart, Ingo and Scovel, Clint},
  journal = {Ann. Stat.},
  title   = {Fast Rates for Support Vector Machines Using {{Gaussian}} Kernels},
  year    = {2007},
  number  = {2},
  volume  = {35},
  doi     = {10.1214/009053606000001226},
}

@Article{Tsybakov2004Optimal,
  author    = {Tsybakov, Alexander B.},
  journal   = {Ann. Stat.},
  title     = {Optimal Aggregation of Classifiers in Statistical Learning},
  year      = {2004},
  number    = {1},
  pages     = {135--166},
  volume    = {32},
  doi       = {10.1214/aos/1079120131},
  publisher = {Institute of Mathematical Statistics},
  ranking   = {rank2},
}

@Article{Hamm2021Adaptive,
  author    = {Hamm, Thomas and Steinwart, Ingo},
  journal   = {Ann. Stat.},
  title     = {Adaptive Learning Rates for Support Vector Machines Working on Data with Low Intrinsic Dimension},
  year      = {2021},
  number    = {6},
  pages     = {3153--3180},
  volume    = {49},
  doi       = {10.1214/21-AOS2078},
  publisher = {Institute of Mathematical Statistics},
}

@Article{Aronszajn1950Theory,
  author  = {Aronszajn, N.},
  journal = {Trans. Amer. Math. Soc.},
  title   = {Theory of Reproducing Kernels},
  year    = {1950},
  number  = {3},
  pages   = {337--404},
  volume  = {68},
  doi     = {10.1090/S0002-9947-1950-0051437-7},
}

@Article{Farrell2021Deep,
  author  = {Farrell, Max H. and Liang, Tengyuan and Misra, Sanjog},
  journal = {Econometrica},
  title   = {Deep Neural Networks for Estimation and Inference},
  year    = {2021},
  number  = {1},
  pages   = {181--213},
  volume  = {89},
  doi     = {10.3982/ECTA16901},
}

@Book{Cucker2007Learning,
  author     = {Cucker, Felipe and Zhou, Ding-Xuan},
  publisher  = {Cambridge University Press},
  title      = {Learning {{Theory}}: {{An Approximation Theory Viewpoint}}},
  year       = {2007},
  shorttitle = {Learning {{Theory}}},
}

@Book{Steinwart2008Support,
  author    = {Steinwart, Ingo and Christmann, Andreas},
  publisher = {Springer},
  title     = {Support {{Vector Machines}}},
  year      = {2008},
  address   = {New York, NY},
  series    = {Information {{Science}} and {{Statistics}}},
  doi       = {10.1007/978-0-387-77242-4},
}

@Article{Lv2021Optimal,
  author    = {Lv, Fusheng and Fan, Jun},
  journal   = {Anal. Appl.},
  title     = {Optimal Learning with {{Gaussians}} and Correntropy Loss},
  year      = {2021},
  number    = {01},
  pages     = {107--124},
  volume    = {19},
  doi       = {10.1142/S0219530519410124},
  publisher = {World Scientific Publishing Co.},
}

@Article{Ye2009SVM,
  author    = {Ye, Gui-Bo and Zhou, Ding-Xuan},
  journal   = {Anal. Appl.},
  title     = {{{SVM}} Learning and ${{L_p}}$ approximation by {{Gaussians}} on {{Riemannian}} manifolds},
  year      = {2009},
  number    = {03},
  pages     = {309--339},
  volume    = {07},
  doi       = {10.1142/S0219530509001384},
  publisher = {World Scientific Publishing Co.},
}

@Article{Lei2024Pairwise,
  author  = {Lei, Guanhang and Shi, Lei},
  journal = {Adv. Comput. Math.},
  title   = {Pairwise Ranking with {{Gaussian}} Kernel},
  year    = {2024},
  number  = {4},
  pages   = {70},
  volume  = {50},
  doi     = {10.1007/s10444-024-10165-0},
}

@Article{Xiang2009Classification,
  author  = {Xiang, Dao-Hong and Zhou, Ding-Xuan},
  journal = {J. Mach. Learn. Res.},
  title   = {Classification with {{Gaussians}} and Convex Loss},
  year    = {2009},
  number  = {49},
  pages   = {1447--1468},
  volume  = {10},
}

@Article{Eberts2013Optimal,
  author    = {Eberts, Mona and Steinwart, Ingo},
  journal   = {Electron. J. Stat.},
  title     = {Optimal Regression Rates for {{SVMs}} Using {{Gaussian}} Kernels},
  year      = {2013},
  pages     = {1--42},
  volume    = {7},
  doi       = {10.1214/12-EJS760},
  publisher = {{Institute of Mathematical Statistics and Bernoulli Society}},
}

@Article{Ye2008Learning,
  author  = {Ye, Gui-Bo and Zhou, Ding-Xuan},
  journal = {Adv. Comput. Math.},
  title   = {Learning and Approximation by {{Gaussians}} on {{Riemannian}} Manifolds},
  year    = {2008},
  number  = {3},
  pages   = {291--310},
  volume  = {29},
  doi     = {10.1007/s10444-007-9049-0},
}

@Article{Tsuji2021Estimation,
  author    = {Tsuji, Kazuma and Suzuki, Taiji},
  journal   = {Electron. J. Stat.},
  title     = {Estimation Error Analysis of Deep Learning on the Regression Problem on the Variable Exponent {{Besov}} Space},
  year      = {2021},
  number    = {1},
  pages     = {1869--1908},
  volume    = {15},
  doi       = {10.1214/21-EJS1828},
  publisher = {{Institute of Mathematical Statistics and Bernoulli Society}},
}

@Article{Karp1989Volume,
  author    = {Karp, Leon and Pinsky, Mark},
  journal   = {Bull. Lond. Math. Soc.},
  title     = {Volume of a Small Extrinsic Ball in a Submanifold},
  year      = {1989},
  number    = {1},
  pages     = {87--92},
  volume    = {21},
  doi       = {10.1112/blms/21.1.87},
  publisher = {Oxford Academic},
}

@Book{Lee2018Introduction,
  author    = {Lee, John M.},
  publisher = {Springer International Publishing},
  title     = {Introduction to {{Riemannian Manifolds}}},
  year      = {2018},
  address   = {Cham},
  series    = {Graduate {{Texts}} in {{Mathematics}}},
  volume    = {176},
  doi       = {10.1007/978-3-319-91755-9},
}

@Article{Tang2026Adaptive,
  author    = {Tang, Tao and Wu, Nan and Cheng, Xiuyuan and Dunson, David},
  journal   = {Ann. Stat.},
  title     = {Adaptive {{Bayesian}} Regression on Data with Low Intrinsic Dimensionality},
  year      = {2026},
  number    = {2},
  pages     = {1080--1099},
  volume    = {54},
  doi       = {10.1214/25-AOS2605},
  publisher = {Institute of Mathematical Statistics},
}

@Misc{Cheng2026Learninga,
  author        = {Cheng, Xiuyuan and Wu, Nan},
  title         = {Learning Manifold Diffusion Semigroups from Graph Transition Matrices},
  year          = {2026},
  archiveprefix = {arXiv},
  doi           = {10.48550/arXiv.2605.25383},
  eprint        = {2605.25383},
  number        = {arXiv:2605.25383},
  primaryclass  = {stat.ML},
  publisher     = {arXiv},
}

@Misc{Pesenson2015Approximations,
  author        = {Pesenson, Isaac Z.},
  title         = {Approximations in {$L_p$}-Norms and {Besov} Spaces on Compact Manifolds},
  year          = {2015},
  howpublished  = {arXiv:1502.03159},
  doi           = {10.48550/arXiv.1502.03159},
}

@Article{Kansa1990Multiquadrics,
  author  = {Kansa, E. J.},
  title   = {Multiquadrics---A Scattered Data Approximation Scheme with Applications to Computational Fluid-Dynamics---{II}. Solutions to Parabolic, Hyperbolic and Elliptic Partial Differential Equations},
  journal = {Comput. Math. Appl.},
  year    = {1990},
  volume  = {19},
  number  = {8--9},
  pages   = {147--161},
  doi     = {10.1016/0898-1221(90)90271-K},
}

@Article{Gilani2019Approximating,
  author  = {Gilani, Faheem and Harlim, John},
  journal = {J. Comput. Phys.},
  title   = {Approximating Solutions of Linear Elliptic {{PDE}}'s on a Smooth Manifold Using Local Kernel},
  year    = {2019},
  pages   = {563--582},
  volume  = {395},
  doi     = {10.1016/j.jcp.2019.06.034},
}

@Article{Shi2017Convergence,
  author   = {Shi, Zuoqiang and Sun, Jian},
  journal  = {Res. Math. Sci.},
  title    = {Convergence of the Point Integral Method for {{Laplace}}--{{Beltrami}} Equation on Point Cloud},
  year     = {2017},
  number   = {1},
  pages    = {22},
  volume   = {4},
  doi      = {10.1186/s40687-017-0111-3},
  fjournal = {Research in the Mathematical Sciences},
}

@Article{Yan2023Spectral,
  author  = {Yan, Qile and Jiang, Shixiao Willing and Harlim, John},
  journal = {J. Comput. Phys.},
  title   = {Spectral Methods for Solving Elliptic {{PDEs}} on Unknown Manifolds},
  year    = {2023},
  pages   = {112132},
  volume  = {486},
  doi     = {10.1016/j.jcp.2023.112132},
}

@Article{Liang2024Solving,
  author   = {Liang, Senwei and Jiang, Shixiao W. and Harlim, John and Yang, Haizhao},
  journal  = {Appl. Comput. Harmon. Anal.},
  title    = {Solving {{PDEs}} on Unknown Manifolds with Machine Learning},
  year     = {2024},
  pages    = {101652},
  volume   = {71},
  doi      = {10.1016/j.acha.2024.101652},
  fjournal = {Applied and Computational Harmonic Analysis},
}

@Article{Trillos2023Rates,
  author   = {Garc{\'i}a Trillos, Nicol{\'a}s and Murray, Ryan and Thorpe, Matthew},
  journal  = {Sampl. Theory Signal Process. Data Anal.},
  title    = {Rates of Convergence for Regression with the Graph Poly-{{Laplacian}}},
  year     = {2023},
  number   = {2},
  pages    = {35},
  volume   = {21},
  doi      = {10.1007/s43670-023-00075-5},
  fjournal = {Sampling Theory, Signal Processing, and Data Analysis},
}

@Article{Davydov2023Error,
  author  = {Davydov, Oleg},
  journal = {Adv. Comput. Math.},
  title   = {Error Bounds for a Least Squares Meshless Finite Difference Method on Closed Manifolds},
  year    = {2023},
  number  = {4},
  pages   = {48},
  volume  = {49},
  doi     = {10.1007/s10444-023-10044-0},
}

@Article{Sanz-Alonso2024Analysis,
  author   = {{Sanz-Alonso}, Daniel and Waniorek, Nathan},
  journal  = {SIAM/ASA J. Uncertain. Quantif.},
  title    = {Analysis of a Computational Framework for {{Bayesian}} Inverse Problems: Ensemble {{Kalman}} Updates and {{MAP}} Estimators under Mesh Refinement},
  year     = {2024},
  number   = {1},
  pages    = {30--68},
  volume   = {12},
  doi      = {10.1137/23M1567035},
  fjournal = {SIAM/ASA Journal on Uncertainty Quantification},
}

@Article{Wu1993Local,
  author   = {Wu, Zong-min and Schaback, Robert},
  journal  = {IMA J. Numer. Anal.},
  title    = {Local Error Estimates for Radial Basis Function Interpolation of Scattered Data},
  year     = {1993},
  number   = {1},
  pages    = {13--27},
  volume   = {13},
  doi      = {10.1093/imanum/13.1.13},
  fjournal = {IMA Journal of Numerical Analysis},
}

@Article{Fuselier2012Scattered,
  author   = {Fuselier, Edward and Wright, Grady B.},
  journal  = {SIAM J. Numer. Anal.},
  title    = {Scattered Data Interpolation on Embedded Submanifolds with Restricted Positive Definite Kernels: {{Sobolev}} Error Estimates},
  year     = {2012},
  number   = {3},
  pages    = {1753--1776},
  volume   = {50},
  doi      = {10.1137/110821846},
  fjournal = {SIAM Journal on Numerical Analysis},
}

\end{document}